%% file: neurips_2026.tex
\documentclass{article}

 \usepackage[preprint]{neurips_2026}

\usepackage{url}
\usepackage{booktabs}
\usepackage{hyperref}
\usepackage{graphicx}
\usepackage{microtype}
\usepackage{multicol}
\usepackage{multirow}
\usepackage{lipsum}
\usepackage{enumitem}
\usepackage{pifont}
\usepackage{colortbl}
\usepackage{tabularx}
\usepackage{caption} 
\usepackage{float}
\usepackage{amsmath,amsfonts,amssymb,bbm}

\usepackage{bm}
\usepackage{lineno}
\usepackage{makecell}
\usepackage{array}
\usepackage{wrapfig}
\usepackage{tikz}
\usepackage[normalem]{ulem}
\usepackage{marvosym}
\usepackage{tcolorbox}
\tcbuselibrary{skins, breakable, theorems}
\usepackage{subcaption}  
\usepackage{float}  
\usepackage{listings}
\usepackage{fontawesome5}
\usepackage{diagbox}

\usepackage{tikz}
\usepackage{amsmath}
\usepackage{graphicx}
\usepackage{caption}
\usepackage{adjustbox}
\usetikzlibrary{3d, arrows.meta, positioning, calc}

\usepackage{lineno}

\usepackage[linesnumbered,ruled,vlined]{algorithm2e} 
\usepackage{amsmath}  
\usepackage{amssymb}  
\usetikzlibrary{shapes, positioning, fit}
\tcbuselibrary{skins, breakable}

\usepackage{CJKutf8}

\usepackage{wrapfig}
\usepackage{tikz}
\usepackage{pgfplots}
\pgfplotsset{compat=1.18}
\usetikzlibrary{arrows.meta}

\usepackage{arydshln}

\newcommand{\latbp}{\textsc{LanBo}}
\newcommand{\tbt}{\textsc{{PreAda}}}
\newcommand{\tdt}{\textsc{InAda}}

\definecolor{1c}{RGB}{194, 213, 247}
\definecolor{2c}{RGB}{252, 225, 198}
\definecolor{3c}{RGB}{29, 108, 171}
\definecolor{4c}{RGB}{255, 116, 16}

\definecolor{darkred}{rgb}{0.7, 0, 0}
\definecolor{darkblue}{rgb}{0, 0., 0.7}
\definecolor{purple}{rgb}{0.7, 0, 0.7}
\definecolor{darkyellow}{rgb}{0.7, 0.5, 0}
\definecolor{darkgreen}{rgb}{0, 0.7, 0}

\hypersetup{colorlinks=true, citecolor=darkblue, linkcolor=darkblue, urlcolor=darkblue}

\definecolor{skyblue}{rgb}{0.75 0.88 0.98}
\definecolor{pink}{rgb}{1.0, 0.752, 0.796}

\usepackage{amsmath}
\newtheorem{theorem}{Theorem}

\newtheorem{definition}{Definition}
\newenvironment{proof}{{\noindent\it Proof.}\quad}{\hfill $\square$\par}

\newenvironment{solution}{{\noindent\it Solution.}\quad}{\hfill $\square$\par}

\definecolor{deepred}{rgb}{0.6, 0, 0}

\definecolor{VoteBlue}{HTML}{C7DBF4}
\definecolor{RewardOrange}{HTML}{F6D7AA}
\definecolor{OracleGreen}{HTML}{BFE5CF}

\newcommand{\BlueShade}[1]{%
  \edef\val{\fpeval{min(max(#1,0),1)}}%
  \edef\shade{\fpeval{round(100 - 100*\val,0)}}%
  \colorbox{VoteBlue!\shade}{\makebox[0.9cm][c]{\color{black}#1}}%
}

\newcommand{\OrangeShade}[1]{%
  \edef\val{\fpeval{min(max(#1,0),1)}}%
  \edef\shade{\fpeval{round(100 - 100*\val,0)}}%
  \colorbox{RewardOrange!\shade}{\makebox[0.9cm][c]{\color{black}#1}}%
}

\newcommand{\GreenShade}[1]{%
  \edef\val{\fpeval{min(max(#1,0),1)}}%
  \edef\shade{\fpeval{round(100 - 100*\val,0)}}%
  \colorbox{OracleGreen!\shade}{\makebox[0.9cm][c]{\color{black}#1}}%
}

\newcommand{\ColorBar}[2]{%
  \begin{tikzpicture}
    \shade[left color=#1, right color=white] (0,0) rectangle (1.0cm,0.2cm);

    \fill[white] (1.0cm,0) rectangle (1.45cm,0.2cm);

    \draw (0,0) rectangle (1.45cm,0.2cm);

    \draw (1.0cm,0) -- (1.0cm,0.2cm);

    \node[below] at (0,-0.02) {\footnotesize 0.0};
    \node[below] at (1.0cm,-0.02) {\footnotesize 1.0};
    \node[below] at (1.45cm,-0.02) {\footnotesize \textgreater 1};

    \node[above] at (0.875cm,0.55) {\footnotesize #2};
  \end{tikzpicture}%
}

\makeatletter
\renewcommand{\tagform@}[1]{%
  \maketag@@@{\textcolor{deepred}{\textbf{(\ignorespaces#1\unskip\@@italiccorr)}}}%
}
\makeatother

\title{On the Overscaling Curse of Parallel Thinking: System Efficacy Contradicts Sample Efficiency}

\author{%
Yiming Wang, ~Zhuosheng Zhang, ~Rui Wang\\
School of Computer Science, Shanghai Jiao Tong University\\
\texttt{yiming.wang@sjtu.edu.cn}
}

\begin{document}

\maketitle

\input{0-abstract}

\input{1-intro}

\input{2-preliminary}

\input{3-overscaling}

\input{4-breaking}

\input{5-application}

\input{6-analysis}

\input{7-conclusion}

\bibliographystyle{plain}
\bibliography{example_paper}


\appendix

\input{appendix}



\end{document}

%% file: 0-abstract.tex
\vspace{-0.2in}
\begin{abstract}



Parallel thinking improves LLM reasoning through multi-path sampling and aggregation.
In standard evaluations, due to a lack of sample-specific priors, all samples share a global budget chosen to maximize dataset accuracy.
However, many samples reach their best accuracy with much smaller budgets, causing low budget utilization.
This {\it contradiction between system efficacy and sample efficiency} constitutes the {\bf Overscaling Curse}.
In this paper, we first provide a formal analysis of the overscaling curse and quantify its prevalence and severity in real-world systems.
To break it, we propose {\bf \underline{Lat}ent \underline{B}udget Predict\underline{o}r (\latbp)}, which probes model latent representations to predict sample-specific optimal budgets.
\latbp{} significantly improves budget utilization while maintaining dataset accuracy.
We further integrate \latbp{} into the full decoding pipeline, inspiring {\bf \underline{Pre}-decoding Budget \underline{Ada}ptation (\tbt)}, a paradigm that allocates budgets before decoding to preserve decoding-time parallelization.
\latbp{} substantially improves hardware-aware efficiency in latency and memory, demonstrating both its practical value and the promise of \tbt{} for efficient parallel decoding.

\end{abstract}

%% file: 1-intro.tex
\vspace{-0.1in}
\section{Introduction}
\label{sec:intro}

Large language models (LLMs) continually update their distributions by large-scale corpora training \citep{vaswani2017attention,brown2020language}, enabling them to encode broad world knowledge \citep{achiam2023gpt,singh2025openai} and evolve strong reasoning abilities \citep{guo2025deepseek}.
As model capacity increases, recent work shows that model distributions contain latent information that is not fully revealed by simple autoregressive decoding \citep{chen2025not}.
This motivates scaling compute \citep{snell2024scaling,muennighoff2025s1} during inference to elicit model potential, termed test-time scaling (TTS).

Among TTS, parallel thinking represents a distinct paradigm that broadens the space of reasoning exploration \citep{fu2025deep,zheng2025parallel}.
This paradigm typically pre-allocates a parallelism budget $N \in \mathbb{N}^+$.
During decoding, the model first generates $N$ {\it independent} paths in parallel via multinomial sampling \citep{hinton2015distilling, holtzman2019curious}, promoting diversity and broader coverage of the output distribution.
Then, an aggregation strategy like majority voting \citep{wang2022self} or external verifiers \citep{liu2025skywork} selects the final answer from the $N$ candidates.

\begin{figure}[t]
    \vspace{-0.1in}
    \centering
    \includegraphics[width=\columnwidth]{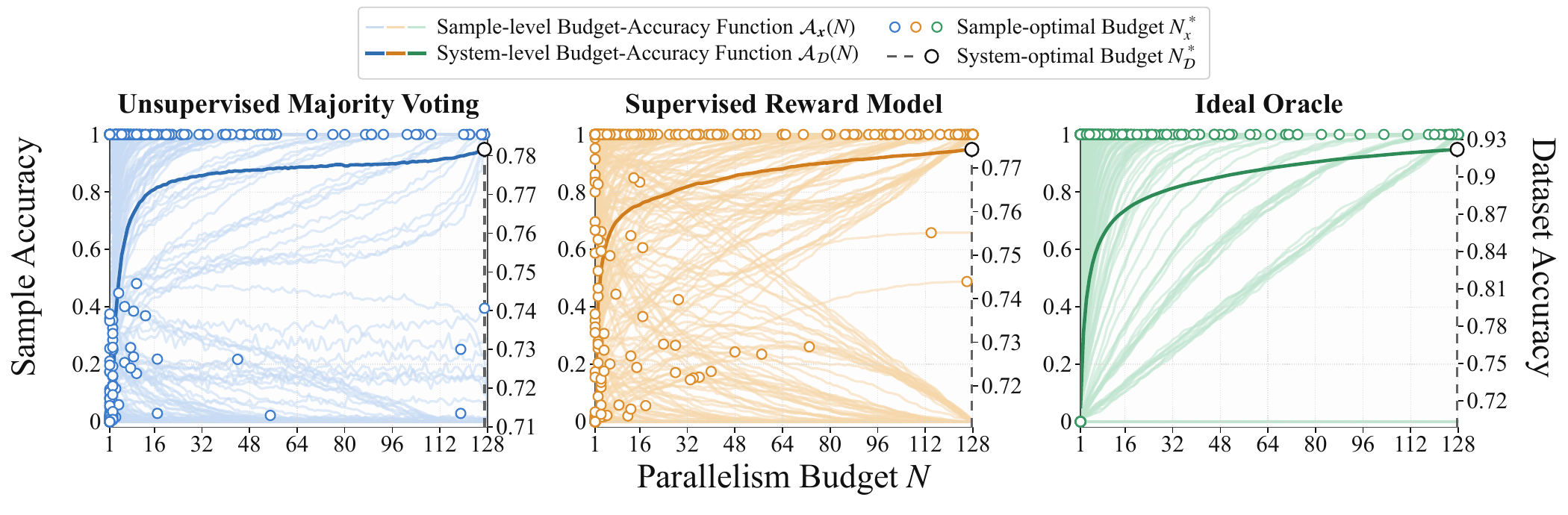}
    \vspace{-0.27in}
    \caption{{\bf Illustration of the Overscaling Curse in Parallel Thinking}. This figure illustrates a single system $(\pi, \mathcal{D})$, \emph{i.e.}, evaluating model $\pi$ on dataset $\mathcal{D}$. Here $\pi = \texttt{Qwen2.5-7B}$ and $\mathcal{D} = \texttt{MATH500}$. 
    Each subfigure corresponds to one answer aggregation strategy.
    The single thick curve shows how overall dataset accuracy varies with the budget, and thin curves show how sample accuracy varies with the budget for each sample $\bm{x} \in \mathcal{D}$. We term these curves the {\it ``budget-accuracy'' functions}.
    Hollow markers on each curve indicate the minimum budget that achieves the highest accuracy, which we term the {\it ``optimal budget''}.
    See \textcolor{deepred}{Section \ref{sec:global}} for experimental setups and \textcolor{deepred}{Section \ref{sec:formalization}} for detailed function computation.
    Illustrations in other systems are shown in \textcolor{deepred}{Figures \ref{fig:illustration-qwen25-amc} -- \ref{fig:illustration-qwen3-aime25}} (\textcolor{deepred}{Appendix \ref{appe:illustration}}).}
    \vspace{-0.25in}
    \label{fig:illustration-qwen25-math}
\end{figure}

In practice, parallel thinking is usually evaluated at the system level: a model-dataset pair $(\pi, \mathcal{D})$ forms a system, and performance is measured by the overall accuracy of model $\pi$ on dataset $\mathcal{D}$.
Due to the lack of sample-specific priors, all samples $\bm{x} \in \mathcal{D}$ share a global budget $N_{\mathcal{D}}$.
Following the TTS principle of prioritizing performance over cost \citep{snell2024scaling}, this budget should first \textit{\textbf{prioritize system efficacy}}.
Thus, the ideal global $N_{\mathcal{D}}$ is the minimum budget that \textit{\textbf{maximizes dataset accuracy}}, which we call the system-optimal budget.
In contrast, if released from this global constraint, each sample $\bm{x}$ would have its own ideal budget $N_{\bm{x}}$, since sampling randomness can induce different expected accuracies under different budgets.
Analogously, we call the sample-optimal budget the minimum budget that maximizes the expected accuracy for that sample.

However, when examining real-world systems, \emph{a consistent observation emerges: the system-optimal budget often exceeds the sample-optimal budgets for many individual samples}, as shown in \textcolor{deepred}{Figure \ref{fig:illustration-qwen25-math}} and \textcolor{deepred}{\ref{fig:illustration-qwen25-amc} -- \ref{fig:illustration-qwen3-aime25}}.
This means that, when these samples are forced to follow this global system-optimal budget rather than their own optima, their \textit{\textbf{budget utilization decreases}}, thereby \textit{\textbf{undermining sample efficiency}}.
This reveals \emph{a fundamental contradiction induced by the global-budget constraint: maximizing system efficacy can come at the cost of sample efficiency}.
We refer to this as the \textbf{Overscaling Curse}.
In this paper, we start from the above phenomenon and provide a formal analysis of the overscaling curse, together with concrete metrics to quantify its severity.
Extensive empirical results show that the overscaling curse is both prevalent and severe in real-world systems.

Next, we explore breaking the overscaling curse by improving budget utilization while preserving dataset accuracy. Our foremost step is to relax the global-budget constraint and allow sample-specific allocation. Since each sample has its own optimal budget, we reduce the problem to predicting this sample-optimal budget. As the budget depends on both the sample itself and the model's perception of it, we propose \textbf{Latent Budget Predictor (\latbp)}, which probes model latent representations of the sample to predict its optimal budget. Empirically, \latbp{} improves budget utilization by $3-6\times$ on average while maintaining dataset accuracy, and a single training can adapt to multiple domains.

Finally, we integrate \latbp{} into the full parallel decoding pipeline to evaluate its practical hardware-aware efficiency beyond merely reducing the budget size.
This integration inspires \textbf{Pre-decoding Budget Adaptation (\tbt)}, a new efficient parallel decoding paradigm that allocates budgets before decoding, in contrast to existing efficient methods that adapt budgets during decoding \citep{li2024escape,wang2025make,fu2025deep}.
By avoiding decoding-time blocking, \latbp{} better preserves full hardware parallelization despite its predictor cost.
Empirically, \latbp{} reduces both latency and memory by over $50\%$ compared to standard parallel decoding, significantly outperforming existing methods, and with less than $2\%$ performance fluctuation, demonstrating its practical value for efficient parallel decoding.
More importantly, as the first exploration under \tbt{}, \latbp{} demonstrates the feasibility of this paradigm and opens a promising direction for future efficient parallel decoding research.

%% file: 2-preliminary.tex
\vspace{-0.05in}
\section{Global Experimental Setup}
\label{sec:global}

\vspace{-0.05in}
\subsection{Implementation of Parallel Thinking}
\label{sec:parallel-thinking}

Let $\pi(\bm{y}|\bm{x})$ be an autoregressive language model that generates an output sequence $\bm{y} = (y_1, \dots, y_{T'})$ conditioned on an input sequence $\bm{x} = (x_1, \dots, x_T)$, where each token $y_t$ is sampled sequentially as $y_t \sim \pi(\cdot \mid \bm{x}, \bm{y}_{\prec t})$.
Parallel thinking consists of two progressive stages \citep{fu2025deep,zeng2025pushing}: {\it multi-path sampling} and {\it answer aggregation}. In the first stage, $\pi$ independently generates $N$ distinct output sequences $\{\bm{y}^{i}\}_{i=1}^N$ for the same $\bm{x}$, each following the same autoregressive process. In the second stage, an aggregation function $\mathcal{F}$ combines these $N$ sequences to produce the final output $\hat{\bm{y}} = \mathcal{F}(\bm{y}^{1}, \dots, \bm{y}^{N})$.

\vspace{-0.1in}
\paragraph{Multi-path Sampling.}
we use multinomial sampling with combined top-$k$, top-$p$, and temperature $T$ \citep{fan2018hierarchical,holtzman2019curious,hinton2015distilling}.
Following prior work \citep{yang2025qwen3,guo2025deepseek}, we set $k=20$, $p=0.95$, and $T=0.6$ by default. Sampling is implemented using the \texttt{vLLM} framework with full parallelization on 80G A100 GPUs.

\vspace{-0.1in}
\paragraph{Answer Aggregation.}
We consider two strategies: \emph{unsupervised majority voting} (self-consistency \citep{wang2022self}) and \emph{supervised external verifiers}. For the latter, we use a general-purpose reward model (\textit{Skywork-Reward-V2} \citep{liu2025skywork}) and include an ideal oracle (no verification error) as an upper bound of verifiers. These choices ensure our findings are not tied to specific aggregation methods.

\vspace{-0.05in}
\subsection{System Selection: Models and Datasets}
\label{sec:system}

In this paper, a system is defined as a model-dataset pair $(\pi, \mathcal{D})$, and the phenomena observed when evaluating $\pi$ on $\mathcal{D}$ form the meta-object of our study. All systems considered are $\{(\pi, \mathcal{D}) \in \bm{\pi} \times \bm{\mathcal{D}}\}$.

\vspace{-0.1in}
\paragraph{Model Set $\bm{\pi}$.}
we select four models for our evaluations: \texttt{Qwen2.5-7B} \citep{yang2024qwen2}, \texttt{Llama3.1-8B} \citep{grattafiori2024llama}, \texttt{Deepseek-R1-Distill-Qwen-7B} \citep{guo2025deepseek}, and \texttt{Qwen3-4B} \citep{yang2025qwen3} (the \texttt{Instruct-2507} version with \texttt{enable\_thinking=True}). The first two are non-reasoning models, and the last two are reasoning models.
{\it Together, they provide a broad diversity in model series and reasoning paradigms.}

\vspace{-0.1in}
\paragraph{Dataset Set $\bm{\mathcal{D}}$.}
we select seven datasets: \texttt{MATH500} \citep{hendrycks2021measuring}, \texttt{AMC} \citep{amc}, \texttt{AIME24} \citep{aime2024}, \texttt{AIME25} \citep{aime2025} from the mathematical domain with progressively increasing difficulty; and \texttt{GPQA} \citep{rein2024gpqa}, \texttt{MMLU-Pro} \citep{wang2024mmlu}, \texttt{BrowseComp} (en \citep{wei2025browsecomp} + zh \citep{zhou2025browsecomp}) from diverse domains, including general QA, multi-subject knowledge, and agent tasks.
{\it Together, they provide a broad diversity in difficulty and domain.}

%% file: 3-overscaling.tex
\vspace{-0.05in}
\section{Revealing the Overscaling Curse}
\label{sec:overscaling}

\vspace{-0.05in}
\subsection{Issue Formalization}
\label{sec:formalization}

The overscaling curse stems from \emph{the contradiction between system efficacy and sample efficiency induced by the global-budget constraint}: the former is reflected by {\it dataset accuracy}, while the latter is reflected by {\it budget utilization}.
We therefore begin by formalizing the \textbf{Budget-Accuracy Function} $\mathcal{A}(N)$, which links budget to accuracy at both sample and dataset levels.

We first consider a sample $\bm{x} \in \mathcal{D}$ without global constraints.
Let $\mathcal{A}_{\bm{x}}(N)$ denote its expected accuracy under budget $N \in [1,N_{\max}]$, where $N_{\max}$ is the largest budget considered\footnote{
We set \(N_{\max}=128\) in our experiments. During pre-evaluation, we started from $N=2$ and doubled $N$ until the change in dataset accuracy fell below $0.01$, indicating performance convergence. This occurred for most cases by $N=2^7$.
}.
For a fixed $N$, one run samples $N$ candidates and aggregates them, producing a binary correctness outcome.
Since sampling is stochastic, $\mathcal{A}_{\bm{x}}(N)$ should be estimated as the expected correctness over $R$ repeated $N$-sampling runs.
Naively computing the full function requires repeating this for every $N \in [1,N_{\max}]$, leading to $\mathcal{O}(N_{\max}^2 R)$ sampling cost.
To make this tractable, we use subsampling:
For $\bm{x}$, we first generate a reference set of $2N_{\max}$ outputs $\{\bm{a}_{\bm{x},n}\}_{n=1}^{2N_{\max}}$.
For every $N$, we uniformly select $N$ outputs without replacement from this set, aggregate them with $\mathcal{F}$, and repeat this operation $M$ times\footnote{
We set $\tau=10^5$ in $M$ to avoid unbounded combinatorial enumeration while keeping the Monte Carlo estimation error negligible in practice.
With this setting, our stability test yields $p<0.01$ in all subsequent experiments, supporting the reliability of the estimates.
This subsampling procedure requires only one reference generation of size $2N_{\max}$ per sample, reducing the sampling cost to $\mathcal{O}(N_{\max})$ while enabling stable estimation of the full budget-accuracy function.
}, yielding:

\vspace{-0.15in}
\begin{small}
\begin{equation}
\begin{aligned}
    \mathcal{A}_{\bm{x}}(N)
    &=
    \mathbb{E}_{m \sim [M]}
    \left[
        \mathbb{I} \left(
            \mathcal{F} \big(\{\bm{a}_{\bm{x},n}^{m}\}_{n=1}^{N}\big) = \bm{y}
        \right)
    \right],
    \\
    M
    &=
    \min \left\{\tau, \binom{N_{\max}}{N}\right\},
    \qquad
    \{\bm{a}_{\bm{x},n}^{m}\}_{n=1}^{N}
    \sim
    \mathrm{Uniform}_{\mathrm{w/o\ repl.}}
    \left(
    \{\bm{a}_{\bm{x},n}\}_{n=1}^{2N_{\max}}, N
    \right).
\end{aligned}
\label{eq:sample-acc}
\end{equation}
\end{small}
\vspace{-0.15in}

Following the TTS principle of prioritizing performance over cost \citep{snell2024scaling}, the ideal budget for $\bm{x}$ is the minimum budget that achieves its best accuracy. We term it the {\bf Sample-optimal Budget} $N_{\bm{x}}^*$,
\emph{i.e.}, $N_{\bm{x}}^* = \min ( \mathrm{argmax}_{N \in [1,N_{\max}]}~\mathcal{A}_{\bm{x}}(N) ).$\refstepcounter{equation}\label{eq:sample-optimal-budget} \textcolor{deepred}{\textbf{\textup{(\theequation)}}}
Suppose that $\bm{x}$ is allocated an actual budget $N_{\bm{x}}$. Then the ratio between $N_{\bm{x}}^*$ and $N_{\bm{x}}$ can be defined as its {\bf Budget Utilization} $\mathcal{U}_{\bm{x}}$, \emph{i.e.}, $\mathcal{U}_{\bm{x}} = N_{\bm{x}}^*/N_{\bm{x}}$,\refstepcounter{equation}\label{eq:sample-budget-utilization} \textcolor{deepred}{\textbf{\textup{(\theequation)}}} measuring how much of the allocated budget is actually needed by $\bm{x}$.
When $\mathcal{U}_{\bm{x}} < 1$, the allocated budget is redundant, since any portion beyond $N_{\bm{x}}^*$ brings no positive accuracy gain for it.

Now, we consider the practical evaluation setting where \emph{sample-specific priors are unavailable, so a global budget is allocated to all samples}.
In this case, the dataset-level budget-accuracy function is
$\mathcal{A}_{\mathcal{D}}(N)=\mathbb{E}_{\bm{x}\sim P_{\mathcal{D}}}[\mathcal{A}_{\bm{x}}(N)]$.
Analogously, the {\bf System-optimal Budget} is defined as
$N_{\mathcal{D}}^*=\min(\operatorname*{argmax}_{N\in[1,N_{\max}]}\mathcal{A}_{\mathcal{D}}(N))$,
and the {\bf Expected Budget Utilization} over $\mathcal{D}$ is
$\mathbb{E}_{\bm{x} \sim P_{\mathcal{D}}}[\mathcal{U}_{\bm{x}}]$.

Under this global-budget constraint, \emph{to maximize dataset accuracy}, the global budget should be set to $N_{\mathcal{D}}^*$.
Consequently, each sample's budget utilization becomes
$\mathcal{U}_{\bm{x}}=N_{\bm{x}}^*/N_{\mathcal{D}}^*$.
However, $N_{\mathcal{D}}^*$ may exceed the sample-optimal budget of many individual samples.
These samples then yield $\mathcal{U}_{\bm{x}}<1$, lowering the expected budget utilization
$\mathbb{E}_{\bm{x}\sim P_{\mathcal{D}}}[\mathcal{U}_{\bm{x}}]$
and thereby \emph{undermining sample efficiency} over $\mathcal{D}$.
Based on this mechanism, \textbf{\textit{the essence of the overscaling curse can be formalized as:}}

\vspace{-0.1in}
\begin{center}
\fbox{
\begin{minipage}{0.95\textwidth}
\centering
\textbf{\textit{
With the global-budget constraint that $N_{\bm{x}} = N_{\mathcal{D}}^*$ for all $\bm{x}\in\mathcal{D}$, many $\bm{x}$ may yield $\mathcal{U}_{\bm{x}} < 1$, resulting in $\mathbb{E}_{\bm{x}\sim P_{\mathcal{D}}}[\mathcal{U}_{\bm{x}}] < 1$.
}}
\end{minipage}
}
\end{center}
\vspace{-0.1in}

In summary, the overscaling curse can be viewed as a \textbf{\textit{sample--system--sample closed-loop constraint process under the global budget}}:
(1) Individual samples exhibit heterogeneous $\mathcal{A}_{\bm{x}}(N)$, each with corresponding $N_{\bm{x}}^*$;
(2) Under the global-budget constraint, these samples are aggregated into $\mathcal{A}_{\mathcal{D}}(N)=\mathbb{E}_{\bm{x}\sim P_{\mathcal{D}}}[\mathcal{A}_{\bm{x}}(N)]$, jointly determining the system-optimal global budget $N_{\mathcal{D}}^*$;
and (3) $N_{\mathcal{D}}^*$ is then imposed back on all samples, 
constraining their $\mathcal{U}_{\bm{x}}$ and ultimately lowering $\mathbb{E}_{\bm{x}\sim P_{\mathcal{D}}}[\mathcal{U}_{\bm{x}}]$.

\subsection{Quantify the Overscaling Curse}
\label{sec:how-overscaling}

\begin{wraptable}{r}{0.55\textwidth}
\vspace{-0.65in}
\centering
\footnotesize
\setlength{\tabcolsep}{1.1mm}
\renewcommand{\arraystretch}{1.0}

\parbox{0.55\textwidth}{\centering
  \ColorBar{VoteBlue}{}%
  \qquad
  \ColorBar{RewardOrange}{}%
  \qquad
  \ColorBar{OracleGreen}{}%
}
\vspace{-0.1in}

\caption{{\it Expected Budget Utilization $\mathbb{E}_{\bm{x}\sim P_{\mathcal{D}}}[\mathcal{U}_{\bm{x}}]$} across systems $(\pi, \mathcal{D})$ under different aggregation strategies.}
\vspace{0.02in}

\resizebox{0.55\textwidth}{!}{
\begin{tabular}{lccccccc}
\toprule
\multicolumn{1}{c}{\diagbox{\bf Model $\pi$}{\bf Dataset $\mathcal{D}$}} & MATH500 & AMC & AIME24 & AIME25 & GPQA & MMLU-Pro & BrowseComp \\
\midrule

\multicolumn{8}{c}{\textbf{Unsupervised Majority Voting}} \\
\midrule
Qwen2.5-7B          
& \BlueShade{0.09} & \BlueShade{0.50} & \BlueShade{0.07} & \BlueShade{0.16} & \BlueShade{0.38} & \BlueShade{0.31} & \BlueShade{0.23}
\\
Llama3.1-8B         
& \BlueShade{0.21} & \BlueShade{0.24} & \BlueShade{0.26} & \BlueShade{0.12} & \BlueShade{0.24} & \BlueShade{0.39} & \BlueShade{0.26}
\\
R1-Distill-Qwen-7B  
& \BlueShade{0.20} & \BlueShade{0.27} & \BlueShade{0.17} & \BlueShade{0.22} & \BlueShade{0.30} & \BlueShade{0.20} & \BlueShade{0.36}
\\
Qwen3-4B
& \BlueShade{0.06} & \BlueShade{0.13} & \BlueShade{0.34} & \BlueShade{0.18} & \BlueShade{0.26} & \BlueShade{0.31} & \BlueShade{0.33}
\\

\midrule
\multicolumn{8}{c}{\textbf{Supervised Reward Model}} \\
\midrule
Qwen2.5-7B
& \OrangeShade{0.17} & \OrangeShade{4.41} & \OrangeShade{1.25} & \OrangeShade{0.41} & \OrangeShade{0.44} & \OrangeShade{0.36} & \OrangeShade{0.27}
\\
Llama3.1-8B
& \OrangeShade{0.82} & \OrangeShade{1.11} & \OrangeShade{0.19} & \OrangeShade{0.05} & \OrangeShade{0.35} & \OrangeShade{0.33} & \OrangeShade{0.35}
\\
R1-Distill-Qwen-7B
& \OrangeShade{0.35} & \OrangeShade{0.30} & \OrangeShade{0.20} & \OrangeShade{0.26} & \OrangeShade{0.35} & \OrangeShade{0.24} & \OrangeShade{0.39}
\\
Qwen3-4B
& \OrangeShade{0.11} & \OrangeShade{4.37} & \OrangeShade{1.42} & \OrangeShade{0.30} & \OrangeShade{0.28} & \OrangeShade{0.45} & \OrangeShade{0.32}
\\

\midrule
\multicolumn{8}{c}{\textbf{Ideal Oracle}} \\
\midrule
Qwen2.5-7B
& \GreenShade{0.10} & \GreenShade{0.27} & \GreenShade{0.28} & \GreenShade{0.18} & \GreenShade{0.33} & \GreenShade{0.27} & \GreenShade{0.18}
\\
Llama3.1-8B
& \GreenShade{0.22} & \GreenShade{0.37} & \GreenShade{0.33} & \GreenShade{0.32} & \GreenShade{0.27} & \GreenShade{0.25} & \GreenShade{0.27}
\\
R1-Distill-Qwen-7B
& \GreenShade{0.25} & \GreenShade{0.24} & \GreenShade{0.15} & \GreenShade{0.18} & \GreenShade{0.27} & \GreenShade{0.15} & \GreenShade{0.28}
\\
Qwen3-4B & \GreenShade{0.03} & \GreenShade{0.08} & \GreenShade{0.21} & \GreenShade{0.16} & \GreenShade{0.19} & \GreenShade{0.34} & \GreenShade{0.24}
\\

\bottomrule
\end{tabular}
}
\vspace{-0.2in}
\label{tab:overscaling-degree}
\end{wraptable}

We now quantify whether the overscaling curse is prevalent in real-world systems.
As formalized in \textcolor{deepred}{Section \ref{sec:formalization}}, the closed-loop process ultimately leads to low expected budget utilization $\mathbb{E}_{\bm{x}\sim P_{\mathcal{D}}}[\mathcal{U}_{\bm{x}}]$.
Therefore, we use $\mathbb{E}_{\bm{x}\sim P_{\mathcal{D}}}[\mathcal{U}_{\bm{x}}]$ as the primary metric to measure the overscaling degree.
A smaller value below $1$ indicates more severe overscaling, while values equal to or above $1$ suggest that the overscaling curse is absent.

\textcolor{deepred}{Table \ref{tab:overscaling-degree}} reports $\mathbb{E}_{\bm{x}\sim P_{\mathcal{D}}}[\mathcal{U}_{\bm{x}}]$ results across systems under different aggregation strategies.
{\it In 94\% of cases (79 out of 84 groups), $\mathbb{E}_{\bm{x}\sim P_{\mathcal{D}}}[\mathcal{U}_{\bm{x}}] < 1$, indicating that the overscaling curse is prevalent in real-world systems.}
Moreover, among these 79 cases, only one exceeds 0.5, while the remaining 78 are below 0.5.
This means that, {\it for almost all affected systems, more than half of the system-optimal budget is redundant for samples on average, showing that the overscaling curse is severe}.
Importantly, this phenomenon is consistent across aggregation strategies, suggesting that it is not an artifact of a specific aggregation but rather inherent to the system itself.
These results consistently \textbf{\textit{demonstrate the prevalence and severity of the overscaling curse in real-world systems}}.

\begin{figure}[t]
    \vspace{-0.2in}
    \centering
    \includegraphics[width=\columnwidth]{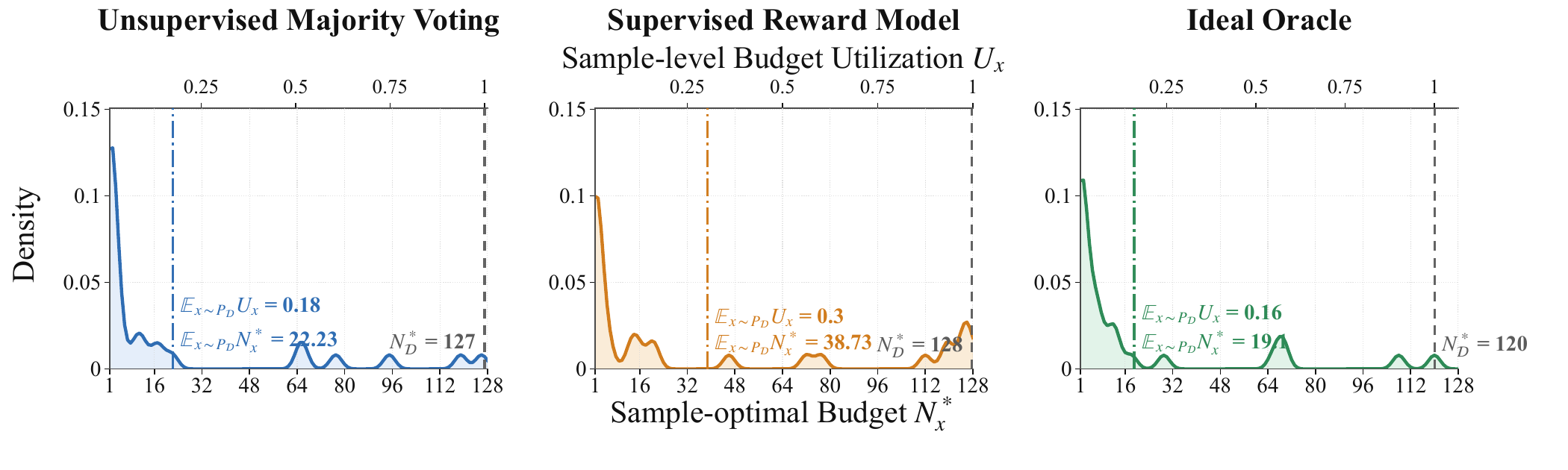}
    \vspace{-0.35in}
    \caption{Distribution of {\it Budget Utilization $\mathcal{U}_{\bm{x}}$} and corresponding sample-optimal budgets $N^*_{\bm{x}}$ in a single systems $(\pi,\mathcal{D})$ under different aggregation strategies, along with $N^*_{\mathcal{D}}$. Here $\pi = \texttt{Qwen3-4B}$ and $\mathcal{D} = \texttt{AIME25}$. See \textcolor{deepred}{Figures \ref{fig:distribution-qwen25-math} -- \ref{fig:distribution-qwen3-browsecomp}} in \textcolor{deepred}{Appendix \ref{appe:quantify}} for distributions in other systems.}
    \vspace{-0.15in}
    \label{fig:distribution-qwen3-aime25}
\end{figure}

\subsection{Better Understand the Overscaling Curse}
\label{sec:better-understanding}

After empirically verifying the overscaling curse, we trace backward from the outcome of this closed-loop process.
Since $\mathbb{E}_{\bm{x}\sim P_{\mathcal{D}}}[\mathcal{U}_{\bm{x}}]$ depends on the distribution of $\mathcal{U}_{\bm{x}}$, and $\mathcal{U}_{\bm{x}}$ is determined by both $N_{\bm{x}}^*$ and the global $N_{\mathcal{D}}^*$, we further examine them in each system.

\textcolor{deepred}{Figure \ref{fig:distribution-qwen25-math}} and \textcolor{deepred}{\ref{fig:distribution-qwen25-amc} -- \ref{fig:distribution-qwen3-browsecomp}} present the results.
A highly imbalanced pattern consistently emerges: the distribution of $N_{\bm{x}}^*$ is heavily concentrated on the left side of the budget axis, indicating that most samples achieve their best accuracy with relatively small budgets.
However, despite the sparse right region, $N_{\mathcal{D}}^*$ is often pulled toward that side by a few high-optimal-budget samples.
As a result, most left-side samples are forced to operate under a much larger global budget. This leads to very low $\mathcal{U}_{\bm{x}}$ and consequently low $\mathbb{E}_{\bm{x}\sim P_{\mathcal{D}}}[\mathcal{U}_{\bm{x}}]$ over $\mathcal{D}$, thereby causing the severe overscaling curse.

This raises a further question: why does this phenomenon emerge and lead to such a severe overscaling curse?
We further trace backward along the closed-loop process and revisit the sample-level budget-accuracy functions $\mathcal{A}_{\bm{x}}(N)$, as illustrated in \textcolor{deepred}{Figure \ref{fig:illustration-qwen25-math}} and \textcolor{deepred}{\ref{fig:illustration-qwen25-amc} -- \ref{fig:illustration-qwen3-aime25}}.
Although sample behaviors are intertwined, most $\mathcal{A}_{\bm{x}}(N)$ exhibit clear monotonic patterns.
We group these dominant patterns into four types:
{\it (1) $\mathcal{A}_{\bm{x}}(N) \equiv 1$;
(2) $\mathcal{A}_{\bm{x}}(N) \equiv 0$;
(3) $\mathcal{A}_{\bm{x}}(N)$ approximately decreases toward $0$;
and (4) $\mathcal{A}_{\bm{x}}(N)$ approximately increases toward $1$.}
All remaining cases are labeled as {\it Type-5, covering non-monotonic functions without a clear trend}.
Accordingly, $\mathcal{D}$ can be partitioned into five disjoint subsets $\{\mathcal{D}_i\}_{i=1}^{5}$, with $\mathcal{D}_i$ containing Type-$i$ samples.
For each $\mathcal{D}_i$, we denote its expected sample-optimal budget as
$\mathbb{E}_{\bm{x}\sim P_{\mathcal{D}_i}}[N_{\bm{x}}^*]$.
Formal definitions of these five types are in \textcolor{deepred}{Appendix \ref{appe:sample-type}}.

\begin{figure}[t]
    \vspace{-0.2in}
    \centering
    \includegraphics[width=\columnwidth]{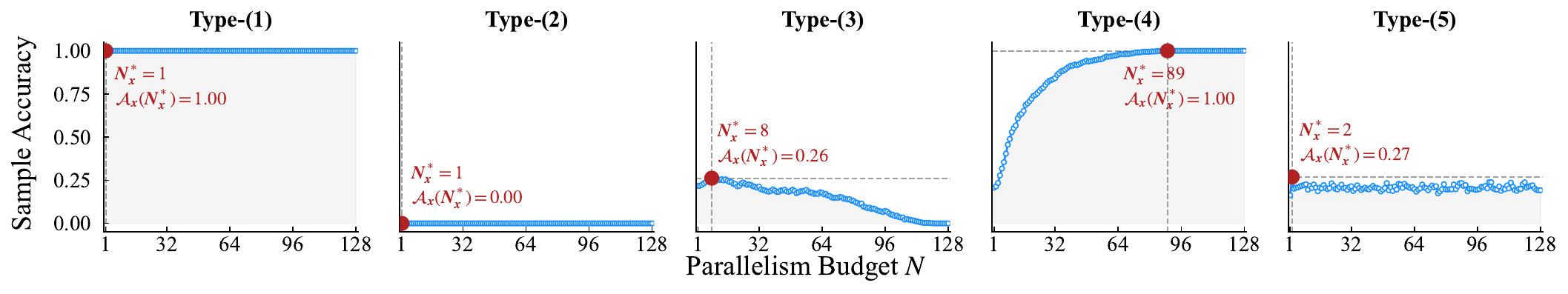}
    \vspace{-0.25in}
    \caption{
    Illustrations of $\mathcal{A}_{\bm{x}}(N)$ from the five sample types based on the monotonic patterns.
    Examples here are from \texttt{Qwen3-4B} on the \texttt{AIME25} dataset. See \textcolor{deepred}{Appendix \ref{appe:sample-type}} for formal definitions.}
    \vspace{-0.25in}
    \label{fig:sample-type-case}
\end{figure}

\textcolor{deepred}{Figure \ref{fig:sample-type-case}} presents one example illustration of $\mathcal{A}_{\bm{x}}(N)$ for the five types, with more illustrations shown in \textcolor{deepred}{Figures \ref{fig:overscaling-sample-qwen25-math-type3} -- \ref{fig:overscaling-sample-qwen3-aime25-type5}}. 
According to the monotonicity feature, samples in $\mathcal{D}_4$ naturally tend to have larger $N_{\bm{x}}^*$, especially compared to those from $\mathcal{D}_1$, $\mathcal{D}_2$, and $\mathcal{D}_3$, while $\mathcal{D}_5$ is less predictable due to its non-monotonicity.
We empirically verify this by computing all $\mathbb{E}_{\bm{x}\sim P_{\mathcal{D}_i}}[N_{\bm{x}}^*]$ under each system (see \textcolor{deepred}{Appendix \ref{appe:sample-type-statistics}}).
The results confirm that
$\mathbb{E}_{\bm{x}\sim P_{\mathcal{D}_4}}[N_{\bm{x}}^*]$
is substantially larger than that of the other types, including Type-5.
This is the key inherent property for Type-4 samples in real-world systems.

Delving deeper, we analyze each subset $\mathcal{D}_i$ by its proportion $p_i$ and total accuracy gain $\bar{\Delta}_i$ over the full budget domain.
For each $\mathcal{D}_i$, let $p_i=|\mathcal{D}_i|/|\mathcal{D}|$ and
$\bar{\Delta}_i = \mathbb{E}_{\bm{x}\sim P_{\mathcal{D}_i}} [\mathcal{A}_{\bm{x}}(N_{\max})-\mathcal{A}_{\bm{x}}(1)]$.
We find that {\it severe overscaling systems commonly satisfy $p_4 > p_3$ and $\bar{\Delta}_4 > [-\bar{\Delta}_3]_+$, where $[z]_+=\max(z,0)$} (see \textcolor{deepred}{Appendix \ref{appe:sample-type-statistics}}).
Since dataset accuracy averages over all samples, the influence of each $\mathcal{D}_i$ for $\mathcal{A}_{\mathcal{D}}(N)$ is jointly determined by $p_i$ and $\bar{\Delta}_i$.
Thus, $\mathcal{D}_4$ dominates the positive improvement of $\mathcal{A}_{\mathcal{D}}(N)$, while $\mathcal{D}_3$ only weakly pulls it in the opposite direction; $\mathcal{D}_1$, $\mathcal{D}_2$, and $\mathcal{D}_5$ are negligible for providing zero or minimal gains.
As a result, $N_{\mathcal{D}}^*$ is pulled toward the high-budget region near the large $\mathbb{E}_{\bm{x}\sim P_{\mathcal{D}_4}}[N_{\bm{x}}^*]$, while the remaining samples cannot pull it back.
This explains why many $N_{\bm{x}}^*$ concentrate in the low-budget region , while $N_{\mathcal{D}}^*$ shifts rightward (\textcolor{deepred}{Figure \ref{fig:distribution-qwen3-aime25}}), causing serve overscaling.
This intuition can be supported by the following theorem, with proof in \textcolor{deepred}{Appendix \ref{appe:proof}}.

\vspace{-0.08in}
\begin{theorem}
\label{thm:type4-dominance}
Let $\mu_i=\mathbb{E}_{\bm{x}\sim P_{\mathcal{D}_i}}[N_{\bm{x}}^*]$, where $\mu_4>\mu_i ~(i\neq 4)$.
With monotonicity features, We have $\bar{\Delta}_1 = \bar{\Delta}_2 = 0$ and $\bar{\Delta}_5 \approx 0$.
If $p_4\bar{\Delta}_4>p_3[-\bar{\Delta}_3]_+$, then the budget utilization $\mathbb{E}_{\bm{x}\sim P_{\mathcal{D}}}[\mathcal{U}_{\bm{x}}]<1$.
\end{theorem}
\vspace{-0.08in}

So far, we have completed the backward tracing of the closed-loop process, and found that
\textit{the key source of the overscaling curse lies in Type-4 samples: their large accuracy gains pull \(N_{\mathcal{D}}^*\) toward the high-budget region, while other samples cannot counteract this pull. Consequently, many low-\(N_{\bm{x}}^*\) samples suffer reduced budget utilization, leading to low \(\mathbb{E}_{\bm{x}\sim P_{\mathcal{D}}}[\mathcal{U}_{\bm{x}}]\) and thus the overscaling curse.}

%% file: 4-breaking.tex
\vspace{-0.05in}
\section{Breaking the Overscaling Curse}
\label{sec:break}

Having demonstrated the prevalence and severity of the overscaling curse, we now seek methods to break it.
Specifically, given a model $\pi$, for any dataset $\mathcal{D}$, we aim to {\it improve the sample efficiency ({\it i.e.}, budget utilization) while maintaining overall system efficacy ({\it i.e.}, dataset accuracy)}.

\vspace{-0.05in}
\subsection{Motivation}
\label{sec:break-motivation}

The overscaling curse is induced by enforcing a single global budget on all samples.
Therefore, a natural solution is to relax this constraint and allocate sample-specific budgets.
Since each sample has its own sample-optimal budget $N_{\bm{x}}^*$ (\textcolor{deepred}{Eq.\ref{eq:sample-optimal-budget}}), ideally, each sample should receive a budget close to $N_{\bm{x}}^*$.
In this case, $\mathcal{U}_{\bm{x}}$ approaches $1$, and the dataset accuracy is maintained or even improved, since $\mathbb{E}[\mathcal{A}_{\bm{x}}(N^*_{\bm{x}})] \geq \mathbb{E}[\mathcal{A}_{\bm{x}}(N_{\mathcal{D}})]$.
Thus, breaking the curse reduces to predicting $N_{\bm{x}}^*$ for each sample.

This raises the immediate question: what features should we use to predict $N^*_{\bm{x}}$?
Intuitively, for a given sample $\bm{x}$, a very strong or very weak model may only need a small budget, as it tends to be consistently right or wrong regardless of sampling. A moderately capable model, however, may require more sampling to reach a reliable decision.
Thus, $N_{\bm{x}}^*$ depends on both $\bm{x}$ and how the model $\pi$ perceives it.
This motivates using {\it model-aware features}: The hidden states of $\pi$ when encoding $\bm{x}$ naturally capture this information, making them suitable for predicting $N_{\bm{x}}^*$.
In contrast, methods that rely only on the raw sample, such as few-shot learning with external LLMs, may miss these signals.

\begin{figure*}[t]
    \vspace{-0.2in}
    \centering
    \includegraphics[width=1\textwidth]{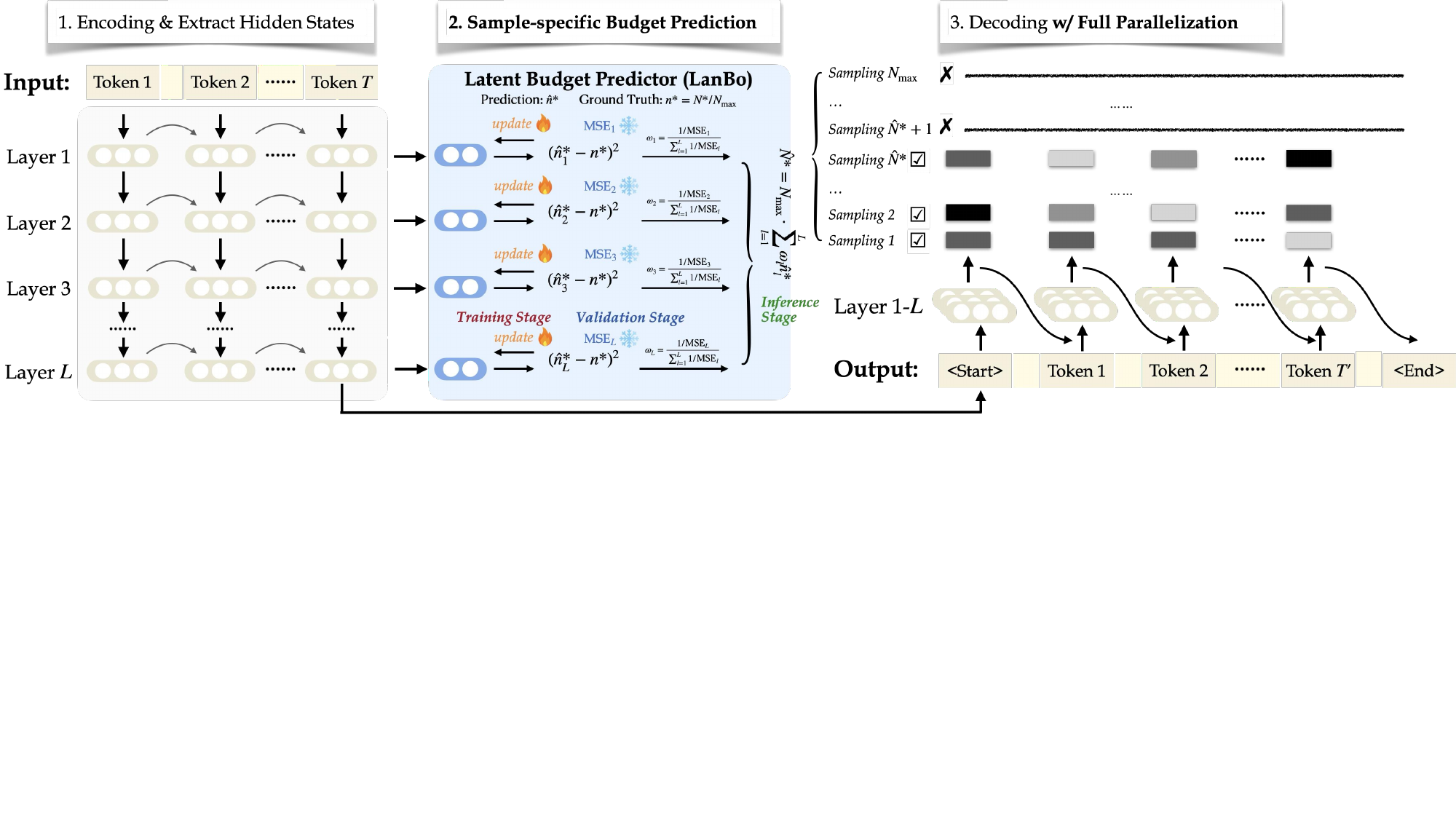}
    \vspace{-1.7in}
    \caption{
    Overview of {\bf our \latbp{} method (middle)} and the {\bf new efficient parallel decoding paradigm \tbt{} (whole)}, formed by integrating \latbp{} into the full decoding pipeline.
    }
    \vspace{-0.2in}
    \label{fig:pipeline}
\end{figure*}

\vspace{-0.05in}
\subsection{Method: Latent Budget Predictor (\latbp{}, \textcolor{deepred}{Figure \ref{fig:pipeline}})}
\label{sec:break-method}

With the above considerations, for a given model $\pi$, we train a lightweight budget predictor that learns to predict $N^*_{\bm{x}}$ from $\pi$'s hidden state of $\bm{x}$ through real-world samples.
Once trained, this predictor should generalize to unseen samples, producing budget predictions for them during inference.

Assume $\pi$ has $L$ decoder layers.
For an input $\bm{x} = (x_1, \dots, x_T)$ of length $T$, the token embeddings $\bm{H}^{0} \in \mathbb{R}^{T \times d}$ are processed layer by layer: $\bm{H}^{l} = F^{l}(\bm{H}^{l-1})$, where $d$ is the hidden dimension and $F^{l}$ is the $l$-th layer function.
Due to causal attention, the last-token representation $\bm{h}^{l}_T = \bm{H}^{l}[:,T]$ can serve as an information aggregation of the entire input at layer $l$ \citep{azaria2023internal,wang2024improving,dong2025emergent}.

\vspace{-0.1in}
\paragraph{Training: Layer-wise Learning.}

We first consider the training process.
Since we do not know in advance which layer's hidden state is most suitable as a learnable feature, we use all $\bm{h}^{l}_T$ from each layer and train $L$ layer-wise predictors $\{\phi^{\pi}_{\theta_l}: \mathbb{R}^d \rightarrow \mathbb{R}\}_{l=1}^L$.
Each predictor is defined as $\phi^{\pi}_{\theta_l}(\bm{h}^{l}_T(\bm{x})) = N_{\bm{x}}^*/N_{\max}$, where the target is normalized as $n_{\bm{x}}^* = N_{\bm{x}}^*/N_{\max} \in [0,1]$ to stabilize training.
Each $\phi^{\pi}_{\theta_l}$ is trained on the training data $\mathcal{D}_{\text{train}}$ to minimize the squared loss:

\vspace{-0.1in}
\begin{small}
\begin{equation}
\mathcal{L}^{\pi}_{l} = \mathbb{E}_{\bm{x}\sim P_{\mathcal{D}_{\text{train}}}} \left[ \left( \phi^{\pi}_{\theta_l}(\bm{h}^{l}_T(\bm{x})) - n_{\bm{x}}^* \right)^{2} \right]
, \quad \text{for } 1 \leq l \leq L,
\label{eq:training-objective}
\end{equation}
\end{small}
\vspace{-0.1in}

\vspace{-0.1in}
\paragraph{Inference: Layer-wise Combination.}
After training, given an unseen $\bm{x}$, we can obtain $L$ outputs from the $L$ learned predictors.
The remaining challenge is how to combine them into a final prediction.

Let $\hat n_{\bm{x},l}^*=\phi^{\pi}_{\theta_l}(\bm{h}_T^l(\bm{x}))$ denote the $l$-th predictor output. We obtain the final prediction by linearly combining all layer-wise outputs with weights $\omega_l$ at layer $l$:

\vspace{-0.1in}
\begin{small}
\begin{equation}
\hat N_{\bm{x}}^* = N_{\max} \cdot \bm \omega^\top \hat{\bm n}_{\bm{x}}^*,
\qquad \text{s.t. } \bm 1^\top \bm \omega = 1,
\label{eq:aggregation-general}
\end{equation}
\end{small}
\vspace{-0.15in}

where $\hat{\bm n}_{\bm{x}}^* = (\hat n_{\bm{x},1}^*, \dots, \hat n_{\bm{x},L}^*)^\top$ and $\bm{\omega} = (\omega_1, \dots, \omega_L)^\top$.
Although the ideal weights are not directly available, our inference objective is clear: {\it minimize the expected error of the final prediction on unseen samples}.
Thus, the optimal weights can be solved by the following optimization problem

\vspace{-0.1in}
\begin{small}
\begin{equation}
\bm \omega^*
=
\arg\min_{\bm \omega}\;
\mathbb{E}_{\bm{x}\sim P_{\mathcal D_{\text{unseen}}}}
\left[
\left(
\bm \omega^\top \hat{\bm n}_{\bm{x}}^* - n_{\bm{x}}^*
\right)^2
\right]
\qquad
\text{s.t. } \bm 1^\top \bm \omega = 1.
\label{eq:weight-opt}
\end{equation}
\end{small}
\vspace{-0.1in}

The solution to this optimization problem is $\bm \omega^* = \frac{\Sigma^{-1} \bm 1}{\bm 1^\top \Sigma^{-1} \bm 1}$, where $\Sigma = \mathbb{E}_{\bm{x} \sim P_{\mathcal{D}}} [(\hat{\bm n}_{\bm{x}}^* - n_{\bm{x}}^* \bm 1)(\hat{\bm n}_{\bm{x}}^* - n_{\bm{x}}^* \bm 1)^\top]$ is the error covariance matrix across layers. The derivation is provided in \textcolor{deepred}{Appendix \ref{appe:LBA-weighting}}.
In principle, computing $\bm \omega^*$ requires the full covariance matrix $\Sigma$, but estimating and inverting it can be unstable in practice. We thus adopt a diagonal surrogate $\Sigma \approx \mathrm{diag}(m_1, \dots, m_L)$, where $m_l = \mathbb{E}_{\bm{x} \sim P_{\mathcal{D}_{\text{unseen}}}} [ ( \hat n_{\bm{x},l}^* - n_{\bm{x}}^* )^2 ]$ is the second moment of the prediction error of the $l$-th predictor.
The approximation error is analyzed in \textcolor{deepred}{Appendix \ref{appe:LBA-weighting}}, which is ignorable in practice.

Under this approximation, the optimal weight vector reduces to inverse-error weighting: $\bm \omega \propto \bm m^{-1}$, where $\bm m = (m_1, \dots, m_L)^\top$. Since the true distribution is unknown, we introduce a \textbf{\textit{Validation Set $\mathcal{D}_{\text{val}}$}} and estimate $m_l$ using the empirical MSE $\hat m_l$ of $l$-th predictor.
The final weights are then:

\vspace{-0.1in}
\begin{small}
\begin{equation}
\bm \omega = \hat{\bm m}^{-1} / \|\hat{\bm m}^{-1}\|_1, \quad \text{where } \hat{\bm m}^{-1} = (1/\hat m_1, \dots, 1/\hat m_L)^\top.
\label{eq:final-weight}
\end{equation}
\end{small}
\vspace{-0.1in}

Applying this to \textcolor{deepred}{Eq.\ref{eq:aggregation-general}} yields the final prediction $\hat N^*_{\bm{x}}$.
This scheme actually assigns higher weights to predictors with lower squared error, aligning with the intuition that better should contribute more.

\begin{figure}[t]
\vspace{-0.2in}
    \centering
    \begin{minipage}[b]{0.48\columnwidth}
        \centering
        \includegraphics[width=\linewidth]{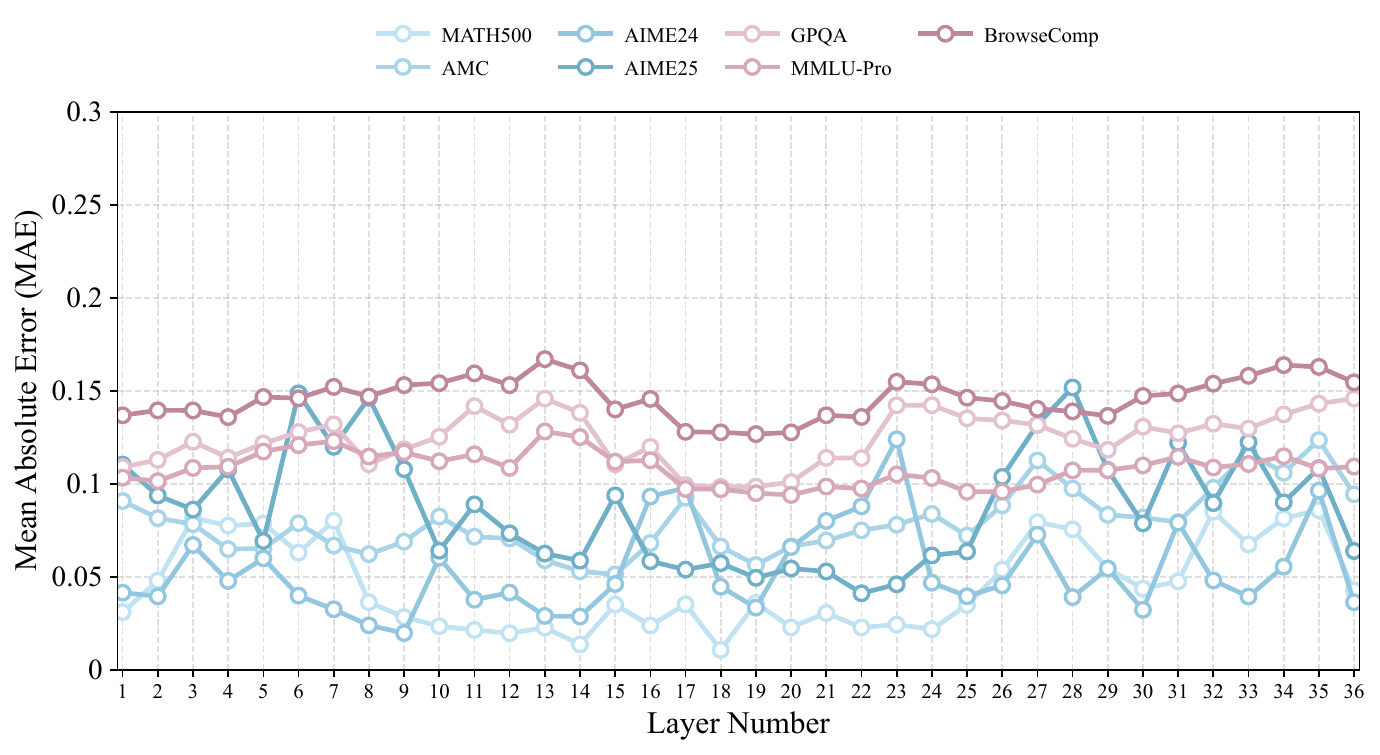}
        \vspace{-0.2in}
        \subcaption{MAE results of the $L$ layer-wise predictors on the validation set, with each averaged {\it over 8 runs}.}
    \end{minipage}
    \hfill
    \begin{minipage}[b]{0.48\columnwidth}
        \centering
        \includegraphics[width=\linewidth]{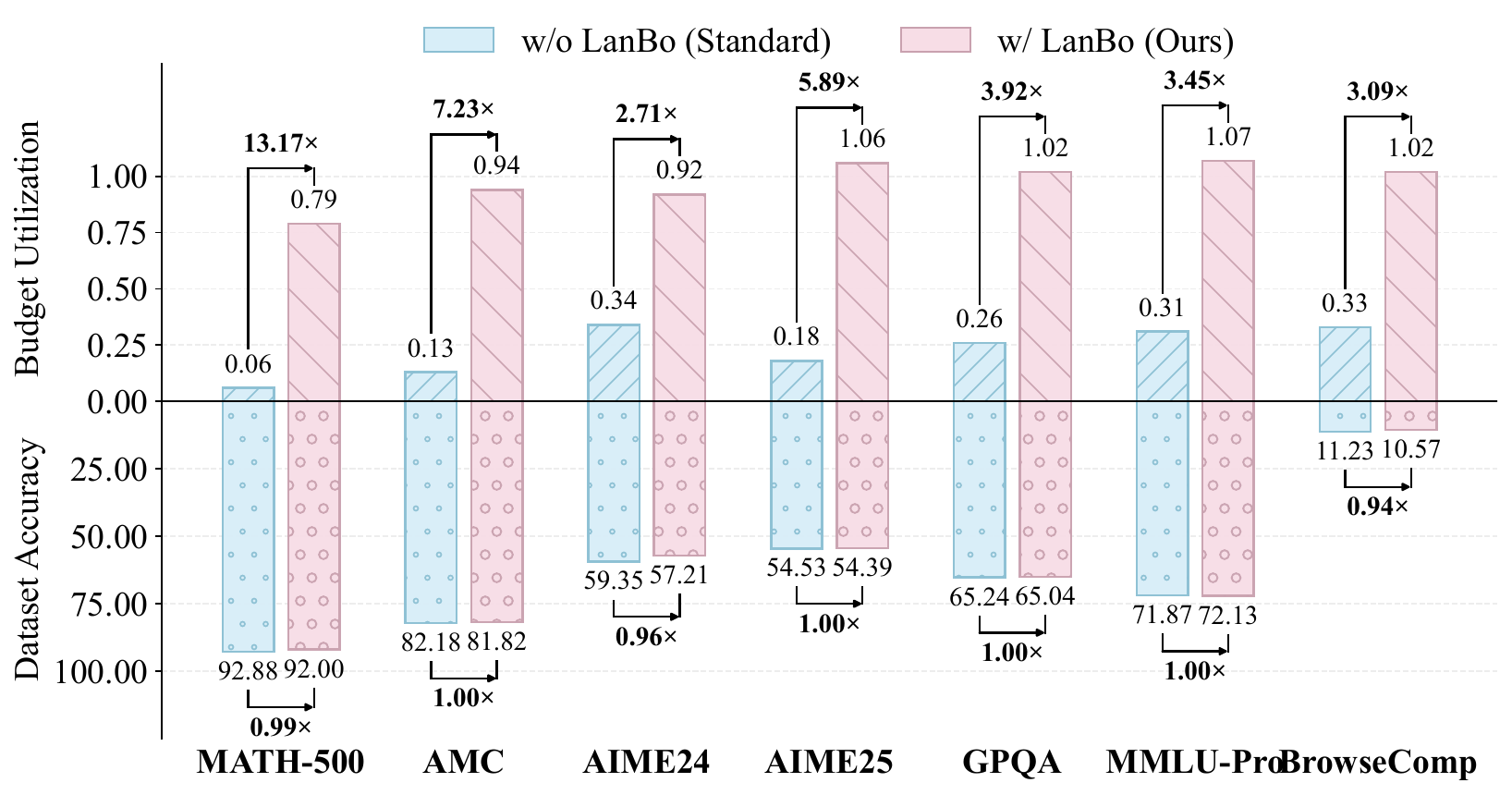}
        \vspace{-0.2in}
        \subcaption{Comparison of expected budget utilization and dataset accuracy before and after applying \latbp.}
    \end{minipage}
    \vspace{-0.05in}
    \caption{{\bf Empirical Validation of \latbp~for Breaking the Overscaling Curse}. Here we use \texttt{Qwen3-4B} with {\it voting} aggregation, with other results presented in  \textcolor{deepred}{Figures \ref{fig:latbp-results-qwen25} -- \ref{fig:latbp-results-qwen3-model}} in \textcolor{deepred}{Appendix \ref{appe:latbp}}.}
    \vspace{-0.25in}
    \label{fig:latbp-results-qwen3}
\end{figure}

\vspace{-0.05in}
\subsection{General Implementation}
\label{sec:predictor-implementation}

Having described the general training and inference procedures, we now specify the implementation.
For the architecture, we follow the principle of minimal algebraic structure to employ a lightweight network: a single-hidden-layer MLP with a ReLU activation, followed by a logistic regression output layer to adapt the regression objective.
We have also tested more complex architectures like deeper MLPs and shallow Transformers, but found no clear performance gains. We therefore adopt the simplest architecture that provides effective learning with minimal cost and no overfitting.
Let $d$ be the hidden dimension of the language model, we set the hidden size of MLP as $\lfloor rd \rfloor$ with $r = 1/8$.

For training data\footnote{Our training data are from open-source corpus and contain no samples in our evaluation sets. We also do not filter by domain to match our evaluation domains, ensuring reliable subsequent conclusions about domain adaptability.}, we consider two main dimensions: {\it difficulty diversity} and {\it domain diversity}.
The former enables the predictors to learn how to allocate budgets to problems of varying difficulty, which is the primary learning objective.
The latter enhances their adaptability across multiple domains, facilitating generalization to unseen scenarios.
The validation data is drawn from the same source as the training data.
Detailed data specifications and training procedures are provided in \textcolor{deepred}{Appendix \ref{appe:LBA-implementation}}.

\vspace{-0.05in}
\subsection{Empirical Validation}
\label{sec:break-experiments}

\paragraph{Evaluation Metric.}

We first check whether the \(L\) predictors are well learned by measuring their Mean Absolute Error (MAE) on the test sets.
Next, we formally evaluate whether \latbp{} breaks the overscaling curse.
Specifically, We replace the global budget with the sample-specific prediction of the actual budget $N_{\bm{x}}$ in the budget utilization $\mathcal{U}_{\bm{x}}$ (\textcolor{deepred}{Eq.\ref{eq:sample-budget-utilization}}), and compare the expected budget utilization \(\mathbb{E}_{\bm{x}\sim P_{\mathcal{D}}}[\mathcal{U}_{\bm{x}}]\) before and after the replacement, along with the corresponding dataset accuracy.

\vspace{-0.1in}
\paragraph{Main Results (\textcolor{deepred}{Figure \ref{fig:latbp-results-qwen3}}).}

\textcolor{deepred}{\ref{fig:latbp-results-qwen3}(a)} shows the learning results.
MAEs on the four mathematical datasets are lowest (mostly below 0.1).
On the other three domains, although MAEs are slightly higher than those on mathematical ones, they remain around 0.15, still indicating good effectiveness.
Across layers, intermediate layers are more stable than early or late ones, which exhibit larger fluctuations, suggesting that intermediate representations offer more robust features, consistent with \citep{skean2025layer}.
\textbf{\textit{Overall, \latbp ~demonstrates strong learning effectiveness and good domain adaptability.}}

\textcolor{deepred}{\ref{fig:latbp-results-qwen3}(b)} presents the inference results.
Clearly, after applying our \latbp, the budget utilization increases significantly, hovering around 1, demonstrating that sample efficiency has been greatly improved. In terms of performance, there is almost no significant degradation; in fact, performance even improves on certain datasets, satisfying the condition of maintaining system efficacy.
Meanwhile, this phenomenon holds across all datasets, indicating that our method is effective across various domains.
\textbf{\textit{Overall, \latbp ~effectively breaks the overscaling curse and adapts well to diverse domains.}}

%% file: 5-application.tex
\vspace{-0.05in}
\section{Practical Application: Real-world Efficient Parallel Decoding}
\label{sec:application}

In the previous section, we revealed the overscaling curse and introduced \latbp{} as a solution.
While \latbp{} improves efficiency in terms of budget utilization, it must ultimately be deployed in real-world decoding pipelines to demonstrate practical usability.
In practice, efficiency depends not only on simple budget size but also on hardware-aware costs such as inference latency and memory overhead \citep{sun2024fast,wang2025sampling}.
Therefore, in this section, we further integrate \latbp{} into the full parallel decoding pipeline and evaluate whether it brings practical efficiency gains beyond budget reduction.

\vspace{-0.05in}
\subsection{New Paradigm: Pre-decoding Budget Adaptation (\tbt)}
\label{sec:paradigm}

When integrating \latbp ~into the full decoding pipeline, the resulting pipeline is illustrated in \textcolor{deepred}{Figure \ref{fig:pipeline}}, with the following three steps:
(1) The model $\pi$ encodes input $\bm{x}$ and extracts hidden states $\bm{h}_T^{l}(\bm{x})$ from all layers;
(2) The well-trained $L$ predictors produce a budget prediction $\hat{N}_{\bm{x}}^*$;
(3) $\hat{N}_{\bm{x}}^*$ sequences are sampled with full hardware parallelization and aggregated to the final answer.

\vspace{-0.1in}
\paragraph{Highlighted Comparison.}

\begin{wraptable}{r}{0.42\textwidth}
\vspace{-0.2in}
\centering
\footnotesize
\setlength{\tabcolsep}{1.1mm}
\renewcommand{\arraystretch}{1.0}

\caption{Extra blocking comparisons of two efficient parallel decoding paradigms.}
\vspace{-0.05in}
\resizebox{0.42\textwidth}{!}{
\begin{tabular}{lccc}
\toprule
& \multirow{2}{*}{\textbf{\makecell{Extra\\ Training}}} & \multicolumn{2}{c}{\textbf{Hardware Blocking}} \\
\cline{3-4}
& & \textbf{Before Decoding} & \textbf{During Decoding} \\
\midrule
\textbf{\tdt} & \ding{55} & \ding{55} & \ding{51} \\
\textbf{\tbt{} (Ours)} & \ding{51} & \ding{51} & \ding{55} \\
\bottomrule
\end{tabular}
}
\vspace{-0.15in}
\label{tab:comparison}
\end{wraptable}

The key feature of our pipeline is that budget allocation is performed \emph{before decoding}.
Existing efficient parallel decoding methods also pursue sample-specific budgets, but typically adapt them \emph{during decoding}.
They often split the original large parallel batch into smaller mini-batches, sometimes of size 1, and decide after each mini-batch whether to stop or continue using internal confidence signals \citep{li2024escape,fu2025deep} or external model signals \citep{wang2025make}.
Based on this distinction, \latbp{} naturally inspires a new efficient parallel decoding paradigm, \textbf{Pre-decoding Budget Adaptation (\tbt)}.
Correspondingly, we refer to existing methods as the {\it Decoding-time Budget Adaptation (\tdt)} paradigm.
\textbf{\textit{Both paradigms aim at adaptive budgeting, but differ in when the budget is determined.}}

Despite these differences, both paradigms introduce extra blocking costs to the decoding pipeline when reducing budgets.
As in \textcolor{deepred}{Table \ref{tab:comparison}}, \tdt{} blocks full hardware parallelization during decoding, while \tbt{} incurs cost for predictor training and inference before decoding.
Therefore, comparing methods under the two paradigms with hardware-aware efficiency can truly reveal the practical value of our \latbp{}, and demonstrate the promise of the new \tbt{} paradigm it represents.

\vspace{-0.05in}
\subsection{Empirical Validation}
\label{sec:efficient-experiments}

\paragraph{Baseline.}
Considering the extra cost of verifiers, we use majority voting as the aggregation strategy \citep{wang2022self}.
It is lightweight and is widely used in parallel decoding.
The standard parallel decoding baseline ({\it STD.}) therefore also uses majority voting, with its global budget set to $N^*_{\mathcal{D}}$ to maximize dataset accuracy.
For the \tdt{} paradigm, we evaluate four representative methods: ASC \citep{aggarwal2023let}, ESC \citep{li2024escape}, DSC \citep{wang2025make}, and the online version of DeepConf \citep{fu2025deep}.
Details are provided in \textcolor{deepred}{Appendix \ref{appe:baselines}}.
The predictors in our \textsc{Latbp} is implemented consistently with \textcolor{deepred}{Section \ref{sec:predictor-implementation}}.

\vspace{-0.1in}
\paragraph{Evaluation Metric.}

When evaluating in a system $(\pi, \mathcal{D})$, we still first evaluate the dataset accuracy.
For efficiency, we use two hardware-aware metrics: {\it memory overhead} and {\it inference latency} \citep{wang2025sampling}:

\vspace{-0.1in}
\begin{itemize}[leftmargin=0.6cm]
    \item For memory, we measure the peak usage per sample $\bm{x}$, which reflects the required hardware capacity. This usage is dominated by the KV cache and model weights, {\it plus any extra overhead from extra models}.
    We compute the expected memory peak usage $\mathcal{M}_{\text{met}}$ per sample over $\mathcal{D}$.
    \item For latency, we measure wall-clock time per sample $\bm{x}$ from encoding to answer aggregation, including all intermediate steps. We compute the total latency $\mathcal{T}_{\text{met}}$ over $\mathcal{D}$. {\it For our \latbp, we include training time in each system evaluation to ensure fair comparison.} 
\end{itemize}
\vspace{-0.1in}

We compare each method against {\it STD.} using relative ratios $\delta_{\mathcal{T}} = \mathcal{T}_{\text{met}} / \mathcal{T}_{\text{std.}}$ and $\delta_{\mathcal{M}} = \mathcal{M}_{\text{met}} / \mathcal{M}_{\text{std.}}$.
Smaller $\delta$ indicates better optimization, and $\delta$ above 1 implies no real improvement over {\it STD.}

\input{tables/qwen3-4B-results}

\vspace{-0.1in}
\paragraph{Main Results (\textcolor{deepred}{Table \ref{tab:qwen3-4B-results}} and \textcolor{deepred}{\ref{tab:qwen25-7B-results} -- \ref{tab:r1-results}}).}

In terms of efficiency, \latbp{} achieves over 50\% memory savings in most settings, ranks among the top two in over 90\% of cases, and remains comparable to other methods.
More importantly, its inference latency is significantly lower than that of the others.
The key reason is that {\it decoding-time parallelization blocking is more costly than expected}.
With strong parallel acceleration such as \texttt{vLLM}, a single parallel request with $N=128$ incurs only about 2--3$\times$ the latency of $N=1$, and this gap shrinks further for longer reasoning.
Therefore, \emph{for \tdt{} methods, serializing parallel requests can hurt latency more than budget reduction helps}, sometimes even making them slower than \emph{STD.}
In contrast, \emph{keeping decoding fully parallelized is key to \latbp{}'s low latency}.
Meanwhile, the pre-decoding predictor is lightweight, adding only milliseconds of inference overhead.
Even including training time, the total time is only tens of seconds, remaining modest compared with the total decoding time.
See \textcolor{deepred}{Appendix \ref{appe:latency}} for detailed latency statistics.
Overall, \textbf{\textit{\latbp{} enables truly efficient parallel decoding in real-world settings.}}

In terms of accuracy, \latbp{} maintains stable performance compared with {\it STD.}, with overall fluctuations within $2\%$, and even outperforms {\it STD.} in about $30\%$ of cases.
Compared with \tdt{} methods, \latbp{} consistently achieves the best or second-best accuracy.
This stability arises from directly learning optimal budget features, thereby avoiding unstable heuristic strategies used by \tdt{} methods.
For example, DSC relies on an extra model to rank problem difficulty, while DeepConf uses model confidence; both perform well on \texttt{R1} but are less effective on \texttt{Llama3.1}.
Thus, \textbf{\textit{\latbp{} maintains stable accuracy, satisfying the prerequisite for efficient parallel decoding.}}

Furthermore, \textbf{\textit{we emphasize that \latbp ~exhibits strong domain adaptability.}}
Even though the predictor for each model is trained only once, it performs consistently well across diverse domains. This further demonstrates the completeness of our training and enhances its practical value.

%% file: tables/qwen3-4B-results.tex
\begin{table}[t]
\vspace{-0.2in}
\caption{{\bf Empirical Validation of Our \latbp ~for Efficient Parallel Decoding.} Under each system, we report the real-world efficiency (memory $\delta_{\mathcal{M}}$ and latency $\delta_{\mathcal{T}}$, both the ratio relative to \textit{STD.}) and accuracy ({\it Acc.}), where \textit{STD.} denotes the standard parallel decoding. {\bf Bold} and \underline{underline} denote the best and second best, respectively. \textcolor{blue}{Blue} means higher accuracy than \textit{STD.}; \textcolor{red}{red} means lower efficiency.
Each result is averaged \textit{\textbf{over 32 runs}}, with standard deviations in \textcolor{deepred}{Appendix \ref{appe:std}}.
This table shows \texttt{Qwen3-4B} results. See \textcolor{deepred}{Table \ref{tab:qwen25-7B-results} -- \ref{tab:r1-results}} in \textcolor{deepred}{Appendix \ref{appe:main-results}} for results in other three models.
}
\vspace{0.05in}
\centering
\footnotesize
\renewcommand\arraystretch{1.}
\setlength{\tabcolsep}{1.3mm}{
\resizebox{1\textwidth}{!}{
\begin{tabular}{llccccccccccccccccccccc}

\toprule

\multirow{2}{*}{\bf Paradigm}
& \multirow{2}{*}{\bf Method}
& \multicolumn{3}{c}{\bf MATH500}
& \multicolumn{3}{c}{\bf AMC}
& \multicolumn{3}{c}{\bf AIME24}
& \multicolumn{3}{c}{\bf AIME25}
& \multicolumn{3}{c}{\bf GPQA}
& \multicolumn{3}{c}{\bf MMLU-Pro}
& \multicolumn{3}{c}{\bf BrowseComp}
\\
\cmidrule(lr){3-5}
\cmidrule(lr){6-8}
\cmidrule(lr){9-11}
\cmidrule(lr){12-14}
\cmidrule(lr){15-17}
\cmidrule(lr){18-20}
\cmidrule(lr){21-23}

& &
$\delta_{\mathcal{M}} \downarrow$ & $\delta_{\mathcal{T}} \downarrow$ & {\it Acc.} $\uparrow$ & 
$\delta_{\mathcal{M}} \downarrow$ & $\delta_{\mathcal{T}} \downarrow$ & {\it Acc.} $\uparrow$ & 
$\delta_{\mathcal{M}} \downarrow$ & $\delta_{\mathcal{T}} \downarrow$ & {\it Acc.} $\uparrow$ & 
$\delta_{\mathcal{M}} \downarrow$ & $\delta_{\mathcal{T}} \downarrow$ & {\it Acc.} $\uparrow$ & 
$\delta_{\mathcal{M}} \downarrow$ & $\delta_{\mathcal{T}} \downarrow$ & {\it Acc.} $\uparrow$ & 
$\delta_{\mathcal{M}} \downarrow$ & $\delta_{\mathcal{T}} \downarrow$ & {\it Acc.} $\uparrow$ &
$\delta_{\mathcal{M}} \downarrow$ & $\delta_{\mathcal{T}} \downarrow$ & {\it Acc.} $\uparrow$ \\

\midrule

--- & {\it STD.}
& {\it 1.00} & {\it 1.00} & {\it 92.88}
& {\it 1.00} & {\it 1.00} & {\it 82.18}
& {\it 1.00} & {\it 1.00} & {\it 59.35}
& {\it 1.00} & {\it 1.00} & {\it 54.53}
& {\it 1.00} & {\it 1.00} & {\it 65.24}
& {\it 1.00} & {\it 1.00} & {\it 71.87}
& {\it 1.00} & {\it 1.00} & {\it 11.23}
\\

\hdashline
\multirow{4}{*}{\tdt}
& AC 
& 0.18 & \textcolor{red}{1.56} & {\bf 92.17}
& \underline{0.14} & \textcolor{red}{2.67} & 81.73
& 0.84 & \textcolor{red}{3.12} & 57.81
& 0.26 & \textcolor{red}{2.41} & 53.81
& {\bf 0.21} & \textcolor{red}{2.11} & 65.03
& \underline{0.30} & \textcolor{red}{2.36} & 71.72
& {\bf 0.26} & \underline{\textcolor{red}{1.90}} & \underline{10.51}
\\
& ESC 
& {\bf 0.12} & \underline{0.88} & \underline{92.11}
& 0.23 & \textcolor{red}{1.84} & \underline{81.77}
& {\bf 0.55} & \underline{\textcolor{red}{1.93}} & 57.34
& 0.28 & \underline{\textcolor{red}{2.25}} & \underline{53.86}
& 0.32 & \underline{\textcolor{red}{1.97}} & \underline{65.12}
& 0.34 & \underline{\textcolor{red}{2.23}} & 71.50
& 0.34 & \textcolor{red}{2.34} & 9.80
\\
& DSC 
& \underline{0.16} & \textcolor{red}{1.35} & 91.17
& 0.28 & \underline{\textcolor{red}{1.71}} & 81.32
& 0.72 & \textcolor{red}{2.21} & 57.18
& \underline{0.22} & \textcolor{red}{2.78} & 50.23
& 0.36 & \textcolor{red}{3.92} & 64.21
& 0.39 & \textcolor{red}{2.97} & 70.38
& 0.45 & \textcolor{red}{2.67} & 10.00
\\
& DeepConf 
& 0.29 & \textcolor{red}{2.77} & 91.87
& 0.32 & \textcolor{red}{3.27} & 81.65
& \textcolor{red}{1.19} & \textcolor{red}{4.63} & {\bf 59.30}
& 0.24 & \textcolor{red}{3.76} & 52.97
& 0.49 & \textcolor{red}{3.20} & 64.87
& 0.58 & \textcolor{red}{4.99} & \underline{\textcolor{blue}{72.15}}
& 0.62 & \textcolor{red}{4.23} & 10.11
\\

\hdashline
\rowcolor{gray!15}
{\bf \tbt} & {\bf \latbp ~(Ours)}
& 0.20 & {\bf 0.45} & 92.07
& {\bf 0.10} & {\bf 0.52} & {\bf 81.88}
& \underline{0.57} & {\bf 0.79} & \underline{59.29}
& {\bf 0.17} & {\bf 0.56} & \textcolor{blue}{\bf 54.55}
& \underline{0.27} & {\bf 0.46} & {\bf 65.17}
& {\bf 0.22} & {\bf 0.40} & {\bf \textcolor{blue}{72.20}}
& \underline{0.28} & {\bf 0.48} & {\bf 10.63}
\\

\bottomrule

\end{tabular}%
}}
\label{tab:qwen3-4B-results}%
\vspace{-0.2in}
\end{table}

%% file: 6-analysis.tex
\vspace{-0.05in}
\section{Extended Analysis}
\label{sec:analysis}

\paragraph{Sensitivity to Sampling Hyperparameters.}
In the main experiments, we use a fixed set of sampling hyperparameters (\textcolor{deepred}{Section \ref{sec:parallel-thinking}}).
We now evaluate the sensitivity of \latbp{} to hyperparameter variations under two settings:
\emph{(1) In-Domain (ID) Adaptability}, where training and decoding use the same changed configuration; and
\emph{(2) Out-of-Domain (OOD) Generalization}, where the predictor is trained with one configuration but another is used for decoding.
Results are in \textcolor{deepred}{Appendix \ref{appe:sampling-generalization}}.

\vspace{-0.1in}
\paragraph{Ablation Study of \latbp{}.}
We perform three ablations:
\emph{(1) MLP hidden size}, \emph{(2) training data size}, and \emph{(3) inference weighting strategy}.
Although the third is derived theoretically, we compare it with common alternatives to validate its practical soundness.
Results are shown in \textcolor{deepred}{Appendix \ref{appe:LBA-ablation}}.

\vspace{-0.1in}
\paragraph{Design Principles of \latbp{}.}
\latbp{} also involves two implicit design principles:
\emph{(1) training objective}, where we use continuous regression rather than discrete classification; and
\emph{(2) layer-wise pattern}, where we train layer-wise $L$ predictors and combine them at inference, rather than concatenating hidden states from all $L$ layers into a single predictor.
We discuss the rationale behind these choices and empirically verify their rationality in \textcolor{deepred}{Appendix \ref{appe:LBA-principle}}.

%% file: 7-conclusion.tex
\vspace{-0.05in}
\section{Conclusion and Future Work}

This paper reveals the overscaling curse in parallel thinking.
When a global budget is selected to maximize system efficacy, it can substantially reduce sample budget utilization, thereby undermining sample efficiency.
We empirically show that this contradiction is both prevalent and severe in real-world systems.
To break the overscaling curse, we propose \latbp{}, which probes the model's latent representations of each sample to learn and predict its sample-optimal budget.
\latbp{} significantly improves budget utilization while maintaining dataset accuracy.
Finally, we integrate \latbp{} into the full decoding pipeline, inspiring a new efficient decoding paradigm, \tbt{},
which substantially reduces latency and memory usage, demonstrating its practical usability.
More {\bf related work}, {\bf limitation statements}, and {\bf broader impacts}, are presented in \textcolor{deepred}{Appendix \ref{sec:related-work}}, \ref{sec:limitations}, \ref{sec:broader-impacts}.

Looking ahead from this work, our proposed \tbt{} is a promising paradigm for efficient parallel decoding, as it avoids decoding-time blocking and preserves full hardware parallelization.
In this work, \latbp{} serves as a first instantiation of this paradigm.
Although effective, it still leaves substantial room for further improvement.
For example, future work can explore training-free budget prediction methods, thereby further reducing overhead and improving efficiency.

%% file: appendix.tex
\newpage

\input{8A-overscaling-illustration}

\input{8B-overscaling-quantification}
\input{8C-overscaling-analysis}

\input{8D-LanBo-setting}


\input{8E-LanBo-results}

\input{8F-PreAda-setting}

\input{8G-PreAda-results}


\input{8H-analysis}

\input{8I-related-work}

%% file: 8A-overscaling-illustration.tex
\section{Illustration of Overscaling Curse}
\label{appe:illustration}

\begin{figure}[H]
    \centering
    \includegraphics[width=\columnwidth]{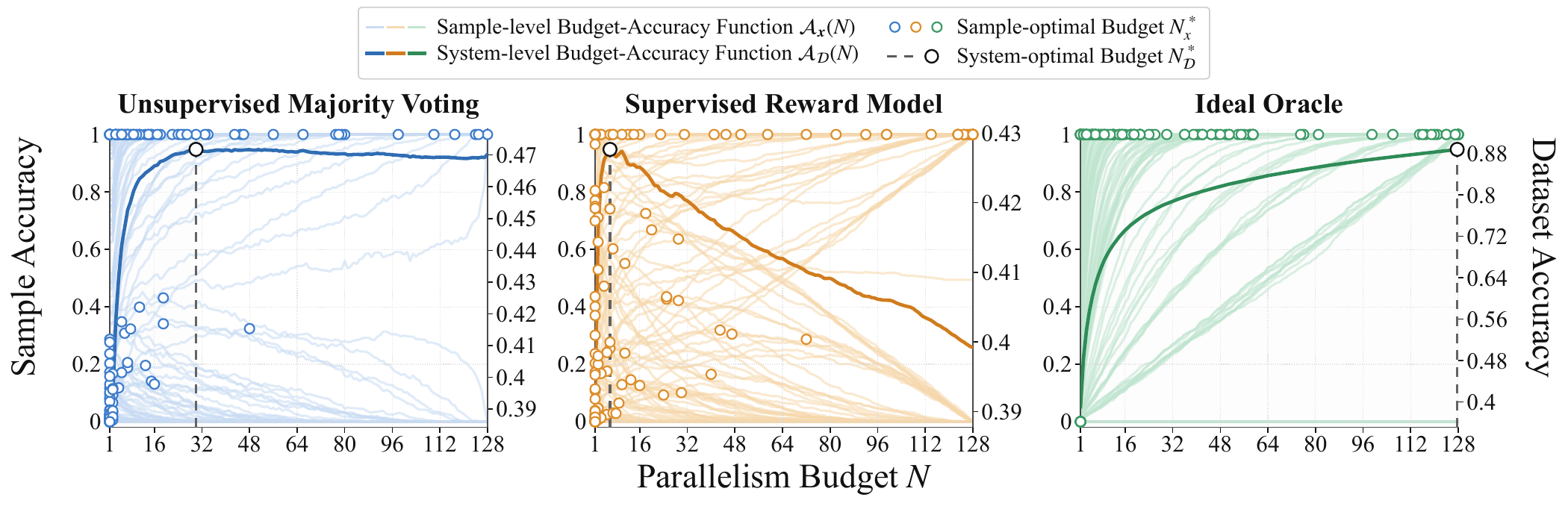}
    \vspace{-0.2in}
    \caption{{\bf Illustration of the Overscaling Curse in Parallel Thinking}. Each subfigure corresponds to a system $(\pi, \mathcal{D})$: evaluating model $\pi$ on dataset $\mathcal{D}$. Here $\pi =$ \texttt{Qwen2.5-7B} and $\mathcal{D} = \texttt{AMC}$.}.
    \label{fig:illustration-qwen25-amc}
\end{figure}

\begin{figure}[H]
    \centering
    \includegraphics[width=\columnwidth]{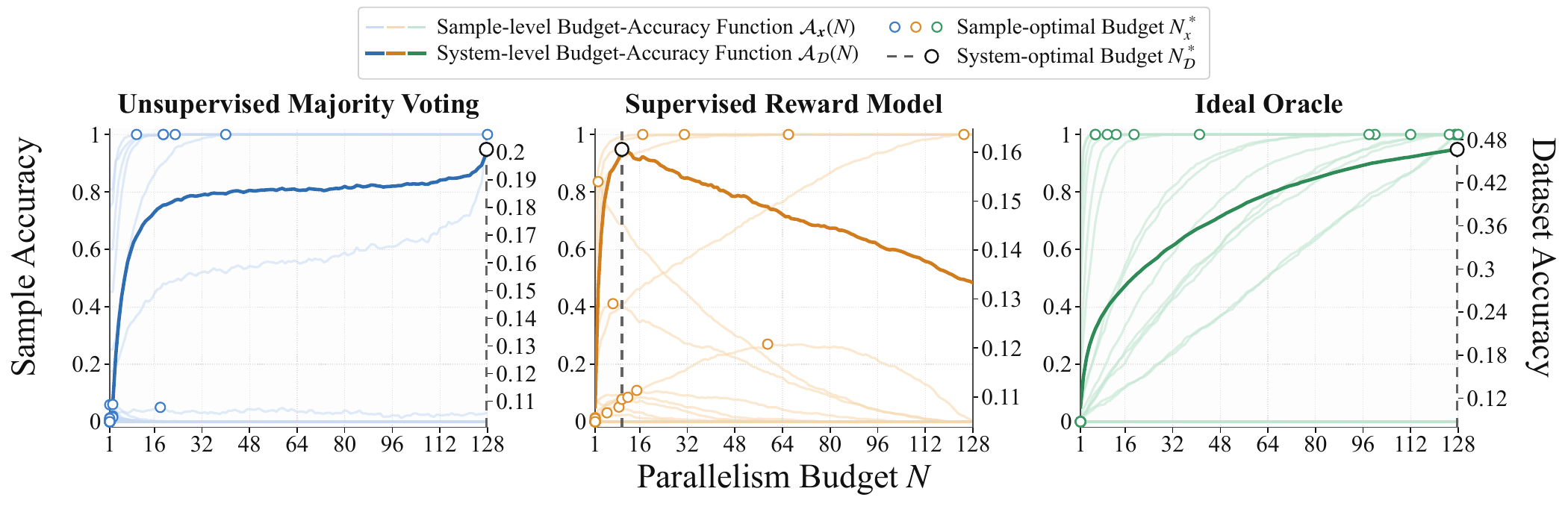}
    \vspace{-0.2in}
    \caption{{\bf Illustration of the Overscaling Curse in Parallel Thinking}. Each subfigure corresponds to a system $(\pi, \mathcal{D})$: evaluating model $\pi$ on dataset $\mathcal{D}$. Here $\pi =$ \texttt{Qwen2.5-7B} and $\mathcal{D} = \texttt{AIME24}$.}.
    \label{fig:illustration-qwen25-aime24}
\end{figure}

\begin{figure}[H]
    \centering
    \includegraphics[width=\columnwidth]{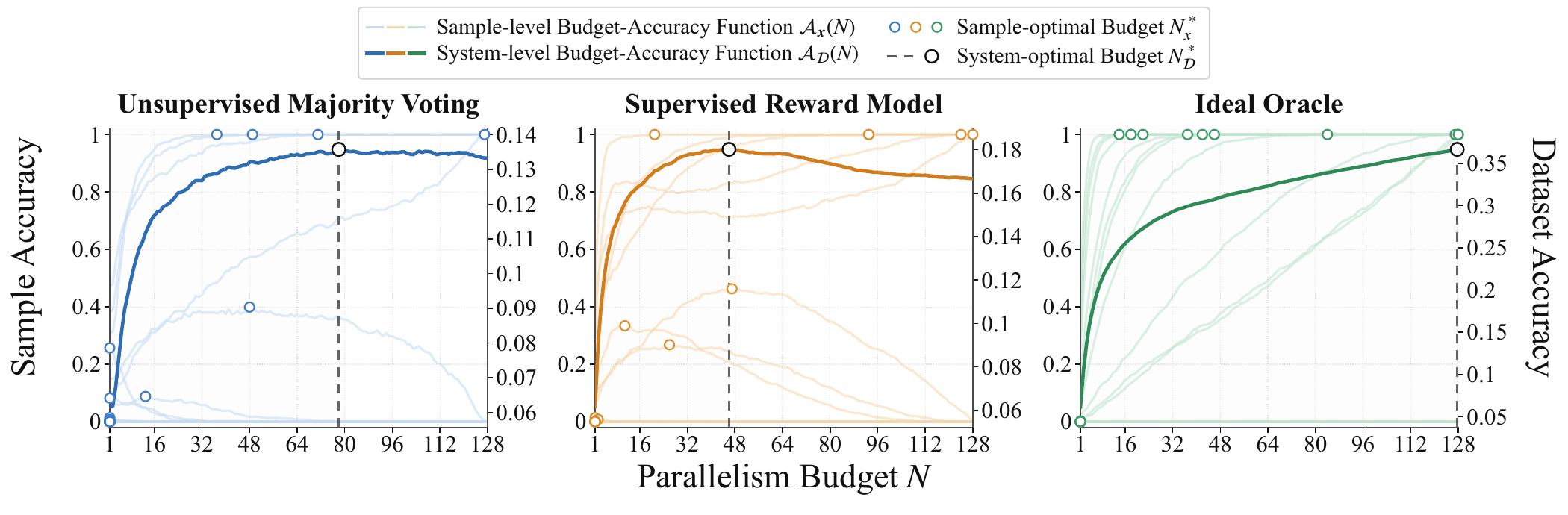}
    \vspace{-0.2in}
    \caption{{\bf Illustration of the Overscaling Curse in Parallel Thinking}. Each subfigure corresponds to a system $(\pi, \mathcal{D})$: evaluating model $\pi$ on dataset $\mathcal{D}$. Here $\pi =$ \texttt{Qwen2.5-7B} and $\mathcal{D} = \texttt{AIME25}$.}.
    \label{fig:illustration-qwen25-aime25}
\end{figure}

\begin{figure}[H]
    \centering
    \includegraphics[width=\columnwidth]{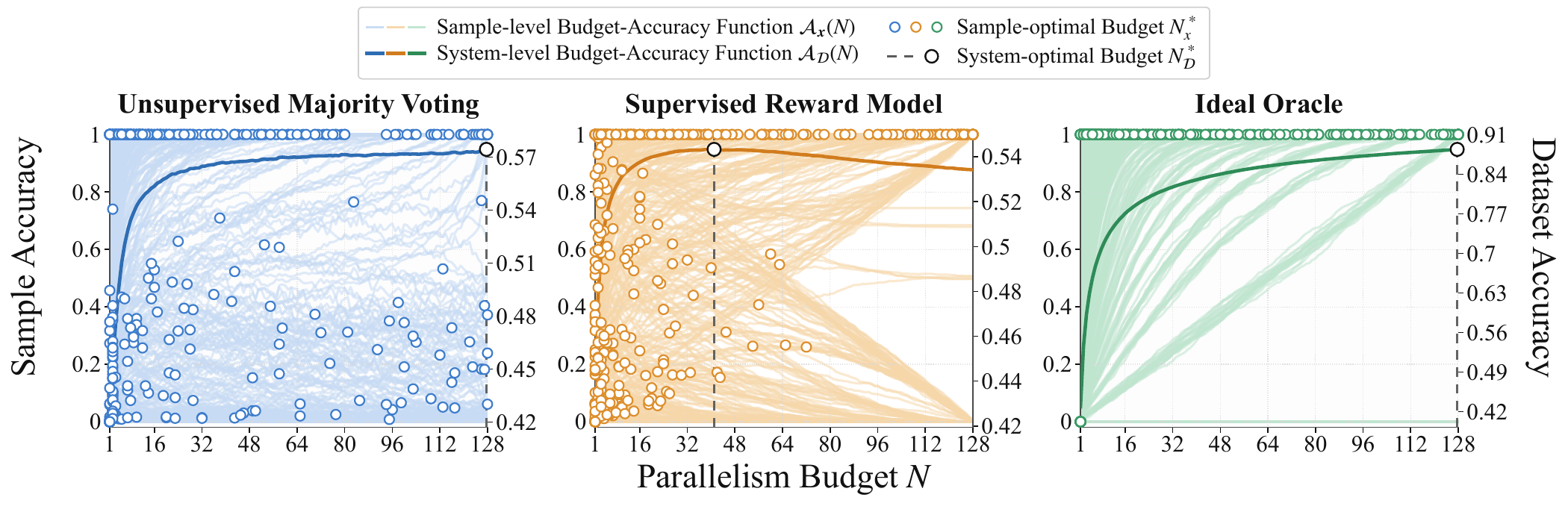}
    \vspace{-0.2in}
    \caption{{\bf Illustration of the Overscaling Curse in Parallel Thinking}. Each subfigure corresponds to a system $(\pi, \mathcal{D})$: evaluating model $\pi$ on dataset $\mathcal{D}$. Here $\pi =$ \texttt{Llama3.1-8B} and $\mathcal{D} = \texttt{MATH-500}$.}.
    \label{fig:illustration-llama31-math}
\end{figure}

\begin{figure}[H]
    \centering
    \includegraphics[width=\columnwidth]{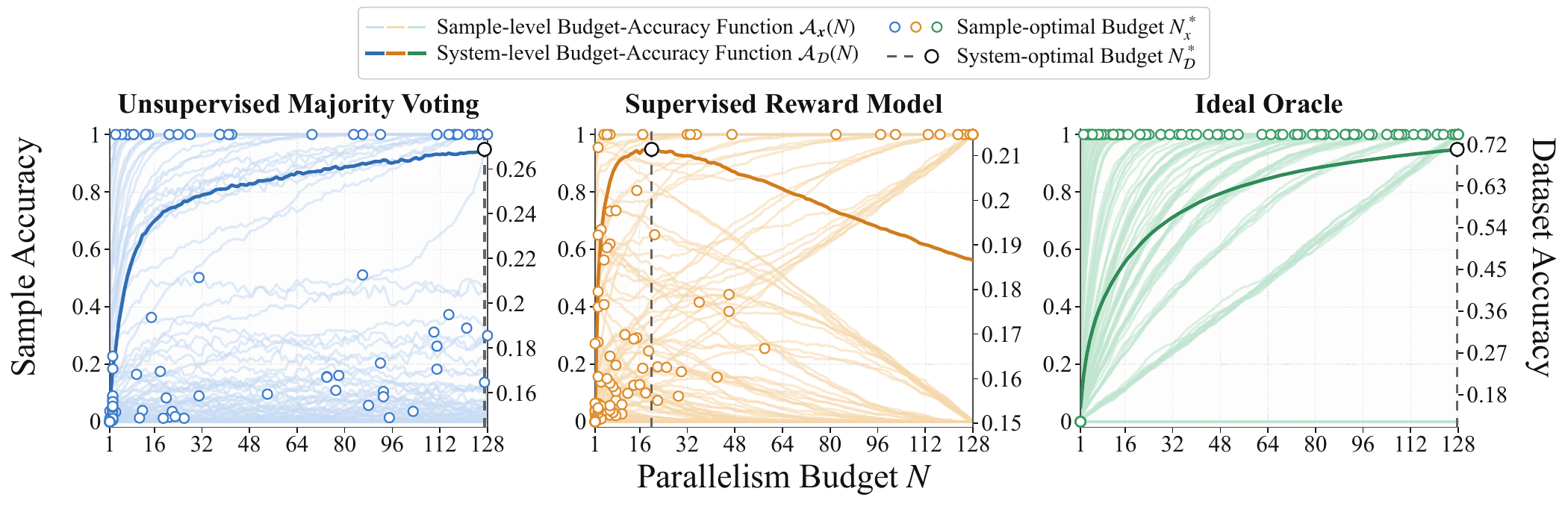}
    \vspace{-0.2in}
    \caption{{\bf Illustration of the Overscaling Curse in Parallel Thinking}. Each subfigure corresponds to a system $(\pi, \mathcal{D})$: evaluating model $\pi$ on dataset $\mathcal{D}$. Here $\pi =$ \texttt{Llama3.1-8B} and $\mathcal{D} = \texttt{AMC}$.}.
    \label{fig:illustration-llama31-amc}
\end{figure}

\begin{figure}[H]
    \centering
    \includegraphics[width=\columnwidth]{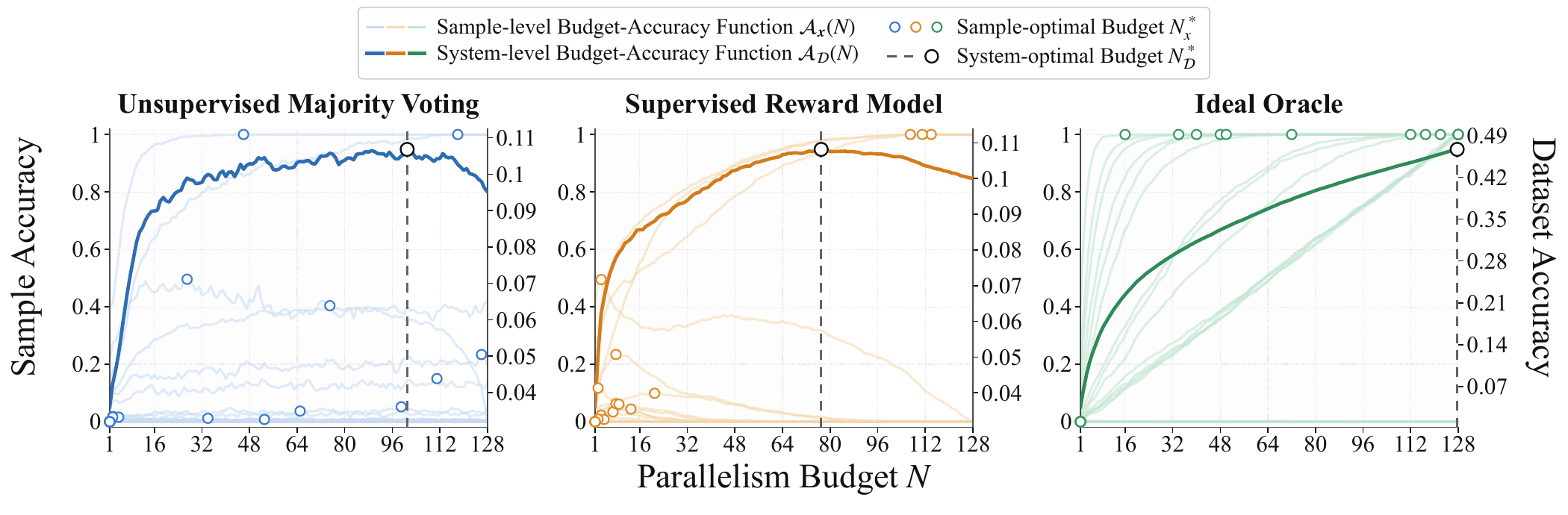}
    \vspace{-0.2in}
    \caption{{\bf Illustration of the Overscaling Curse in Parallel Thinking}. Each subfigure corresponds to a system $(\pi, \mathcal{D})$: evaluating model $\pi$ on dataset $\mathcal{D}$. Here $\pi =$ \texttt{Llama3.1-8B} and $\mathcal{D} = \texttt{AIME24}$.}.
    \label{fig:illustration-llama31-aime24}
\end{figure}

\begin{figure}[H]
    \centering
    \includegraphics[width=\columnwidth]{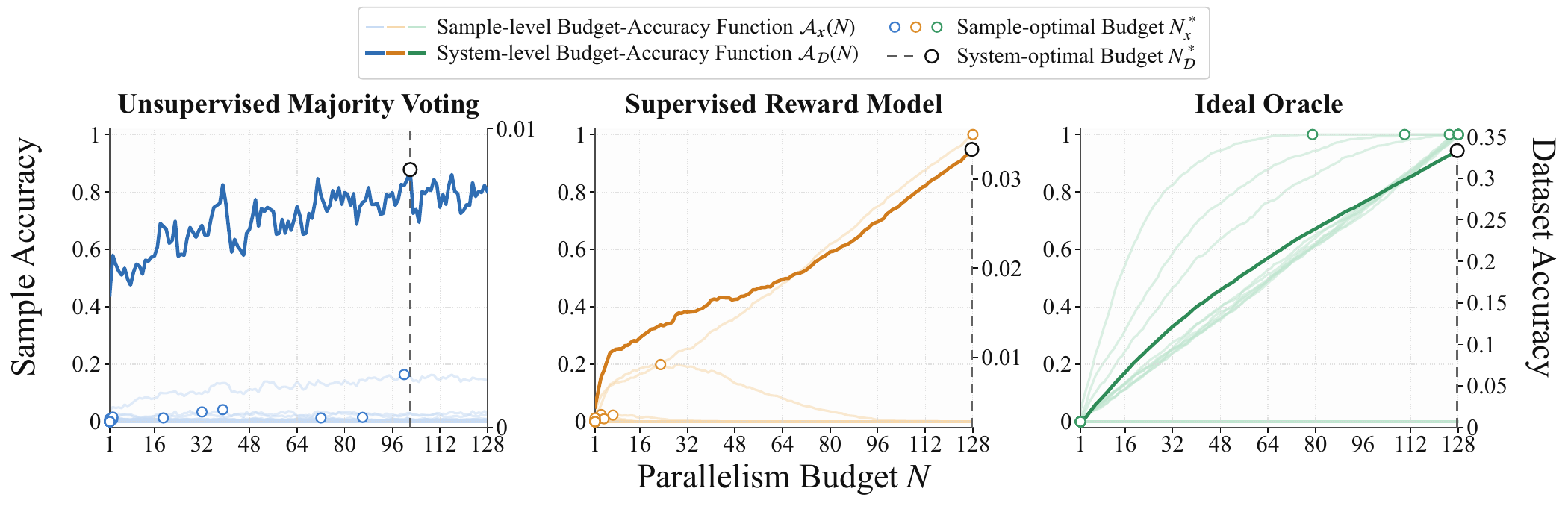}
    \vspace{-0.2in}
    \caption{{\bf Illustration of the Overscaling Curse in Parallel Thinking}. Each subfigure corresponds to a system $(\pi, \mathcal{D})$: evaluating model $\pi$ on dataset $\mathcal{D}$. Here $\pi =$ \texttt{Llama3.1-8B} and $\mathcal{D} = \texttt{AIME25}$.}.
    \label{fig:illustration-llama31-aime25}
\end{figure}

\begin{figure}[H]
    \centering
    \includegraphics[width=\columnwidth]{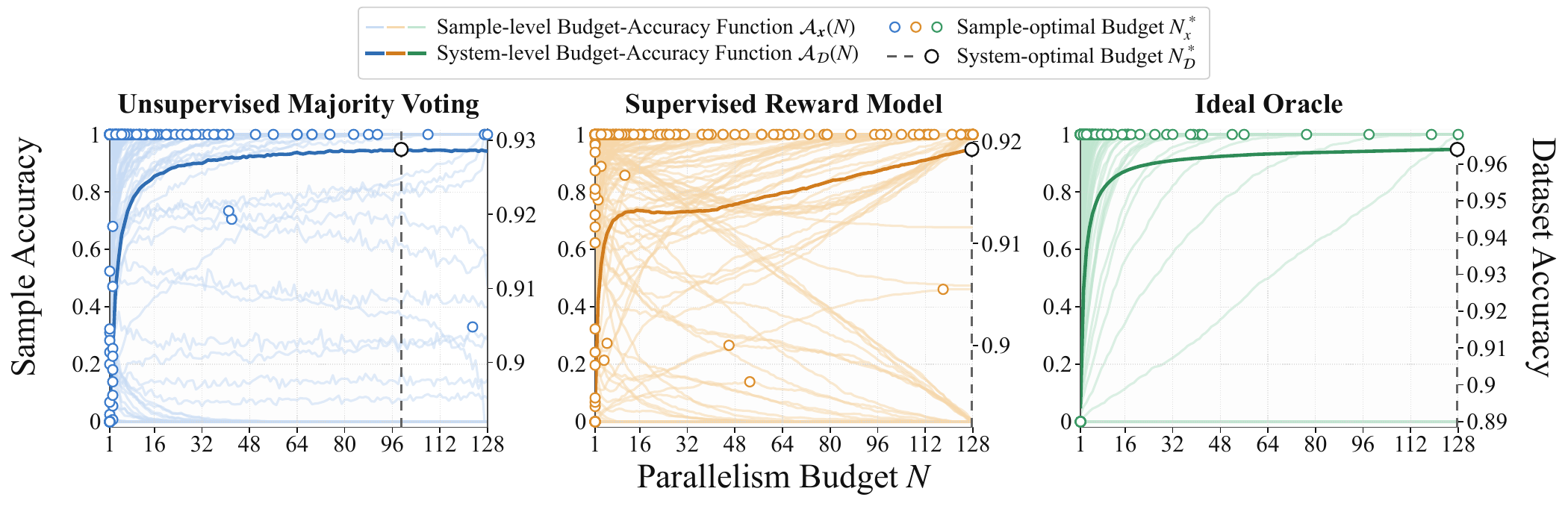}
    \vspace{-0.2in}
    \caption{{\bf Illustration of the Overscaling Curse in Parallel Thinking}. Each subfigure corresponds to a system $(\pi, \mathcal{D})$: evaluating model $\pi$ on dataset $\mathcal{D}$. Here $\pi =$ \texttt{Qwen3-4B} and $\mathcal{D} = \texttt{MATH-500}$.}.
    \label{fig:illustration-qwen3-math}
\end{figure}

\begin{figure}[H]
    \centering
    \includegraphics[width=\columnwidth]{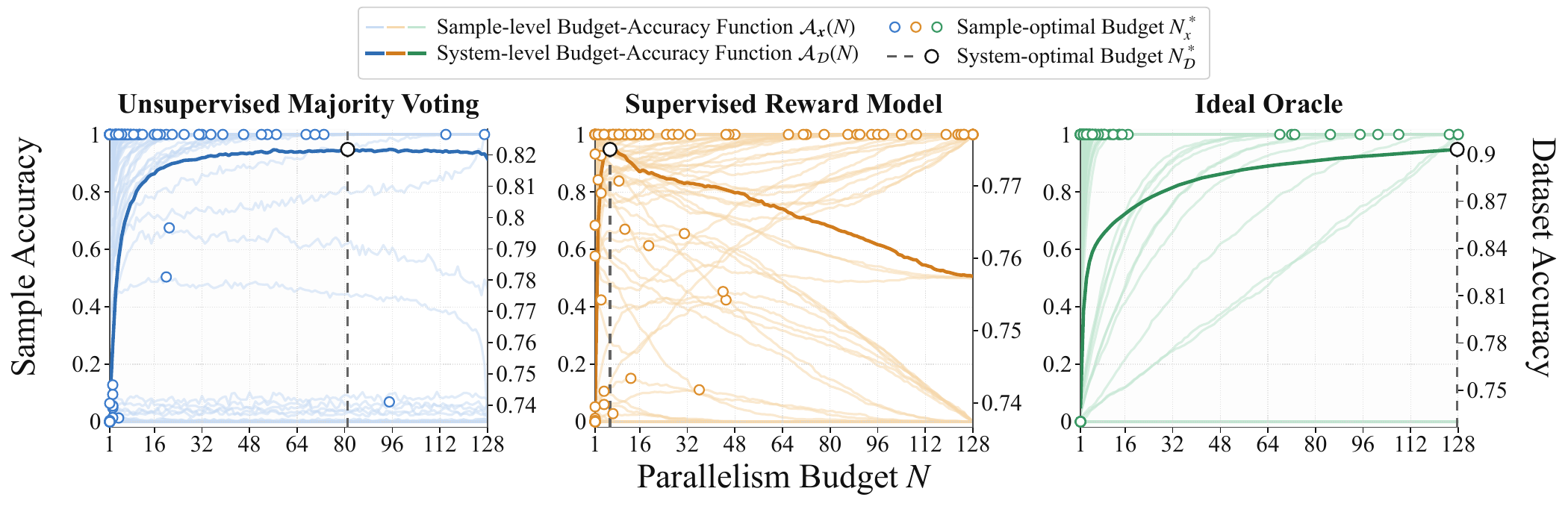}
    \vspace{-0.2in}
    \caption{{\bf Illustration of the Overscaling Curse in Parallel Thinking}. Each subfigure corresponds to a system $(\pi, \mathcal{D})$: evaluating model $\pi$ on dataset $\mathcal{D}$. Here $\pi =$ \texttt{Qwen3-4B} and $\mathcal{D} = \texttt{AMC}$.}.
    \label{fig:illustration-qwen3-amc}
\end{figure}

\begin{figure}[H]
    \centering
    \includegraphics[width=\columnwidth]{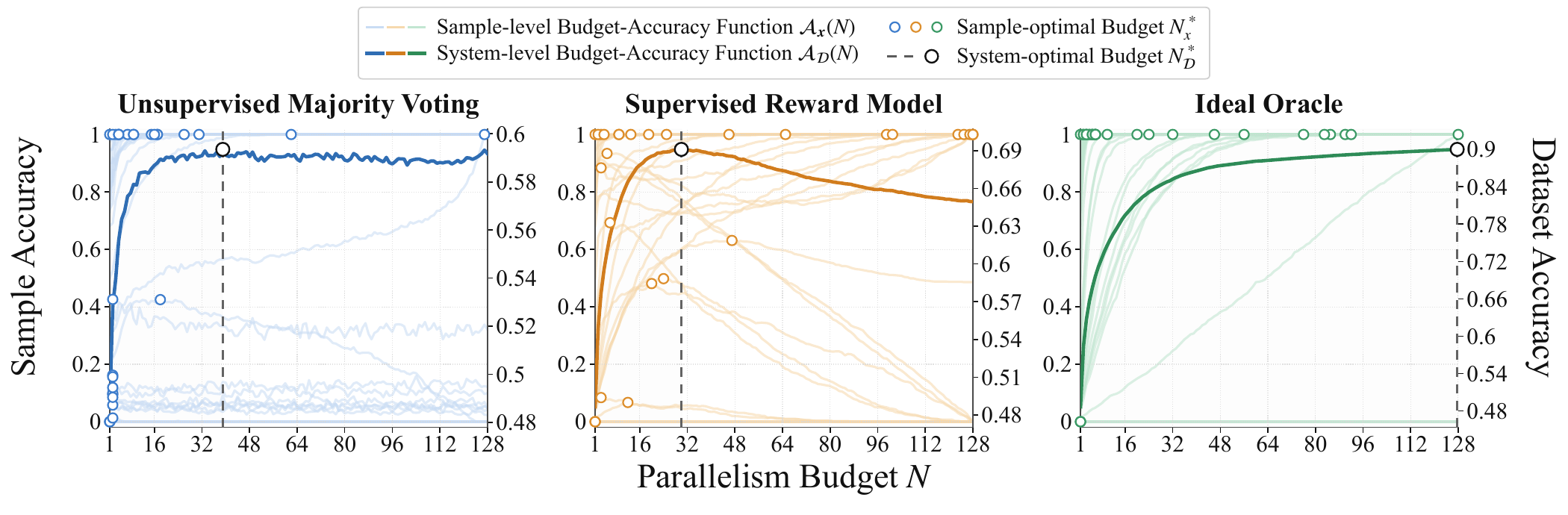}
    \vspace{-0.2in}
    \caption{{\bf Illustration of the Overscaling Curse in Parallel Thinking}. Each subfigure corresponds to a system $(\pi, \mathcal{D})$: evaluating model $\pi$ on dataset $\mathcal{D}$. Here $\pi =$ \texttt{Qwen3-4B} and $\mathcal{D} = \texttt{AIME24}$.}.
    \label{fig:illustration-qwen3-aime24}
\end{figure}

\begin{figure}[H]
    \centering
    \includegraphics[width=\columnwidth]{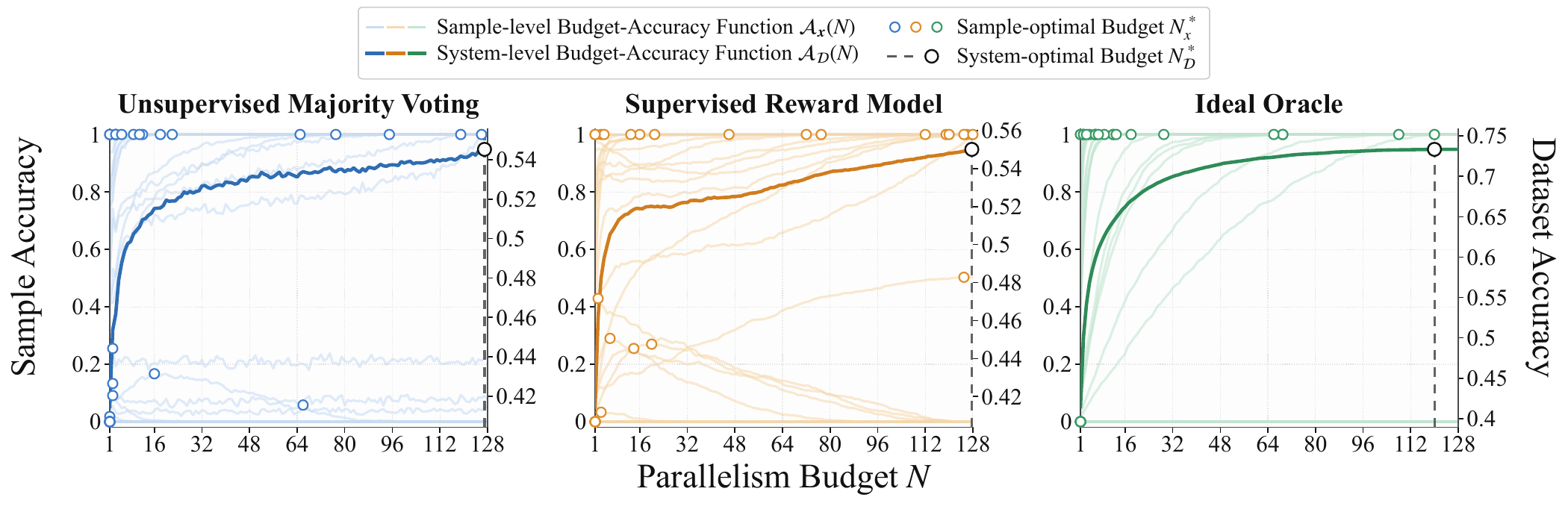}
    \vspace{-0.2in}
    \caption{{\bf Illustration of the Overscaling Curse in Parallel Thinking}. Each subfigure corresponds to a system $(\pi, \mathcal{D})$: evaluating model $\pi$ on dataset $\mathcal{D}$. Here $\pi =$ \texttt{Qwen3-4B} and $\mathcal{D} = \texttt{AIME25}$.}.
    \label{fig:illustration-qwen3-aime25}
\end{figure}

\textcolor{deepred}{Figures \ref{fig:illustration-qwen25-amc} -- \ref{fig:illustration-qwen3-aime25}} present all illustrations of the overscaling curse across all systems under various aggregation strategies, serving as a complete supplement to \textcolor{deepred}{Figure \ref{fig:illustration-qwen25-math}}.

Within each figure, each subfigure corresponds to one answer aggregation strategy.
The single thick curve shows how overall dataset accuracy varies with the budget, and thin curves show how sample accuracy varies with the budget for each sample $\bm{x} \in \mathcal{D}$. We term these curves the {\it ``budget-accuracy'' functions}.
Hollow markers on each curve indicate the minimum budget that achieves the highest accuracy, which we term the {\it ``optimal budget''}.
See \textcolor{deepred}{Section \ref{sec:global}} for experimental setups and \textcolor{deepred}{Section \ref{sec:formalization}} for detailed function computation.

Due to the large data sizes of the GPQA (448 samples), MMLU-Pro (12,000 samples), and BrowseComp (1,255 + 289 samples) datasets, many individual curves would overlap and become unreadable. Therefore, we omit their illustrations here and present only the quantitative results in \textcolor{deepred}{Section \ref{sec:how-overscaling}}.


%% file: 8B-overscaling-quantification.tex
\section{Quantification of Overscaling Curse}
\label{appe:quantify}

\begin{figure}[H]
    \vspace{-0.2in}
    \centering
    \includegraphics[width=\columnwidth]{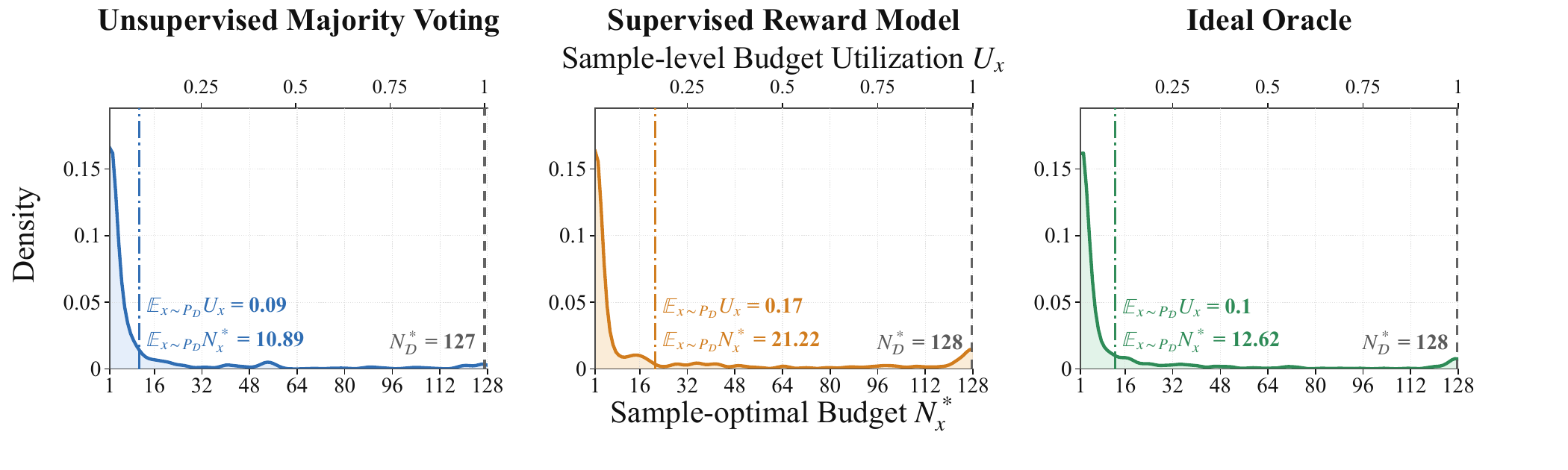}
    \vspace{-0.35in}
    \caption{Distribution of {\it Sample-level Budget Utilization $\mathcal{U}_{\bm{x}}$} and corresponding optimal budgets $N^*_{\bm{x}}$ in a single systems $(\pi,\mathcal{D})$ under different aggregation strategies, along with $N^*_{\mathcal{D}}$. Here $\pi = \texttt{Qwen2.5-7B}$ and $\mathcal{D} = \texttt{MATH-500}$.}
    \label{fig:distribution-qwen25-math}
\end{figure}

\begin{figure}[H]
    \vspace{-0.2in}
    \centering
    \includegraphics[width=\columnwidth]{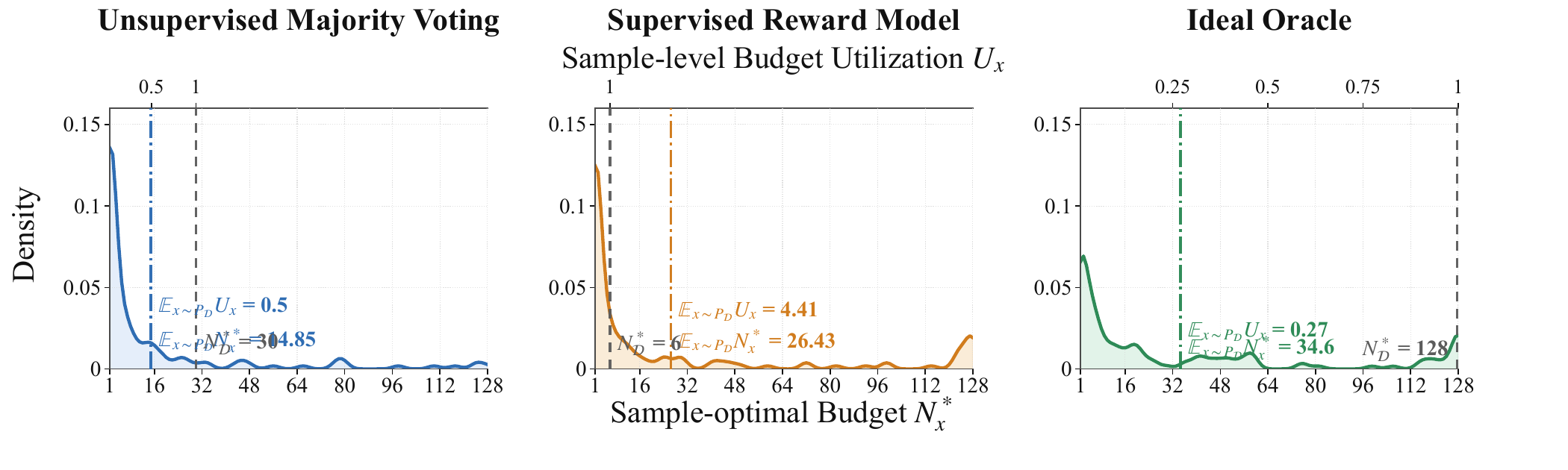}
    \vspace{-0.35in}
    \caption{Distribution of {\it Sample-level Budget Utilization $\mathcal{U}_{\bm{x}}$} and corresponding optimal budgets $N^*_{\bm{x}}$ in a single systems $(\pi,\mathcal{D})$ under different aggregation strategies, along with $N^*_{\mathcal{D}}$. Here $\pi = \texttt{Qwen2.5-7B}$ and $\mathcal{D} = \texttt{AMC}$.}
    \label{fig:distribution-qwen25-amc}
\end{figure}

\begin{figure}[H]
    \vspace{-0.2in}
    \centering
    \includegraphics[width=\columnwidth]{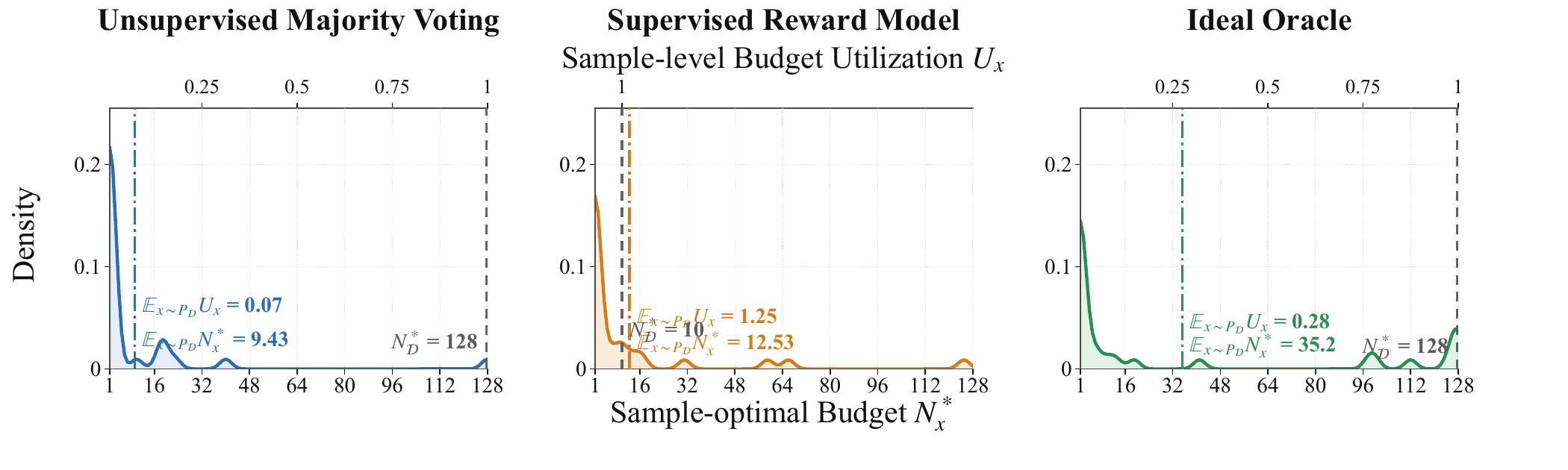}
    \vspace{-0.35in}
    \caption{Distribution of {\it Sample-level Budget Utilization $\mathcal{U}_{\bm{x}}$} and corresponding optimal budgets $N^*_{\bm{x}}$ in a single systems $(\pi,\mathcal{D})$ under different aggregation strategies, along with $N^*_{\mathcal{D}}$. Here $\pi = \texttt{Qwen2.5-7B}$ and $\mathcal{D} = \texttt{AIME24}$.}
    \label{fig:distribution-qwen25-aime24}
\end{figure}

\begin{figure}[H]
    \vspace{-0.2in}
    \centering
    \includegraphics[width=\columnwidth]{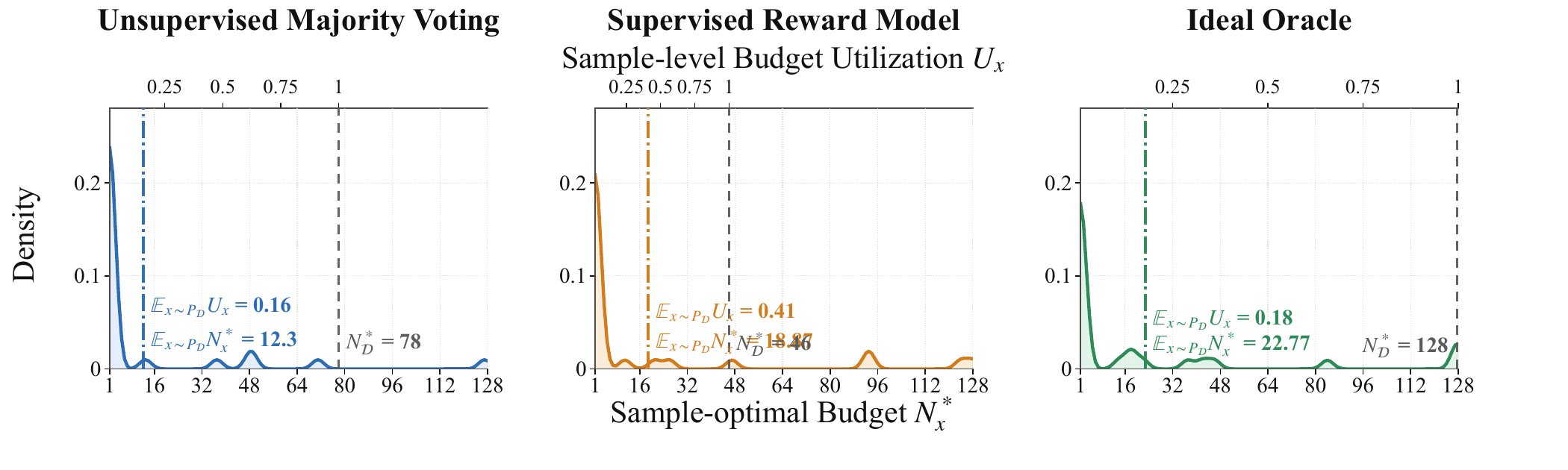}
    \vspace{-0.35in}
    \caption{Distribution of {\it Sample-level Budget Utilization $\mathcal{U}_{\bm{x}}$} and corresponding optimal budgets $N^*_{\bm{x}}$ in a single systems $(\pi,\mathcal{D})$ under different aggregation strategies, along with $N^*_{\mathcal{D}}$. Here $\pi = \texttt{Qwen2.5-7B}$ and $\mathcal{D} = \texttt{AIME25}$.}
    \label{fig:distribution-qwen25-aime25}
\end{figure}

\begin{figure}[H]
    \vspace{-0.2in}
    \centering
    \includegraphics[width=\columnwidth]{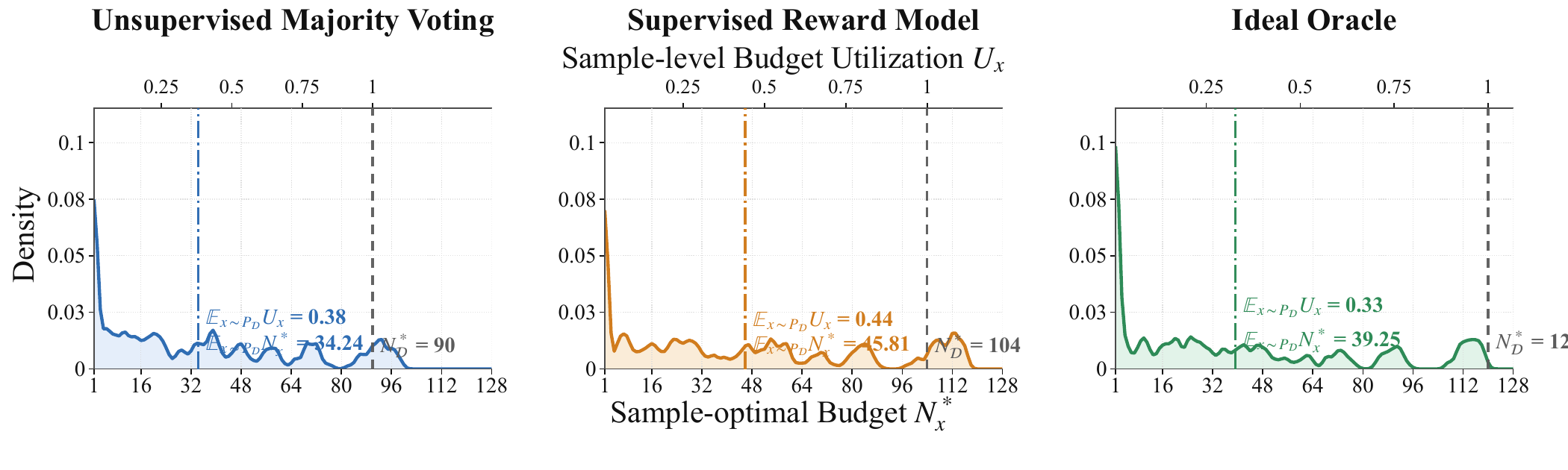}
    \vspace{-0.35in}
    \caption{Distribution of {\it Sample-level Budget Utilization $\mathcal{U}_{\bm{x}}$} and corresponding optimal budgets $N^*_{\bm{x}}$ in a single systems $(\pi,\mathcal{D})$ under different aggregation strategies, along with $N^*_{\mathcal{D}}$. Here $\pi = \texttt{Qwen2.5-7B}$ and $\mathcal{D} = \texttt{GPQA}$.}
    \label{fig:distribution-qwen25-gpqa}
\end{figure}

\begin{figure}[H]
    \vspace{-0.2in}
    \centering
    \includegraphics[width=\columnwidth]{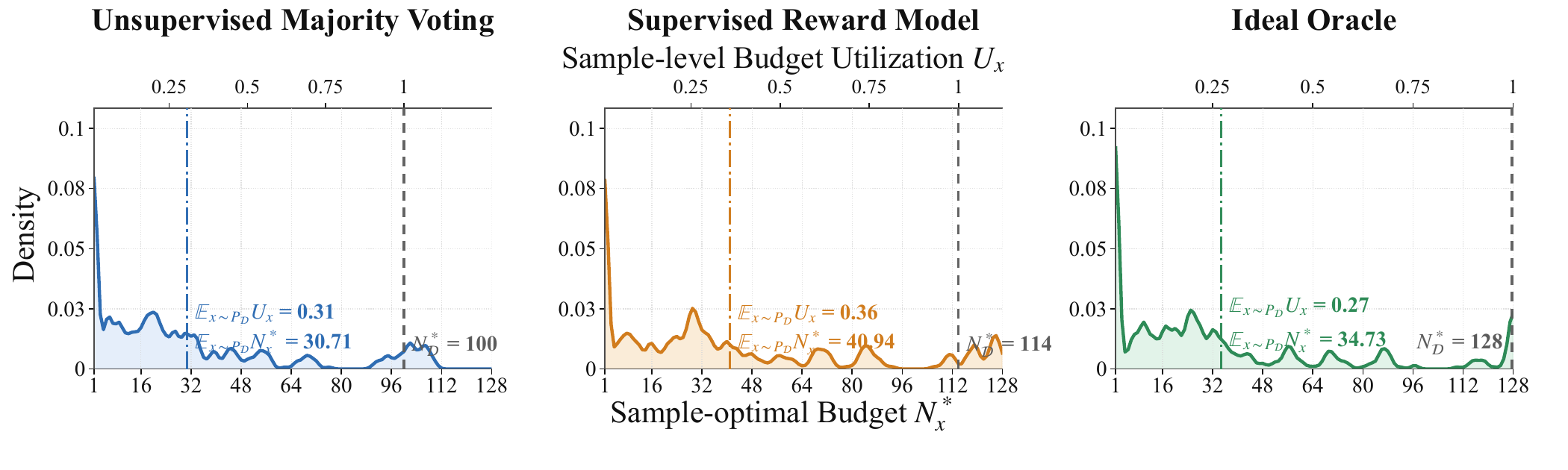}
    \vspace{-0.35in}
    \caption{Distribution of {\it Sample-level Budget Utilization $\mathcal{U}_{\bm{x}}$} and corresponding optimal budgets $N^*_{\bm{x}}$ in a single systems $(\pi,\mathcal{D})$ under different aggregation strategies, along with $N^*_{\mathcal{D}}$. Here $\pi = \texttt{Qwen2.5-7B}$ and $\mathcal{D} = \texttt{MMLU-Pro}$.}
    \label{fig:distribution-qwen25-mmlu}
\end{figure}

\begin{figure}[H]
    \vspace{-0.2in}
    \centering
    \includegraphics[width=\columnwidth]{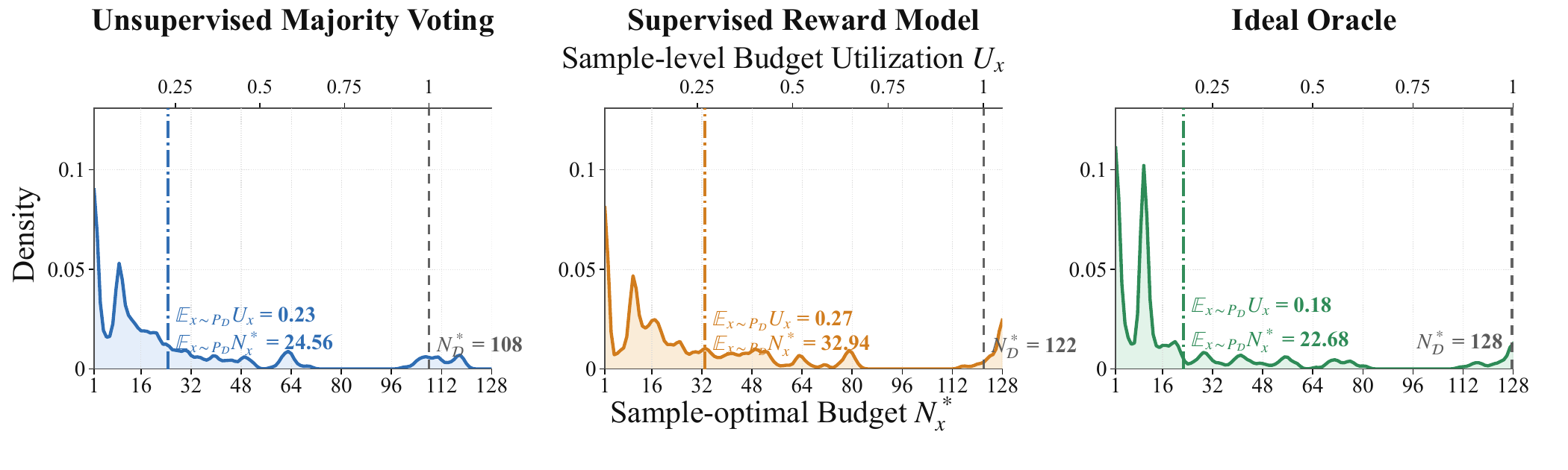}
    \vspace{-0.35in}
    \caption{Distribution of {\it Sample-level Budget Utilization $\mathcal{U}_{\bm{x}}$} and corresponding optimal budgets $N^*_{\bm{x}}$ in a single systems $(\pi,\mathcal{D})$ under different aggregation strategies, along with $N^*_{\mathcal{D}}$. Here $\pi = \texttt{Qwen2.5-7B}$ and $\mathcal{D} = \texttt{BrowseComp}$.}
    \label{fig:distribution-qwen25-browsecomp}
\end{figure}

\begin{figure}[H]
    \vspace{-0.2in}
    \centering
    \includegraphics[width=\columnwidth]{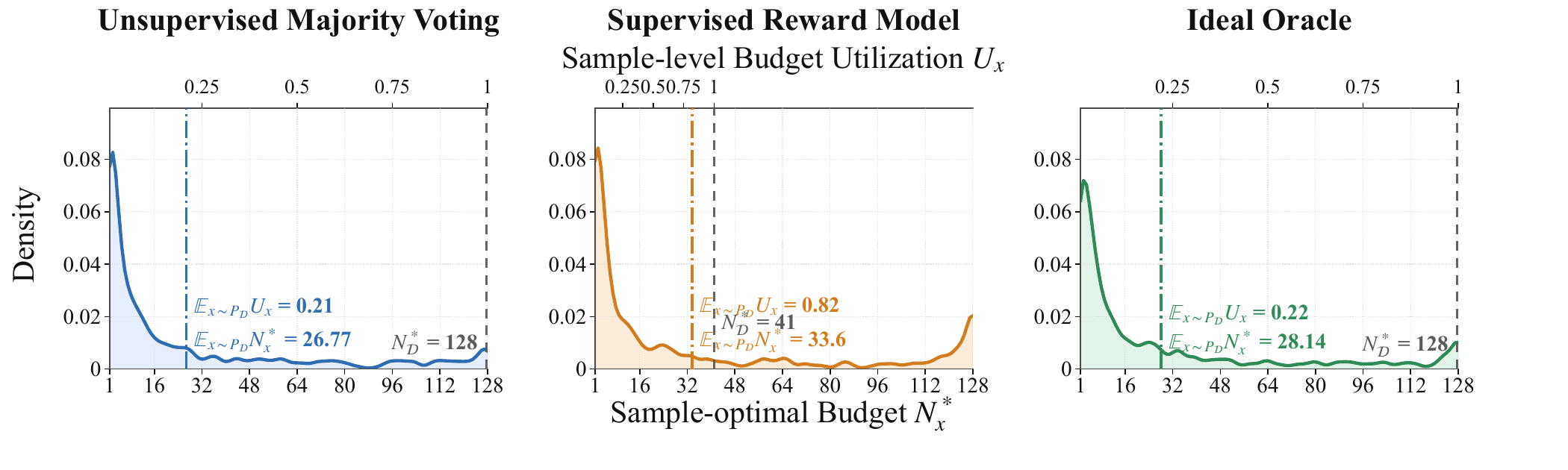}
    \vspace{-0.35in}
    \caption{Distribution of {\it Sample-level Budget Utilization $\mathcal{U}_{\bm{x}}$} and corresponding optimal budgets $N^*_{\bm{x}}$ in a single systems $(\pi,\mathcal{D})$ under different aggregation strategies, along with $N^*_{\mathcal{D}}$. Here $\pi = \texttt{Llama3.1-8B}$ and $\mathcal{D} = \texttt{MATH-500}$.}
    \label{fig:distribution-llama31-math}
\end{figure}

\begin{figure}[H]
    \vspace{-0.2in}
    \centering
    \includegraphics[width=\columnwidth]{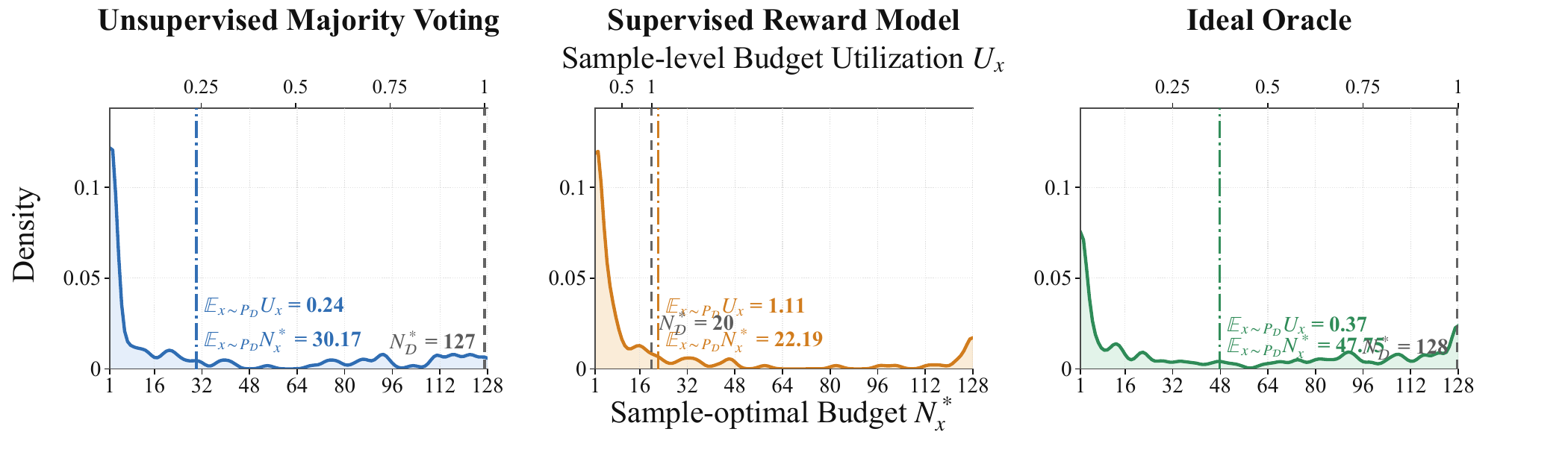}
    \vspace{-0.35in}
    \caption{Distribution of {\it Sample-level Budget Utilization $\mathcal{U}_{\bm{x}}$} and corresponding optimal budgets $N^*_{\bm{x}}$ in a single systems $(\pi,\mathcal{D})$ under different aggregation strategies, along with $N^*_{\mathcal{D}}$. Here $\pi = \texttt{Llama3.1-8B}$ and $\mathcal{D} = \texttt{AMC}$.}
    \label{fig:distribution-llama31-amc}
\end{figure}

\begin{figure}[H]
    \vspace{-0.2in}
    \centering
    \includegraphics[width=\columnwidth]{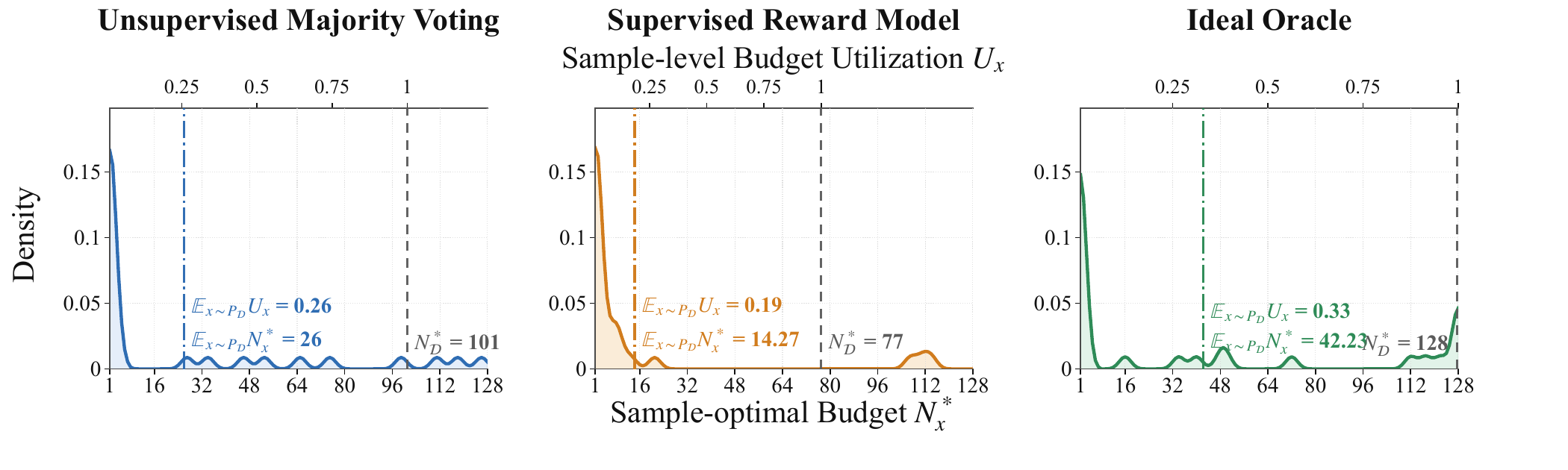}
    \vspace{-0.35in}
    \caption{Distribution of {\it Sample-level Budget Utilization $\mathcal{U}_{\bm{x}}$} and corresponding optimal budgets $N^*_{\bm{x}}$ in a single systems $(\pi,\mathcal{D})$ under different aggregation strategies, along with $N^*_{\mathcal{D}}$. Here $\pi = \texttt{Llama3.1-8B}$ and $\mathcal{D} = \texttt{AIME24}$.}
    \label{fig:distribution-llama31-aime24}
\end{figure}

\begin{figure}[H]
    \vspace{-0.2in}
    \centering
    \includegraphics[width=\columnwidth]{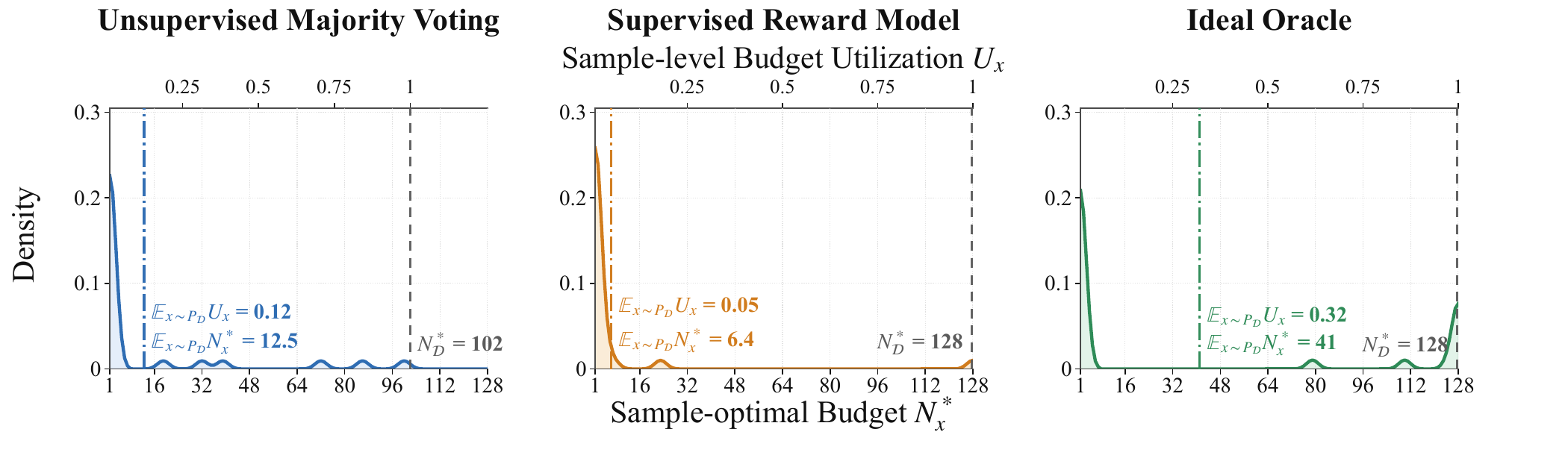}
    \vspace{-0.35in}
    \caption{Distribution of {\it Sample-level Budget Utilization $\mathcal{U}_{\bm{x}}$} and corresponding optimal budgets $N^*_{\bm{x}}$ in a single systems $(\pi,\mathcal{D})$ under different aggregation strategies, along with $N^*_{\mathcal{D}}$. Here $\pi = \texttt{Llama3.1-8B}$ and $\mathcal{D} = \texttt{AIME25}$.}
    \label{fig:distribution-llama31-aime25}
\end{figure}

\begin{figure}[H]
    \vspace{-0.2in}
    \centering
    \includegraphics[width=\columnwidth]{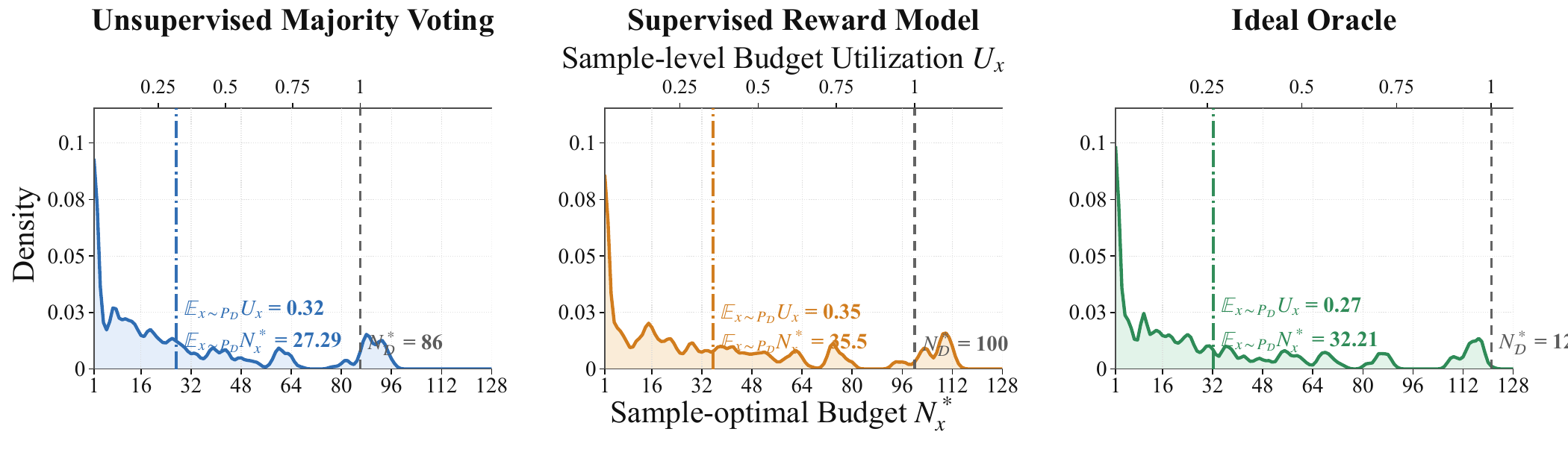}
    \vspace{-0.35in}
    \caption{Distribution of {\it Sample-level Budget Utilization $\mathcal{U}_{\bm{x}}$} and corresponding optimal budgets $N^*_{\bm{x}}$ in a single systems $(\pi,\mathcal{D})$ under different aggregation strategies, along with $N^*_{\mathcal{D}}$. Here $\pi = \texttt{Llama3.1-8B}$ and $\mathcal{D} = \texttt{GPQA}$.}
    \label{fig:distribution-llama31-gpqa}
\end{figure}

\begin{figure}[H]
    \vspace{-0.2in}
    \centering
    \includegraphics[width=\columnwidth]{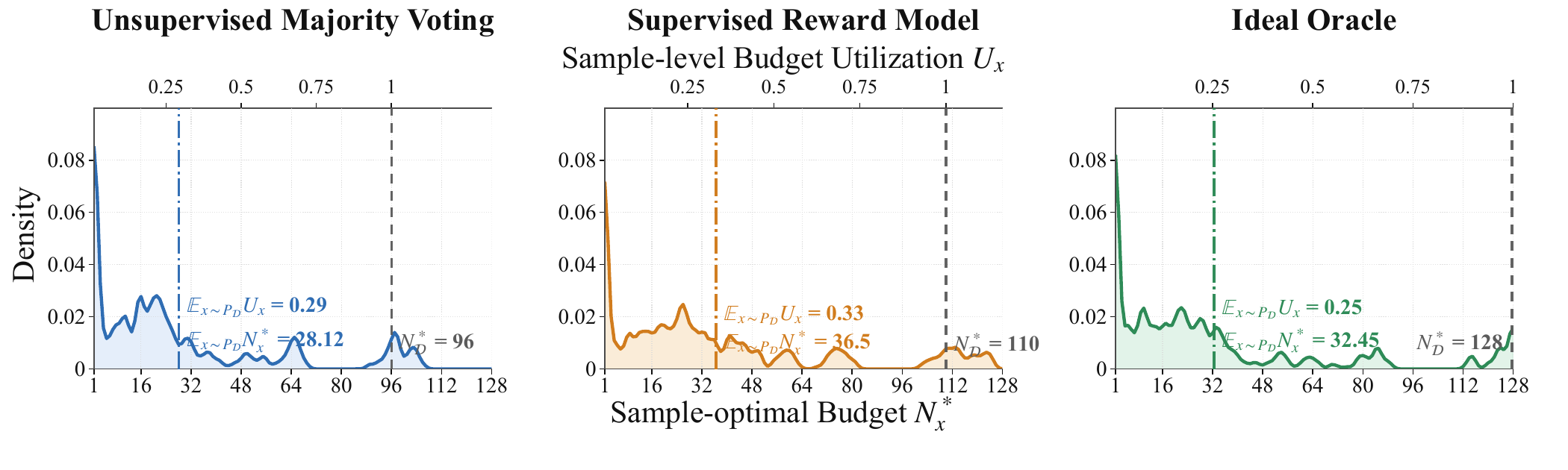}
    \vspace{-0.35in}
    \caption{Distribution of {\it Sample-level Budget Utilization $\mathcal{U}_{\bm{x}}$} and corresponding optimal budgets $N^*_{\bm{x}}$ in a single systems $(\pi,\mathcal{D})$ under different aggregation strategies, along with $N^*_{\mathcal{D}}$. Here $\pi = \texttt{Llama3.1-8B}$ and $\mathcal{D} = \texttt{MMLU-Pro}$.}
    \label{fig:distribution-llama31-mmlu}
\end{figure}

\begin{figure}[H]
    \vspace{-0.2in}
    \centering
    \includegraphics[width=\columnwidth]{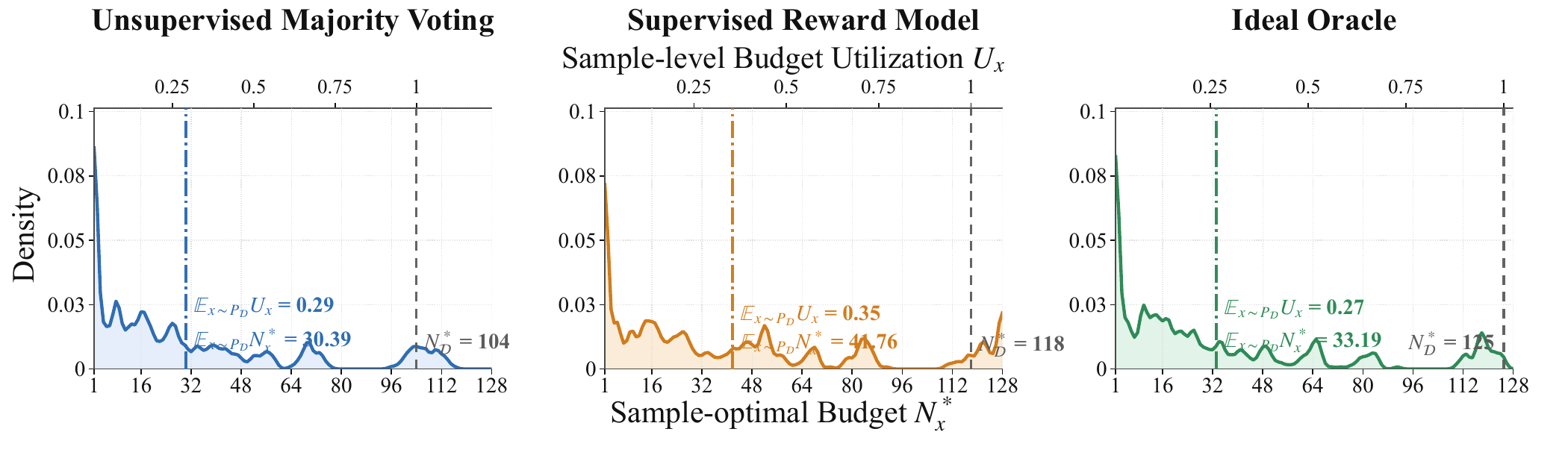}
    \vspace{-0.35in}
    \caption{Distribution of {\it Sample-level Budget Utilization $\mathcal{U}_{\bm{x}}$} and corresponding optimal budgets $N^*_{\bm{x}}$ in a single systems $(\pi,\mathcal{D})$ under different aggregation strategies, along with $N^*_{\mathcal{D}}$. Here $\pi = \texttt{Llama3.1-8B}$ and $\mathcal{D} = \texttt{BrowseComp}$.}
    \label{fig:distribution-llama31-browsecomp}
\end{figure}

\begin{figure}[H]
    \vspace{-0.2in}
    \centering
    \includegraphics[width=\columnwidth]{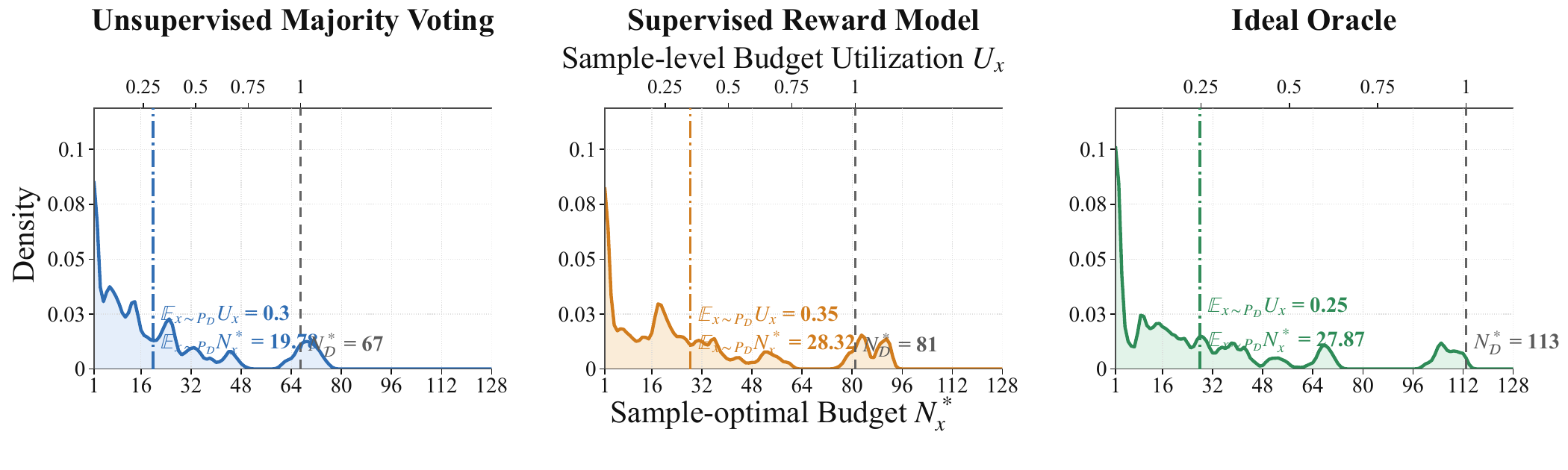}
    \vspace{-0.35in}
    \caption{Distribution of {\it Sample-level Budget Utilization $\mathcal{U}_{\bm{x}}$} and corresponding optimal budgets $N^*_{\bm{x}}$ in a single systems $(\pi,\mathcal{D})$ under different aggregation strategies, along with $N^*_{\mathcal{D}}$. Here $\pi = \texttt{Deepseek-R1-Distill-Qwen-7B}$ and $\mathcal{D} = \texttt{MATH-500}$.}
    \label{fig:distribution-r1-math}
\end{figure}

\begin{figure}[H]
    \vspace{-0.2in}
    \centering
    \includegraphics[width=\columnwidth]{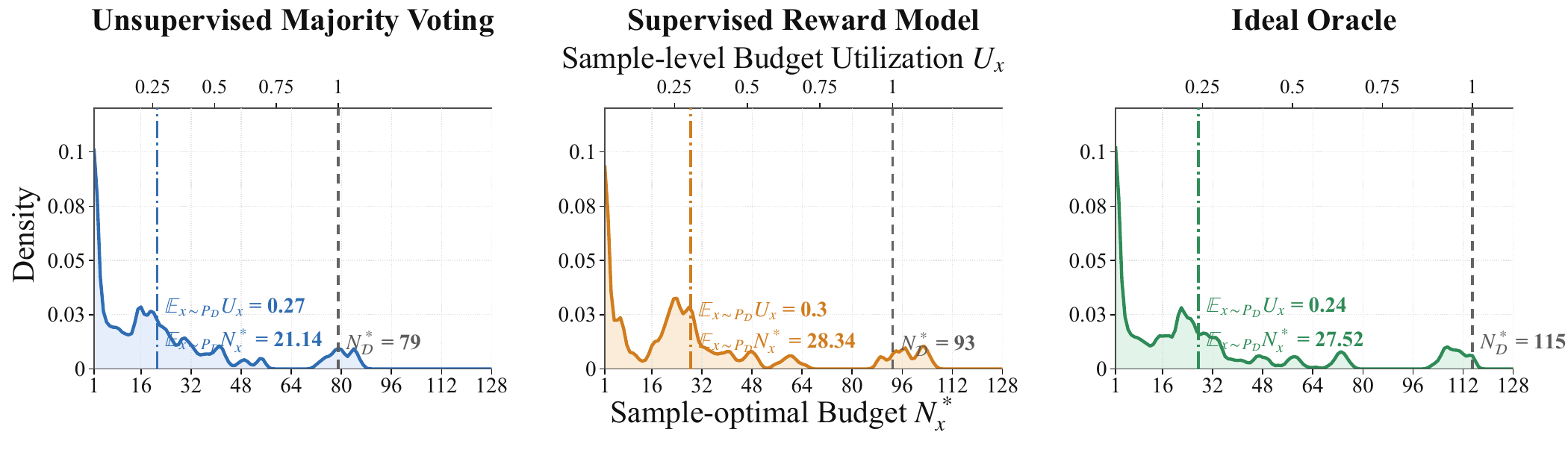}
    \vspace{-0.35in}
    \caption{Distribution of {\it Sample-level Budget Utilization $\mathcal{U}_{\bm{x}}$} and corresponding optimal budgets $N^*_{\bm{x}}$ in a single systems $(\pi,\mathcal{D})$ under different aggregation strategies, along with $N^*_{\mathcal{D}}$. Here $\pi = \texttt{Deepseek-R1-Distill-Qwen-7B}$ and $\mathcal{D} = \texttt{AMC}$.}
    \label{fig:distribution-r1-amc}
\end{figure}

\begin{figure}[H]
    \vspace{-0.2in}
    \centering
    \includegraphics[width=\columnwidth]{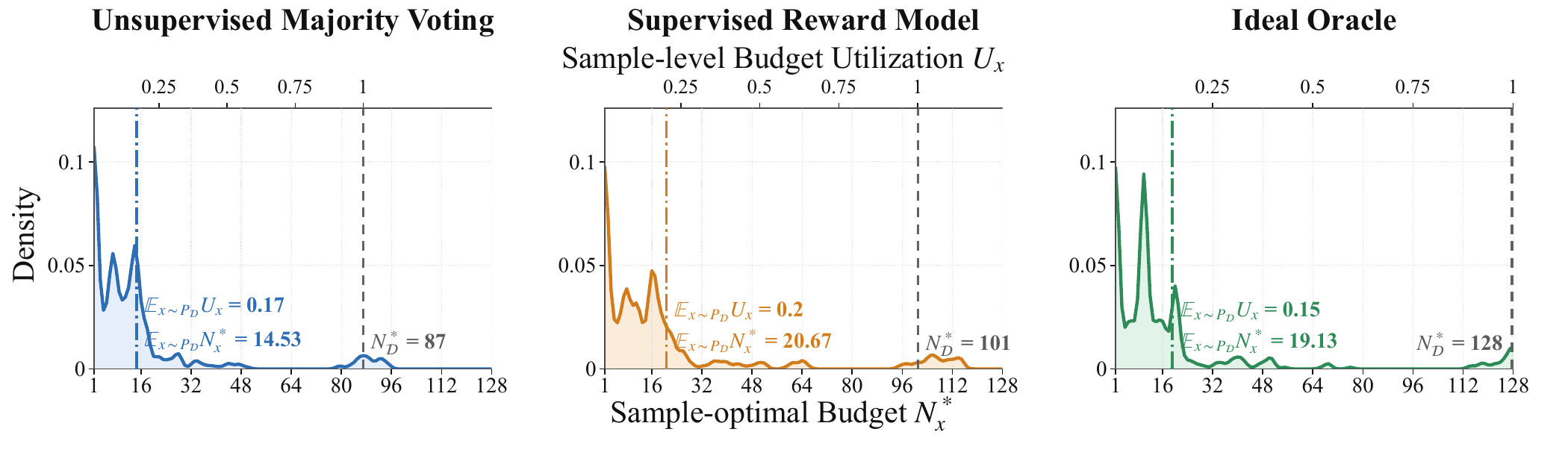}
    \vspace{-0.35in}
    \caption{Distribution of {\it Sample-level Budget Utilization $\mathcal{U}_{\bm{x}}$} and corresponding optimal budgets $N^*_{\bm{x}}$ in a single systems $(\pi,\mathcal{D})$ under different aggregation strategies, along with $N^*_{\mathcal{D}}$. Here $\pi = \texttt{Deepseek-R1-Distill-Qwen-7B}$ and $\mathcal{D} = \texttt{AIME24}$.}
    \label{fig:distribution-r1-aime24}
\end{figure}

\begin{figure}[H]
    \vspace{-0.2in}
    \centering
    \includegraphics[width=\columnwidth]{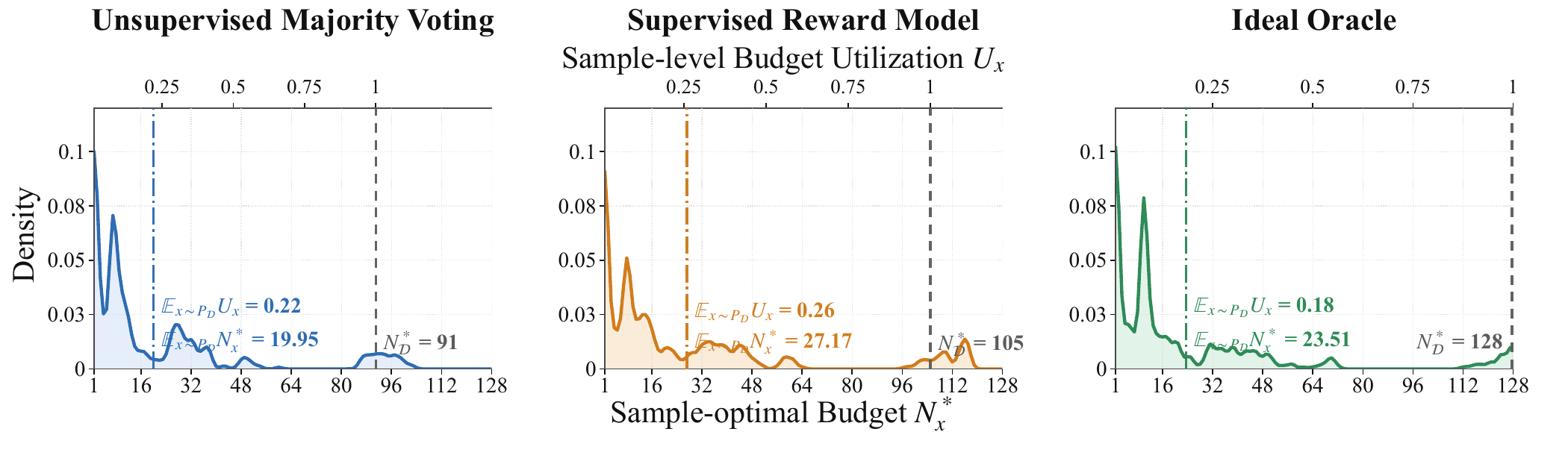}
    \vspace{-0.35in}
    \caption{Distribution of {\it Sample-level Budget Utilization $\mathcal{U}_{\bm{x}}$} and corresponding optimal budgets $N^*_{\bm{x}}$ in a single systems $(\pi,\mathcal{D})$ under different aggregation strategies, along with $N^*_{\mathcal{D}}$. Here $\pi = \texttt{Deepseek-R1-Distill-Qwen-7B}$ and $\mathcal{D} = \texttt{AIME25}$.}
    \label{fig:distribution-r1-aime25}
\end{figure}

\begin{figure}[H]
    \vspace{-0.2in}
    \centering
    \includegraphics[width=\columnwidth]{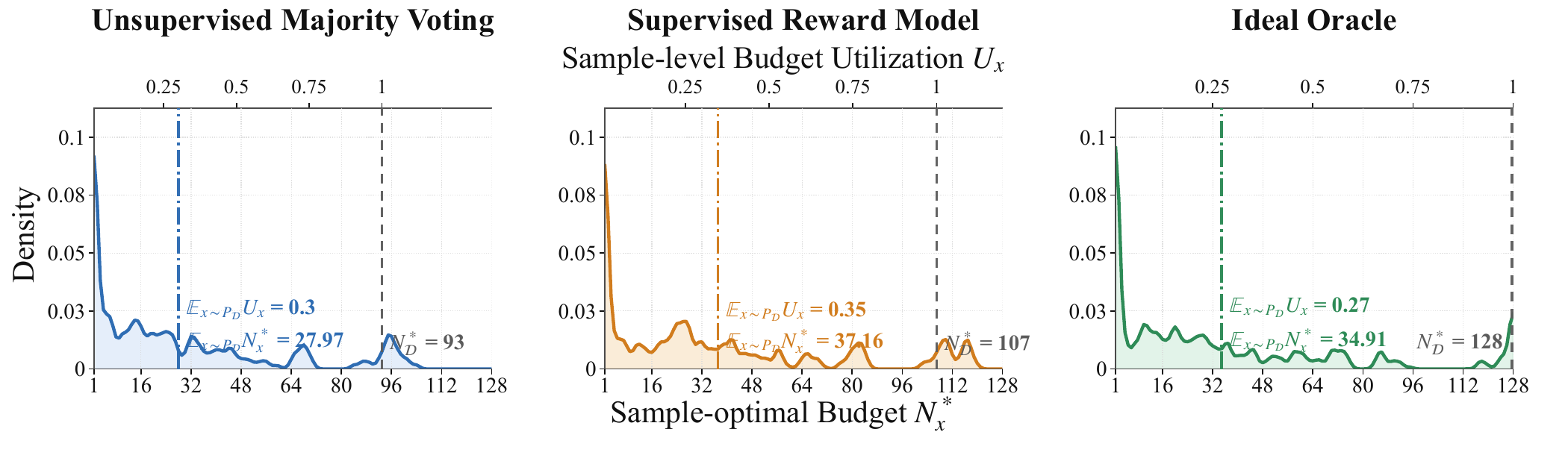}
    \vspace{-0.35in}
    \caption{Distribution of {\it Sample-level Budget Utilization $\mathcal{U}_{\bm{x}}$} and corresponding optimal budgets $N^*_{\bm{x}}$ in a single systems $(\pi,\mathcal{D})$ under different aggregation strategies, along with $N^*_{\mathcal{D}}$. Here $\pi = \texttt{Deepseek-R1-Distill-Qwen-7B}$ and $\mathcal{D} = \texttt{GPQA}$.}
    \label{fig:distribution-r1-gpqa}
\end{figure}

\begin{figure}[H]
    \vspace{-0.2in}
    \centering
    \includegraphics[width=\columnwidth]{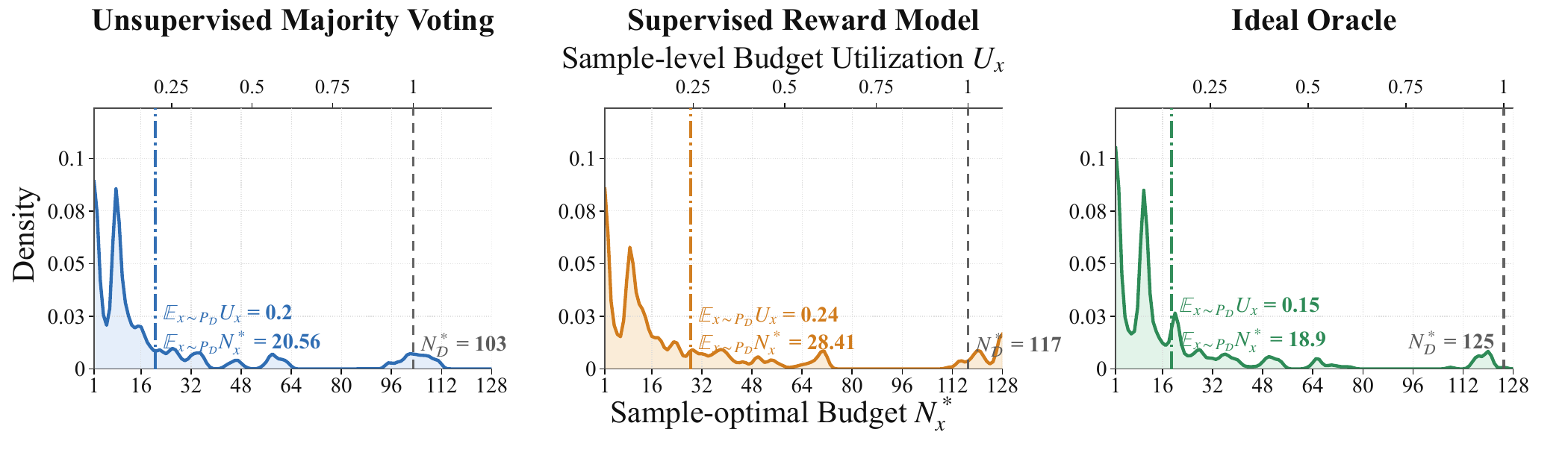}
    \vspace{-0.35in}
    \caption{Distribution of {\it Sample-level Budget Utilization $\mathcal{U}_{\bm{x}}$} and corresponding optimal budgets $N^*_{\bm{x}}$ in a single systems $(\pi,\mathcal{D})$ under different aggregation strategies, along with $N^*_{\mathcal{D}}$. Here $\pi = \texttt{Deepseek-R1-Distill-Qwen-7B}$ and $\mathcal{D} = \texttt{MMLU-Pro}$.}
    \label{fig:distribution-r1-mmlu}
\end{figure}

\begin{figure}[H]
    \vspace{-0.2in}
    \centering
    \includegraphics[width=\columnwidth]{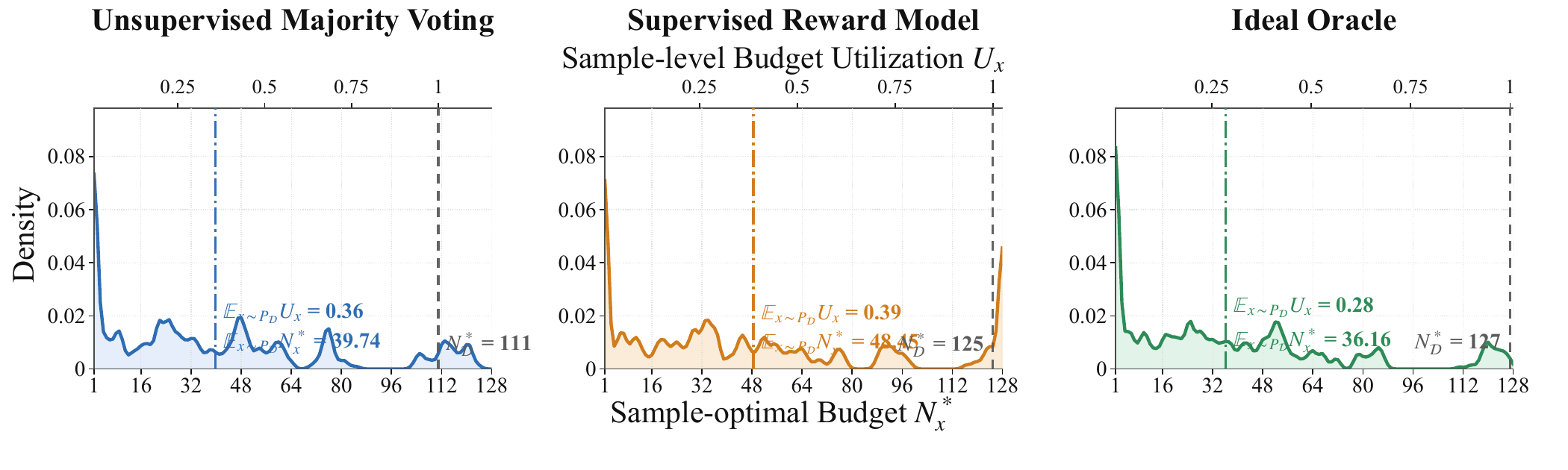}
    \vspace{-0.35in}
    \caption{Distribution of {\it Sample-level Budget Utilization $\mathcal{U}_{\bm{x}}$} and corresponding optimal budgets $N^*_{\bm{x}}$ in a single systems $(\pi,\mathcal{D})$ under different aggregation strategies, along with $N^*_{\mathcal{D}}$. Here $\pi = \texttt{Deepseek-R1-Distill-Qwen-7B}$ and $\mathcal{D} = \texttt{BrowseComp}$.}
    \label{fig:distribution-r1-browsecomp}
\end{figure}

\begin{figure}[H]
    \vspace{-0.2in}
    \centering
    \includegraphics[width=\columnwidth]{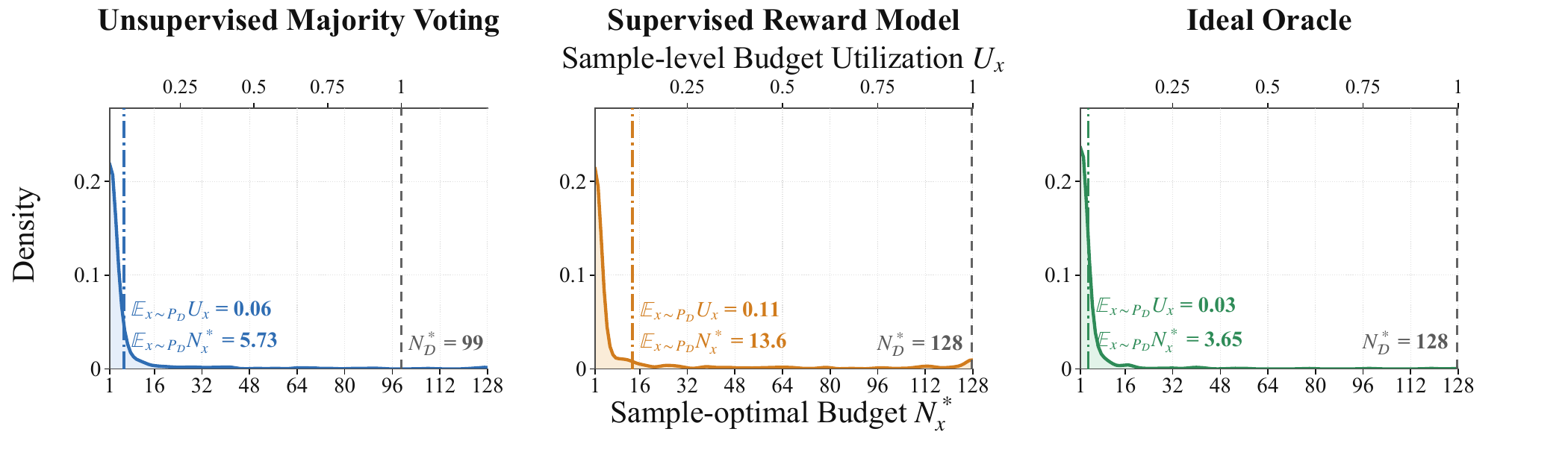}
    \vspace{-0.35in}
    \caption{Distribution of {\it Sample-level Budget Utilization $\mathcal{U}_{\bm{x}}$} and corresponding optimal budgets $N^*_{\bm{x}}$ in a single systems $(\pi,\mathcal{D})$ under different aggregation strategies, along with $N^*_{\mathcal{D}}$. Here $\pi = \texttt{Qwen3-4B}$ and $\mathcal{D} = \texttt{MATH-500}$.}
    \label{fig:distribution-qwen3-math}
\end{figure}

\begin{figure}[H]
    \vspace{-0.2in}
    \centering
    \includegraphics[width=\columnwidth]{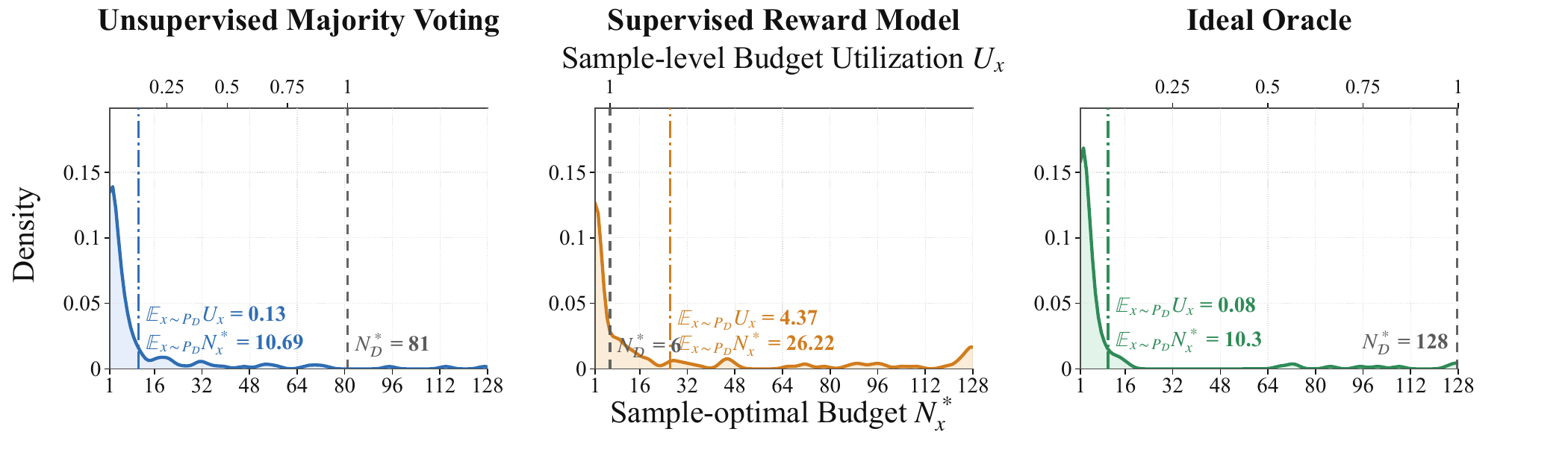}
    \vspace{-0.35in}
    \caption{Distribution of {\it Sample-level Budget Utilization $\mathcal{U}_{\bm{x}}$} and corresponding optimal budgets $N^*_{\bm{x}}$ in a single systems $(\pi,\mathcal{D})$ under different aggregation strategies, along with $N^*_{\mathcal{D}}$. Here $\pi = \texttt{Qwen3-4B}$ and $\mathcal{D} = \texttt{AMC}$.}
    \label{fig:distribution-qwen3-amc}
\end{figure}

\begin{figure}[H]
    \vspace{-0.2in}
    \centering
    \includegraphics[width=\columnwidth]{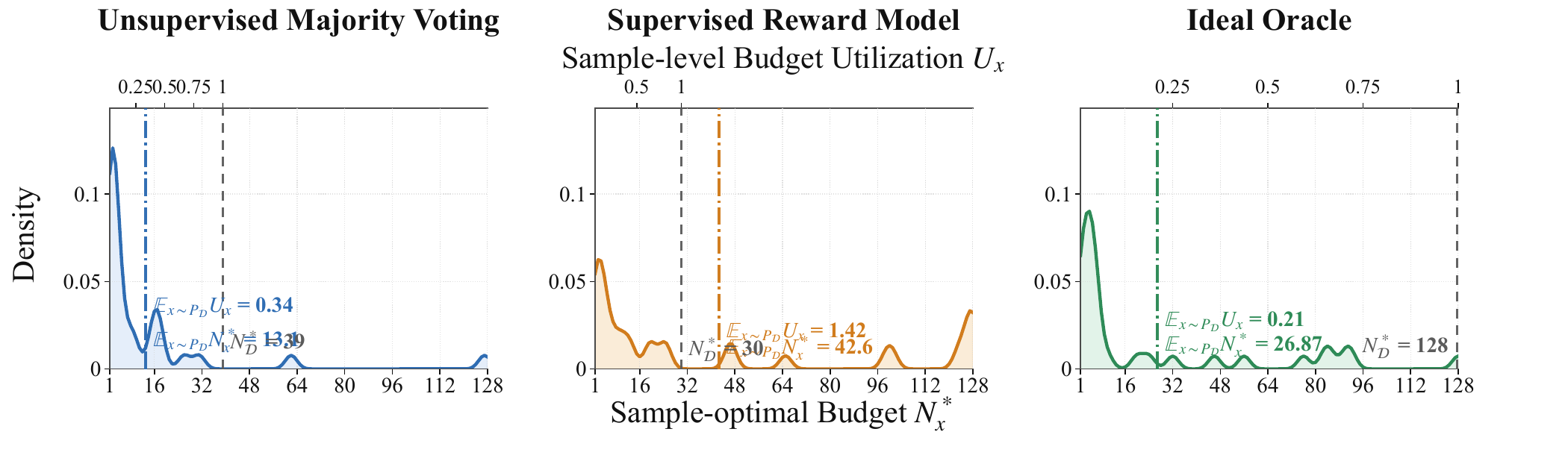}
    \vspace{-0.35in}
    \caption{Distribution of {\it Sample-level Budget Utilization $\mathcal{U}_{\bm{x}}$} and corresponding optimal budgets $N^*_{\bm{x}}$ in a single systems $(\pi,\mathcal{D})$ under different aggregation strategies, along with $N^*_{\mathcal{D}}$. Here $\pi = \texttt{Qwen3-4B}$ and $\mathcal{D} = \texttt{AIME24}$.}
    \label{fig:distribution-qwen3-aime24}
\end{figure}

\begin{figure}[H]
    \vspace{-0.2in}
    \centering
    \includegraphics[width=\columnwidth]{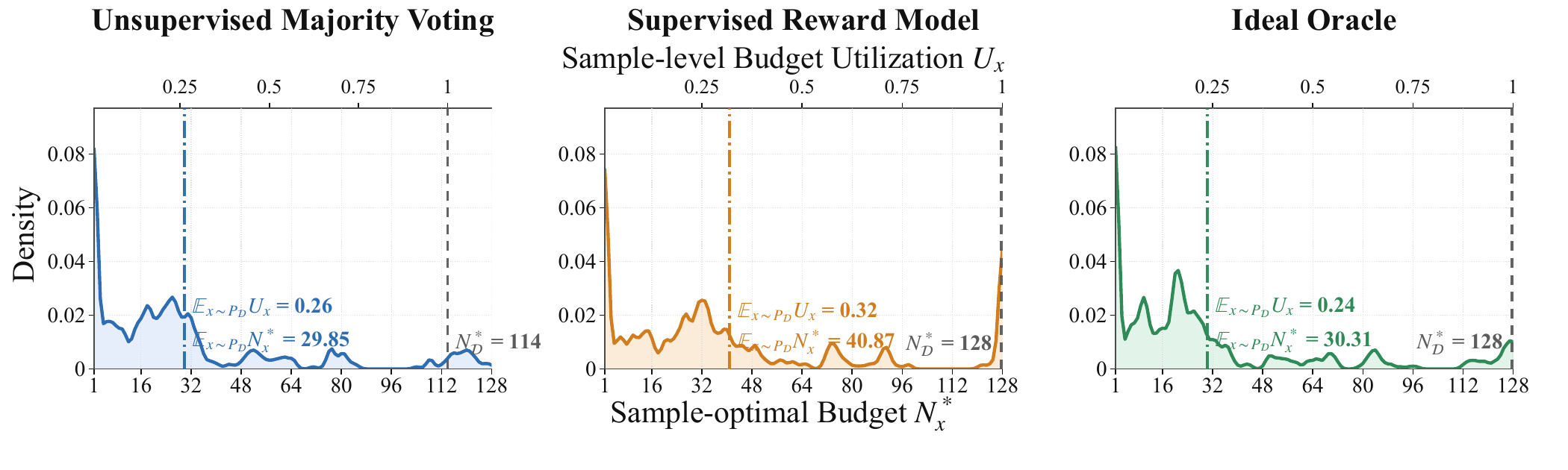}
    \vspace{-0.35in}
    \caption{Distribution of {\it Sample-level Budget Utilization $\mathcal{U}_{\bm{x}}$} and corresponding optimal budgets $N^*_{\bm{x}}$ in a single systems $(\pi,\mathcal{D})$ under different aggregation strategies, along with $N^*_{\mathcal{D}}$. Here $\pi = \texttt{Qwen3-4B}$ and $\mathcal{D} = \texttt{BrowseComp}$.}
    \label{fig:distribution-qwen3-browsecomp}
\end{figure}

\textcolor{deepred}{Figures \ref{fig:distribution-qwen25-math} -- \ref{fig:distribution-qwen3-aime24}} present all the distributions of budget utilization and corresponding sample-optimal budgets, serving as a complete supplement to \textcolor{deepred}{Figure \ref{fig:distribution-qwen3-browsecomp}}.

\begin{figure}[H]
    \vspace{-0.2in}
    \centering
    \includegraphics[width=\columnwidth]{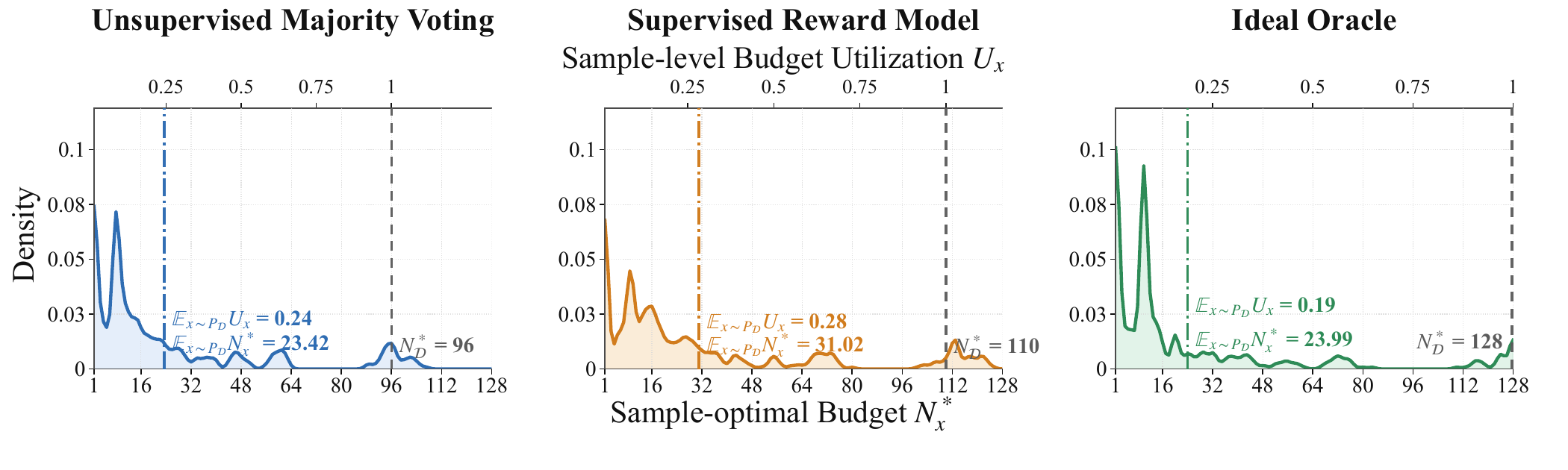}
    \vspace{-0.35in}
    \caption{Distribution of {\it Sample-level Budget Utilization $\mathcal{U}_{\bm{x}}$} and corresponding optimal budgets $N^*_{\bm{x}}$ in a single systems $(\pi,\mathcal{D})$ under different aggregation strategies, along with $N^*_{\mathcal{D}}$. Here $\pi = \texttt{Qwen3-4B}$ and $\mathcal{D} = \texttt{GPQA}$.}
    \label{fig:distribution-qwen3-gpqa}
\end{figure}

\begin{figure}[H]
    \vspace{-0.2in}
    \centering
    \includegraphics[width=\columnwidth]{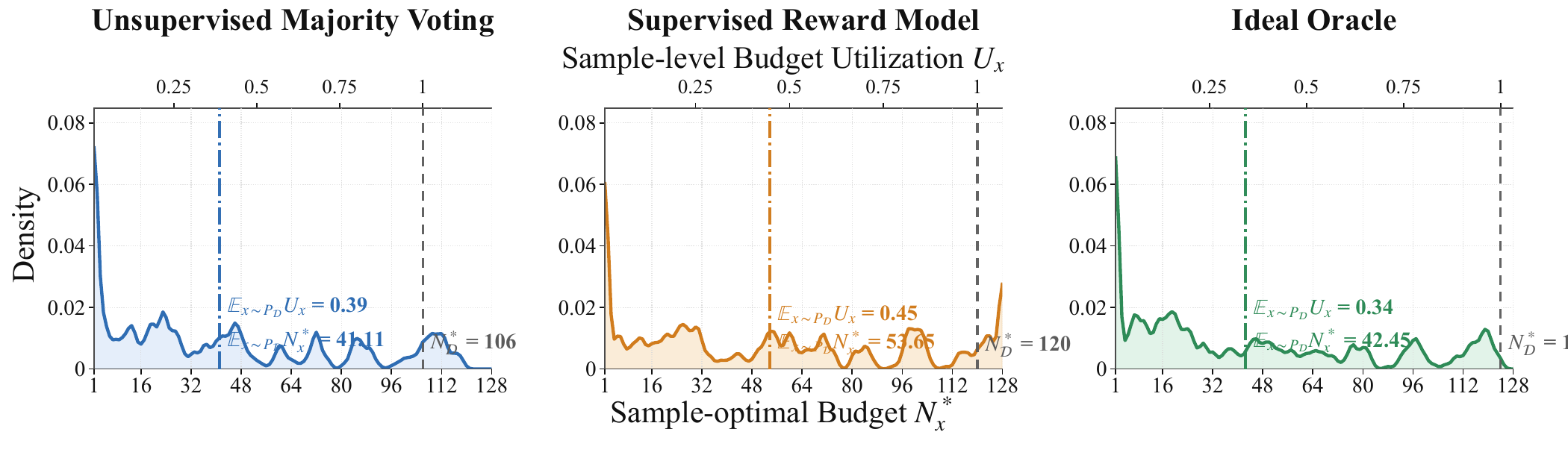}
    \vspace{-0.35in}
    \caption{Distribution of {\it Sample-level Budget Utilization $\mathcal{U}_{\bm{x}}$} and corresponding optimal budgets $N^*_{\bm{x}}$ in a single systems $(\pi,\mathcal{D})$ under different aggregation strategies, along with $N^*_{\mathcal{D}}$. Here $\pi = \texttt{Qwen3-4B}$ and $\mathcal{D} = \texttt{MMLU-Pro}$.}
    \label{fig:distribution-qwen3-mmlu}
\end{figure}

%% file: 8C-overscaling-analysis.tex
\section{Better Understanding the Overscaling Curse}

\subsection{Sample Type Classification}
\label{appe:sample-type}

We classify samples according to the monotonicity pattern of their sample-level budget-accuracy functions $\mathcal{A}_{\bm{x}}(N)$.

\begin{definition}[{\bf Approximate Monotonicity of $\mathcal{A}_{\bm{x}}(N)$}]
Let $s:=\lfloor \sqrt{N_{\max}} \rfloor$ be the step size for monotonicity checking.
For $\alpha \in \{+1,-1\}$, we say that $\mathcal{A}_{\bm{x}}(N)$ is \emph{approximately monotone} in direction $\alpha$, denoted by
$\mathcal{A}_{\bm{x}}(N)\approx_{\alpha}\mathrm{mono}$, if
\begin{equation}
\label{eq:approx-mono}
\frac{1}{N_{\max}-s}
\sum_{i=1}^{N_{\max}-s}
\mathbb{I}
\left\{
\alpha \left(
\mathcal{A}_{\bm{x}}(i+s)-\mathcal{A}_{\bm{x}}(i)
\right)
\ge 0
\right\}
\ge 0.80 .
\end{equation}
Here, $\mathcal{A}_{\bm{x}}(N)\approx_{+1}\mathrm{mono}$ denotes approximate increasing monotonicity, while
$\mathcal{A}_{\bm{x}}(N)\approx_{-1}\mathrm{mono}$ denotes approximate decreasing monotonicity.
\end{definition}
\vspace{-0.08in}

\begin{figure}[H]
    \vspace{-0.in}
    \centering
    \includegraphics[width=1\columnwidth]{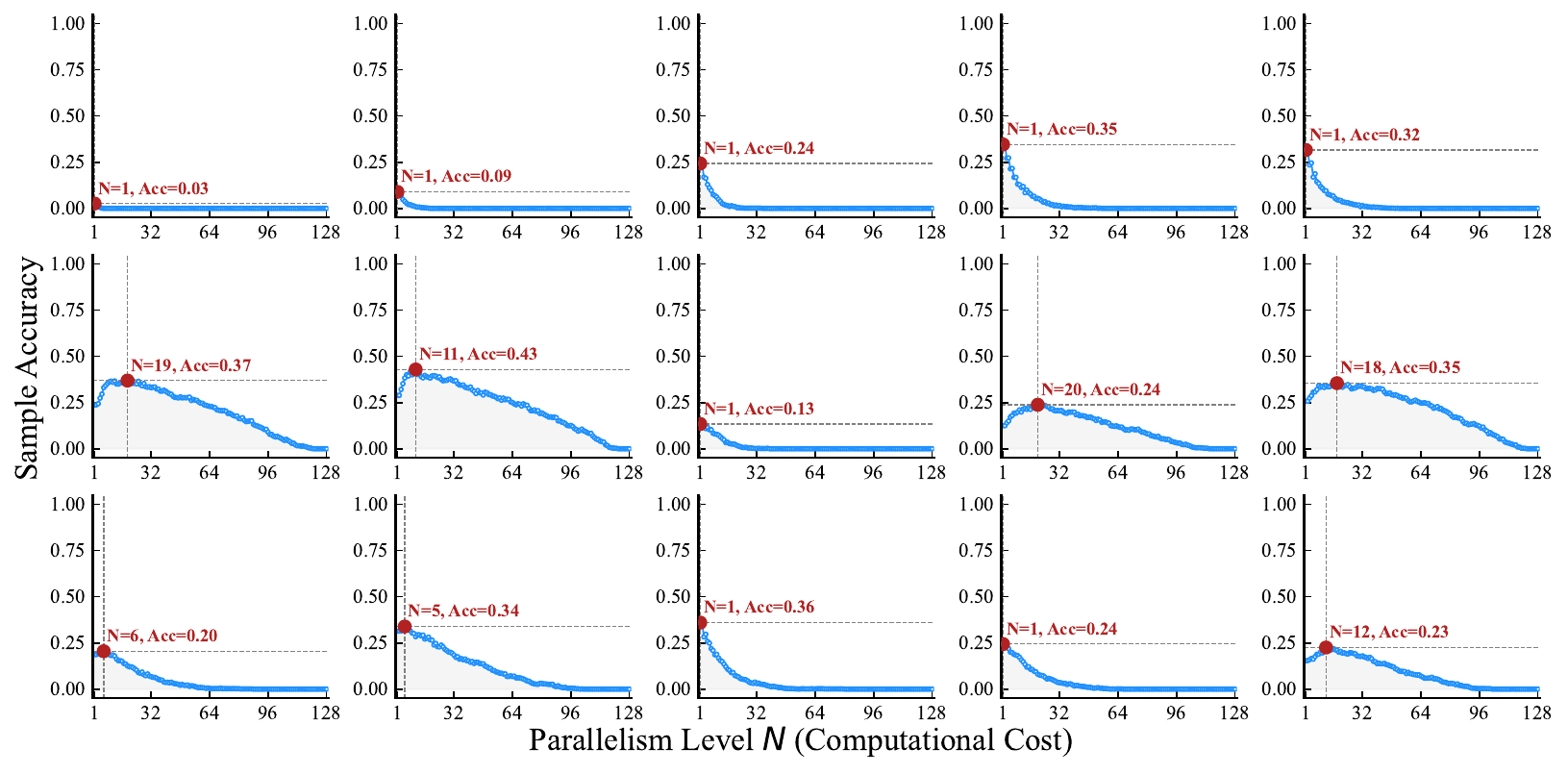}
    \vspace{-0.25in}
    \caption{Examples of the ``budget-accuracy'' function $\mathcal{A}_{\bm{x}}(N)$ for {\bf Type-(3)} samples from \texttt{Qwen2.5-7B} on the \texttt{MATH500} dataset.}
    \vspace{-0.in}
    \label{fig:overscaling-sample-qwen25-math-type3}
\end{figure}

\begin{figure}[H]
    \vspace{-0.in}
    \centering
    \includegraphics[width=1\columnwidth]{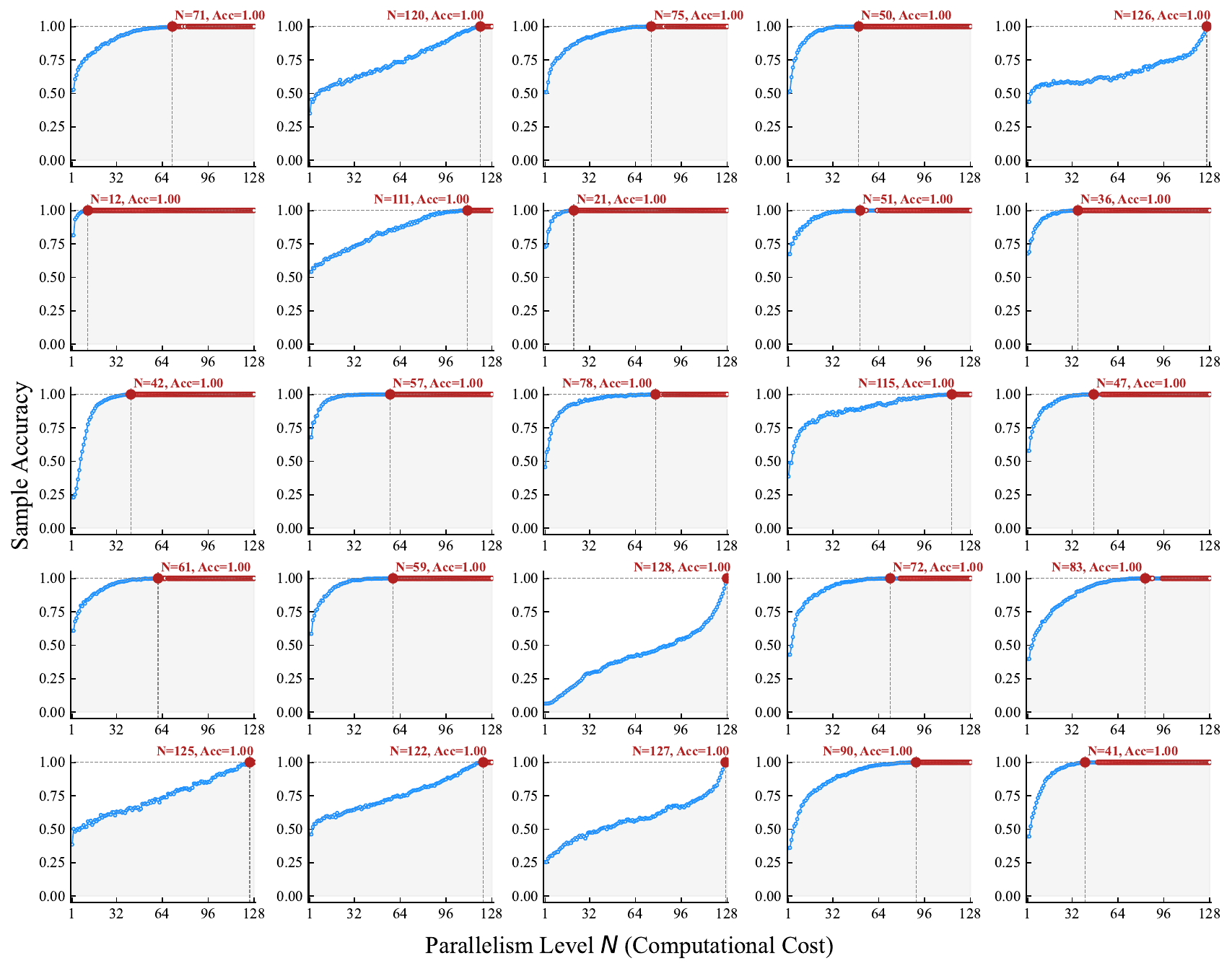}
    \vspace{-0.25in}
    \caption{Examples of the ``budget-accuracy'' function $\mathcal{A}_{\bm{x}}(N)$ for {\bf Type-(4)} samples from \texttt{Qwen2.5-7B} on the \texttt{MATH500} dataset.}
    \vspace{-0.in}
    \label{fig:overscaling-sample-qwen25-math-type4}
\end{figure}

\begin{figure}[H]
    \vspace{-0.in}
    \centering
    \includegraphics[width=1\columnwidth]{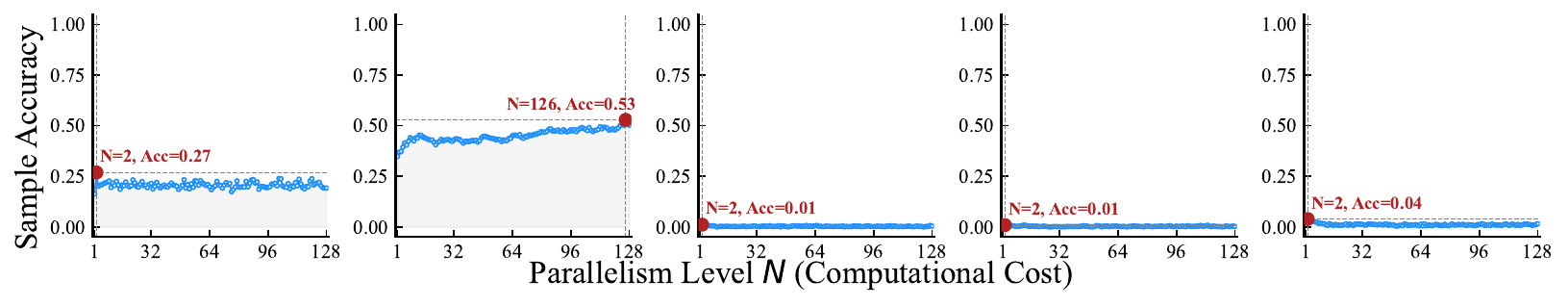}
    \vspace{-0.25in}
    \caption{Examples of the ``budget-accuracy'' function $\mathcal{A}_{\bm{x}}(N)$ for {\bf Type-(5)} samples from \texttt{Qwen2.5-7B} on the \texttt{MATH500} dataset.}
    \vspace{-0.in}
    \label{fig:overscaling-sample-qwen25-math-type5}
\end{figure}

\begin{figure}[H]
    \vspace{-0.in}
    \centering
    \includegraphics[width=1\columnwidth]{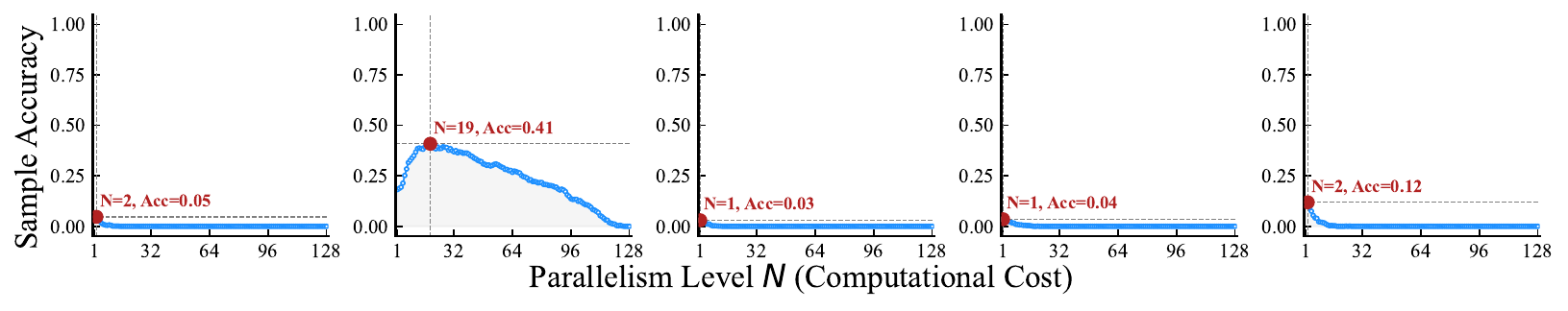}
    \vspace{-0.25in}
    \caption{Examples of the ``budget-accuracy'' function $\mathcal{A}_{\bm{x}}(N)$ for {\bf Type-(3)} samples from \texttt{Qwen2.5-7B} on the \texttt{AIME24} dataset.}
    \vspace{-0.in}
    \label{fig:overscaling-sample-qwen25-aime24-type3}
\end{figure}

\begin{figure}[H]
    \vspace{-0.in}
    \centering
    \includegraphics[width=1\columnwidth]{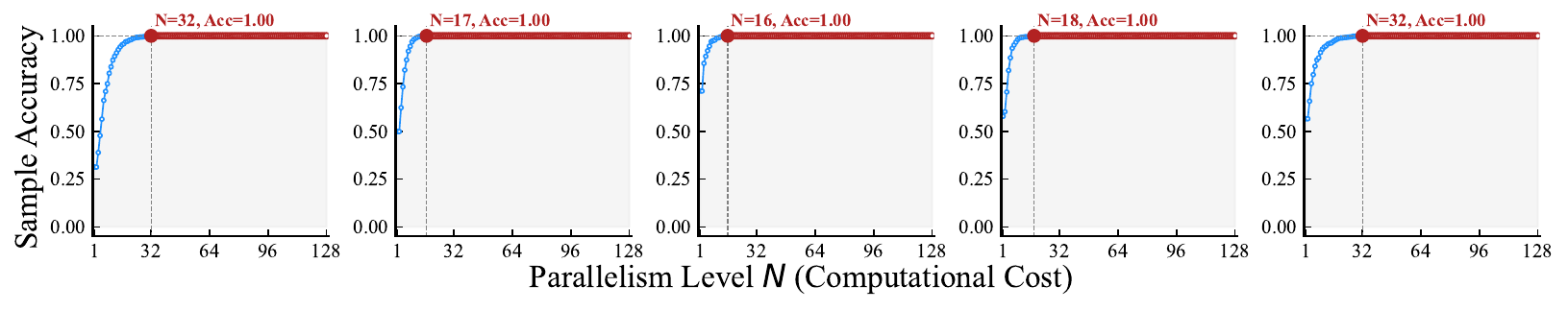}
    \vspace{-0.25in}
    \caption{Examples of the ``budget-accuracy'' function $\mathcal{A}_{\bm{x}}(N)$ for {\bf Type-(4)} samples from \texttt{Qwen2.5-7B} on the \texttt{AIME24} dataset.}
    \vspace{-0.in}
    \label{fig:overscaling-sample-qwen25-aime24-type4}
\end{figure}

\begin{figure}[H]
    \vspace{-0.in}
    \centering
    \includegraphics[width=1\columnwidth]{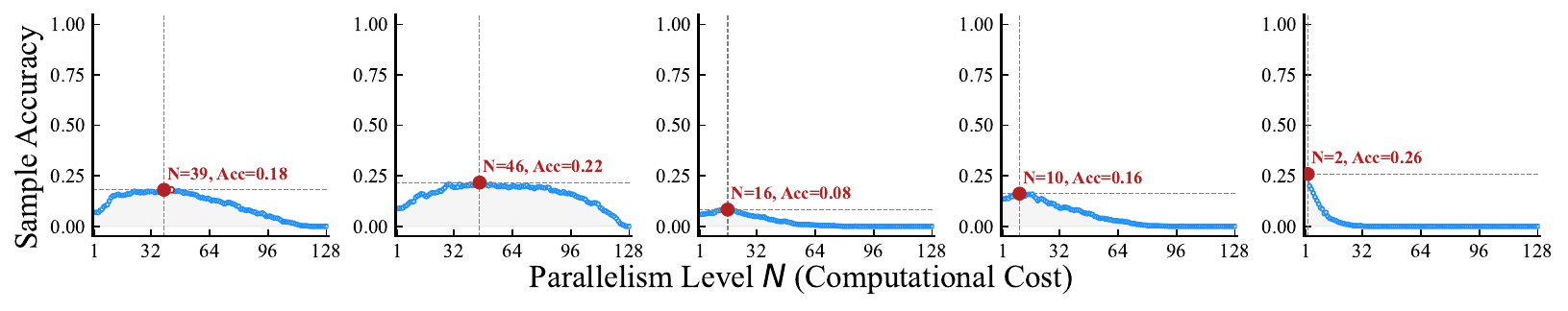}
    \vspace{-0.25in}
    \caption{Examples of the ``budget-accuracy'' function $\mathcal{A}_{\bm{x}}(N)$ for {\bf Type-(3)} samples from \texttt{Qwen2.5-7B} on the \texttt{AIME25} dataset.}
    \vspace{-0.in}
    \label{fig:overscaling-sample-qwen25-aime25-type3}
\end{figure}

\begin{figure}[H]
    \vspace{-0.in}
    \centering
    \includegraphics[width=0.6\columnwidth]{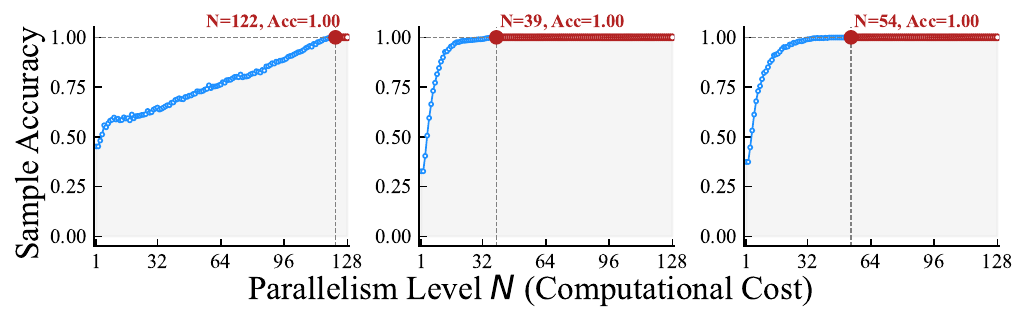}
    \vspace{-0.1in}
    \caption{Examples of the ``budget-accuracy'' function $\mathcal{A}_{\bm{x}}(N)$ for {\bf Type-(4)} samples from \texttt{Qwen2.5-7B} on the \texttt{AIME25} dataset.}
    \vspace{-0.in}
    \label{fig:overscaling-sample-qwen25-aime25-type4}
\end{figure}

\begin{figure}[H]
    \vspace{-0.in}
    \centering
    \includegraphics[width=0.4\columnwidth]{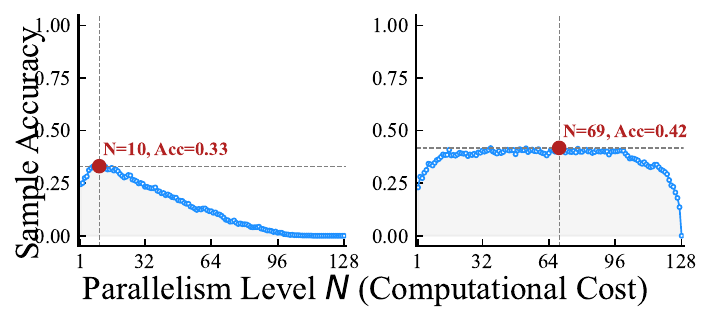}
    \vspace{-0.1in}
    \caption{Examples of the ``budget-accuracy'' function $\mathcal{A}_{\bm{x}}(N)$ for {\bf Type-(3)} samples from \texttt{Qwen3-4B} on the \texttt{AIME24} dataset.}
    \vspace{-0.in}
    \label{fig:overscaling-sample-qwen3-aime24-type3}
\end{figure}

\begin{figure}[H]
    \vspace{-0.in}
    \centering
    \includegraphics[width=1\columnwidth]{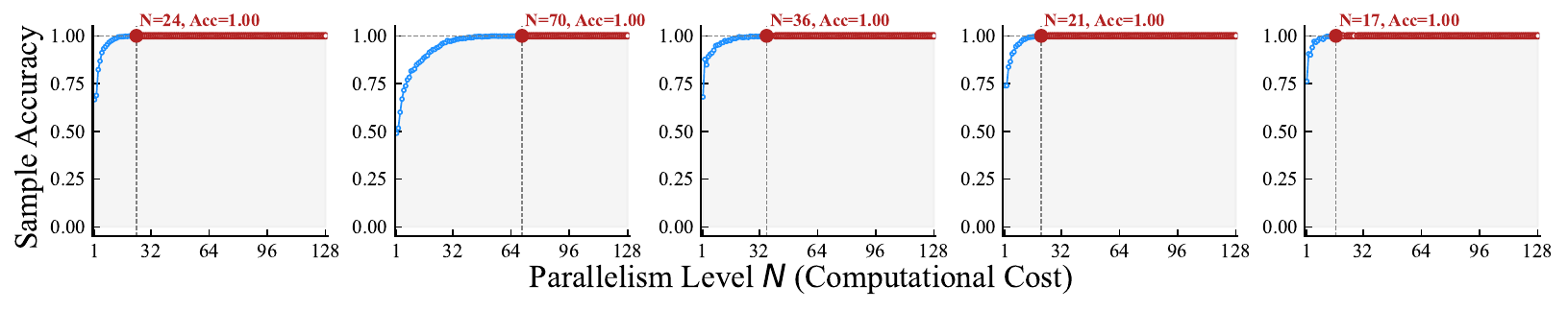}
    \vspace{-0.1in}
    \caption{Examples of the ``budget-accuracy'' function $\mathcal{A}_{\bm{x}}(N)$ for {\bf Type-(4)} samples from \texttt{Qwen3-4B} on the \texttt{AIME24} dataset.}
    \vspace{-0.in}
    \label{fig:overscaling-sample-qwen3-aime24-type4}
\end{figure}

\begin{figure}[H]
    \vspace{-0.in}
    \centering
    \includegraphics[width=0.6\columnwidth]{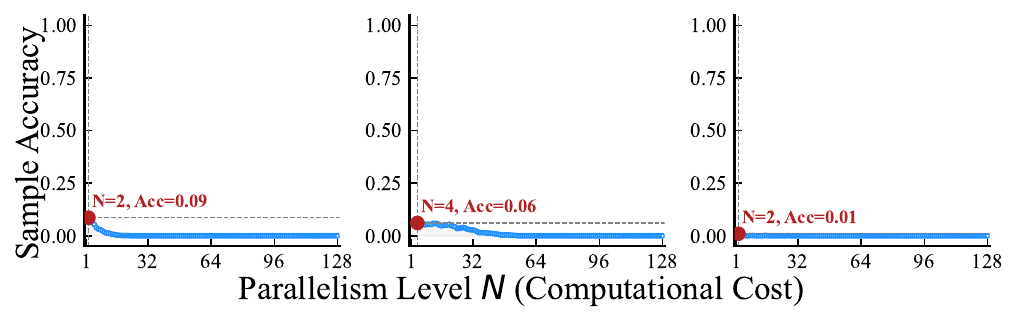}
    \vspace{-0.1in}
    \caption{Examples of the ``budget-accuracy'' function $\mathcal{A}_{\bm{x}}(N)$ for {\bf Type-(3)} samples from \texttt{Qwen3-4B} on the \texttt{AIME25} dataset.}
    \vspace{-0.in}
    \label{fig:overscaling-sample-qwen3-aime25-type3}
\end{figure}

\begin{figure}[H]
    \vspace{-0.in}
    \centering
    \includegraphics[width=1\columnwidth]{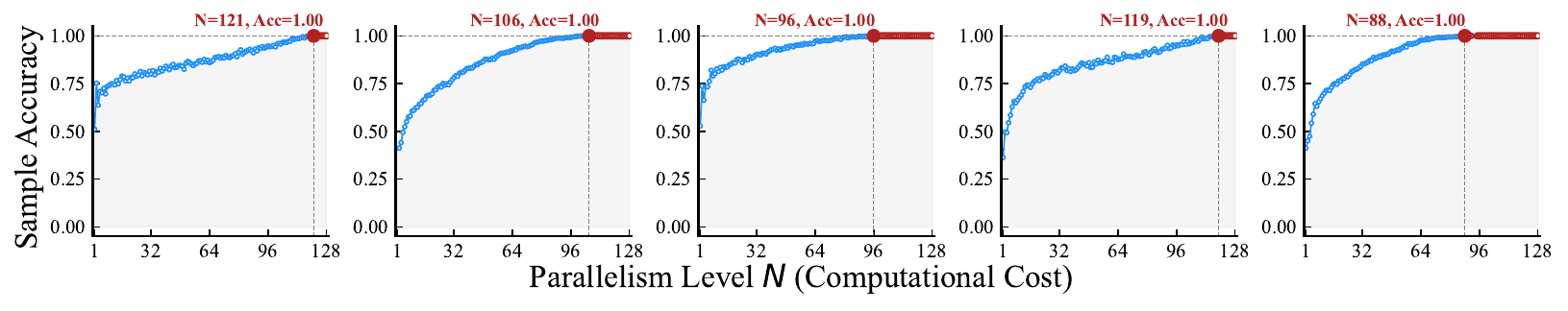}
    \vspace{-0.1in}
    \caption{Examples of the ``budget-accuracy'' function $\mathcal{A}_{\bm{x}}(N)$ for {\bf Type-(4)} samples from \texttt{Qwen3-4B} on the \texttt{AIME25} dataset.}
    \vspace{-0.in}
    \label{fig:overscaling-sample-qwen3-aime25-type4}
\end{figure}

\begin{figure}[H]
    \vspace{-0.in}
    \centering
    \includegraphics[width=0.6\columnwidth]{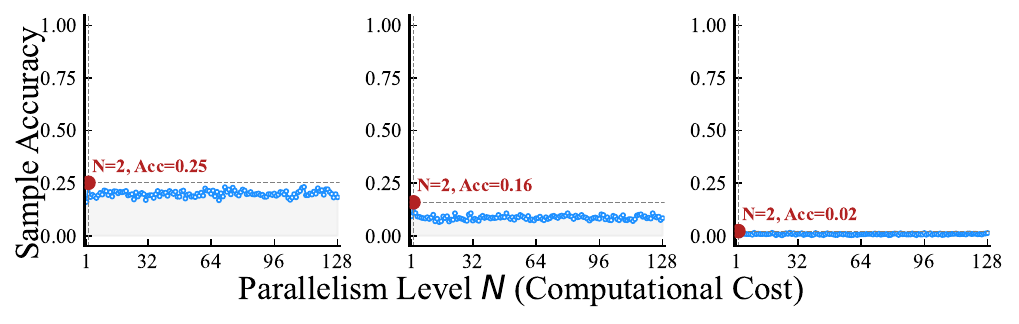}
    \vspace{-0.1in}
    \caption{Examples of the ``budget-accuracy'' function $\mathcal{A}_{\bm{x}}(N)$ for {\bf Type-(5)} samples from \texttt{Qwen3-4B} on the \texttt{AIME25} dataset.}
    \vspace{-0.in}
    \label{fig:overscaling-sample-qwen3-aime25-type5}
\end{figure}

Based on this definition, we partition samples into five types, as summarized in \textcolor{deepred}{Table \ref{tab:classification}}.
Representative examples are shown in \textcolor{deepred}{Figures \ref{fig:overscaling-sample-qwen25-math-type3} -- \ref{fig:overscaling-sample-qwen3-aime25-type5}}.

\begin{table}[H]
\caption{
Sample type classification according to the monotonicity pattern of $\mathcal{A}_{\bm{x}}(N)$.
The five subsets form a partition of $\mathcal{D}$, \emph{i.e.}, $\mathcal{D}_i \cap \mathcal{D}_j=\varnothing$ for $i\neq j$ and $\bigcup_{i=1}^{5}\mathcal{D}_i=\mathcal{D}$.
}
\vspace{0.05in}
\centering
\footnotesize
\renewcommand\arraystretch{1.1}
\setlength{\tabcolsep}{2mm}{
\resizebox{1\textwidth}{!}{
\begin{tabular}{ccp{5.5cm}p{8.5cm}}
\toprule
Type ID & Subset & $\mathcal{A}_{\bm{x}}(N)$ Pattern & Sample Feature \\
\midrule

1 & $\mathcal{D}_1$
& \textcolor{deepred}{$\mathcal{A}_{\bm{x}}(N) \equiv 1$}
& The sample is always solved correctly under all budgets. \\

\midrule
2 & $\mathcal{D}_2$
& \textcolor{deepred}{$\mathcal{A}_{\bm{x}}(N) \equiv 0$}
& The sample is never solved correctly under any budget. \\

\midrule
3 & $\mathcal{D}_3$
& \textcolor{deepred}{$\mathcal{A}_{\bm{x}}(N)\approx_{-1}\mathrm{mono}$, with $\mathcal{A}_{\bm{x}}(N)\not\equiv 1$ and $\mathcal{A}_{\bm{x}}(N)\not\equiv 0$}
& Accuracy tends to decrease as the budget increases. This can occur when incorrect answers become more dominant under aggregation. \\

\midrule
4 & $\mathcal{D}_4$
& \textcolor{deepred}{$\mathcal{A}_{\bm{x}}(N)\approx_{+1}\mathrm{mono}$, with $\mathcal{A}_{\bm{x}}(N)\not\equiv 1$ and $\mathcal{A}_{\bm{x}}(N)\not\equiv 0$}
& Accuracy tends to increase as the budget increases. This indicates that larger budgets make the correct answer more likely to be selected. \\

\midrule
5 & $\mathcal{D}_5$
& \textcolor{deepred}{No clear monotonicity}
& None of the above conditions hold, indicating that the sample does not exhibit a stable monotonic trend over budgets. \\

\bottomrule
\end{tabular}
}
}
\label{tab:classification}
\end{table}

\subsection{Empirical Statistics}
\label{appe:sample-type-statistics}

We report the $\mathbb{E}_{\bm{x}\sim P_{\mathcal{D}_i}}[N_{\bm{x}}^*]$, $p_i$, and $\bar{\Delta}_i$ of each sample type subset $\mathcal{D}_i \subset \mathcal{D}$ under all systems with the majority voting aggregation strategy.

\textcolor{deepred}{Tables \ref{tab:type-nstar-qwen25} -- \ref{tab:type-nstar-qwen3}} show the $\mathbb{E}_{\bm{x}\sim P_{\mathcal{D}_i}}[N_{\bm{x}}^*]$ results.
Across systems, Type-4 consistently shows much larger expected $N_{\bm{x}}^*$ than the other types.

\textcolor{deepred}{Tables \ref{tab:type-gain-math500} -- \ref{tab:type-gain-aime25}} show the results of $p_i$, and $\bar{\Delta}_i$. These systems all suffer from severe overscaling curse.
Across them, $p_4 > p_3$ and $\bar{\Delta}_4 > [-\bar{\Delta}_3]_+$, where $[z]_+=\max(z,0)$, usually happen; $p_5$ and $\bar{\Delta}_5$ is very small that can negligible.

\begin{table}[H]
\vspace{-0in}
\caption{Expected sample-optimal budget $\mathbb{E}[N_{\bm{x}}^*]$ of each sample type under majority voting. Results are from \texttt{Qwen2.5-7B-Instruct}. ``--'' indicates the absence of such a sample type.}
\vspace{0.05in}
\centering
\footnotesize
\renewcommand\arraystretch{1.}
\setlength{\tabcolsep}{1.5mm}{
\resizebox{1\columnwidth}{!}{
\begin{tabular}{lccccccc}
\toprule
\textbf{Type} 
& \textbf{MATH500} 
& \textbf{AMC} 
& \textbf{AIME24} 
& \textbf{AIME25} 
& \textbf{GPQA} 
& \textbf{MMLU-Pro} 
& \textbf{BrowseComp} \\
\midrule
Type-1: Constant-1 
& 1.0 & 1.0 & -- & 1.0 & 1.0 & 1.0 & 1.0 \\
Type-2: Constant-0 
& 1.0 & 1.0 & 1.0 & 1.0 & 1.0 & 1.0 & 1.0 \\
Type-3: Decreasing 
& 3.0 & 3.4 & 1.4 & 3.0 & 2.4 & 5.5 & 6.5 \\
\textbf{Type-4: Increasing} 
& 23.6 & 33.8 & 39.8 & 71.0 & 43.6 & 44.5 & 35.8 \\
Type-5: Non-monotonic 
& 10.9 & 4.5 & 18.0 & 8.0 & 12.9 & 11.4 & 1.4 \\
\bottomrule
\end{tabular}%
}}
\label{tab:type-nstar-qwen25}
\vspace{-0.in}
\end{table}

\begin{table}[H]
\vspace{-0in}
\caption{Expected sample-optimal budget $\mathbb{E}[N_{\bm{x}}^*]$ of each sample type under majority voting. Results are from \texttt{Llama-3.1-8B-Instruct}. ``--'' indicates the absence of such a sample type.}
\vspace{0.05in}
\centering
\footnotesize
\renewcommand\arraystretch{1.}
\setlength{\tabcolsep}{1.5mm}{
\resizebox{1\columnwidth}{!}{
\begin{tabular}{lccccccc}
\toprule
\textbf{Type} 
& \textbf{MATH500} 
& \textbf{AMC} 
& \textbf{AIME24} 
& \textbf{AIME25} 
& \textbf{GPQA} 
& \textbf{MMLU-Pro} 
& \textbf{BrowseComp} \\
\midrule
Type-1: Constant-1 
& 1.0 & -- & -- & -- & 1.0 & 1.0 & 1.0 \\
Type-2: Constant-0 
& 1.0 & 1.0 & 1.0 & 1.0 & 1.0 & 1.0 & 1.0 \\
Type-3: Decreasing 
& 4.8 & 3.1 & -- & 1.0 & 3.5 & 9.9 & 1.1 \\
\textbf{Type-4: Increasing} 
& 32.0 & 59.6 & 82.0 & -- & 90.2 & 67.5 & 35.7 \\
Type-5: Non-monotonic 
& 15.6 & 44.0 & 50.0 & 19.2 & 35.6 & 2.2 & 4.0 \\
\bottomrule
\end{tabular}%
}}
\label{tab:type-nstar-llama31}
\vspace{-0.in}
\end{table}

\begin{table}[H]
\vspace{-0in}
\caption{Expected sample-optimal budget $\mathbb{E}[N_{\bm{x}}^*]$ of each sample type under majority voting. Results are from \texttt{DeepSeek-R1-Distill-Qwen-7B}. ``--'' indicates the absence of such a sample type.}
\vspace{0.05in}
\centering
\footnotesize
\renewcommand\arraystretch{1.}
\setlength{\tabcolsep}{1.5mm}{
\resizebox{1\columnwidth}{!}{
\begin{tabular}{lccccccc}
\toprule
\textbf{Type} 
& \textbf{MATH500} 
& \textbf{AMC} 
& \textbf{AIME24} 
& \textbf{AIME25} 
& \textbf{GPQA} 
& \textbf{MMLU-Pro} 
& \textbf{BrowseComp} \\
\midrule
Type-1: Constant-1 
& 1.0 & 1.0 & 1.0 & 1.0 & 1.0 & 1.0 & 1.0 \\
Type-2: Constant-0 
& 1.0 & 1.0 & 1.0 & 1.0 & 1.0 & 1.0 & 1.0 \\
Type-3: Decreasing 
& 6.5 & 5.1 & 9.2 & 7.8 & 1.3 & 2.4 & 3.8 \\
\textbf{Type-4: Increasing} 
& 22.4 & 75.6 & 88.2 & 35.7 & 48.9 & 62.3 & 36.5 \\
Type-5: Non-monotonic 
& 15.3 & 8.6 & 57.1 & 12.8 & 21.4 & 13.9 & 1.2 \\
\bottomrule
\end{tabular}%
}}
\label{tab:type-nstar-r1}
\vspace{-0.in}
\end{table}

\begin{table}[H]
\vspace{-0in}
\caption{Expected sample-optimal budget $\mathbb{E}[N_{\bm{x}}^*]$ of each sample type under majority voting. Results are from \texttt{Qwen3-4B-Instruct-2507}. ``--'' indicates the absence of such a sample type.}
\vspace{0.05in}
\centering
\footnotesize
\renewcommand\arraystretch{1.}
\setlength{\tabcolsep}{1.5mm}{
\resizebox{1\columnwidth}{!}{
\begin{tabular}{lccccccc}
\toprule
\textbf{Type} 
& \textbf{MATH500} 
& \textbf{AMC} 
& \textbf{AIME24} 
& \textbf{AIME25} 
& \textbf{GPQA} 
& \textbf{MMLU-Pro} 
& \textbf{BrowseComp} \\
\midrule
Type-1: Constant-1 
& 1.0 & 1.0 & 1.0 & 1.0 & 1.0 & 1.0 & 1.0 \\
Type-2: Constant-0 
& 1.0 & 1.0 & 1.0 & 1.0 & 1.0 & 1.0 & 1.0 \\
Type-3: Decreasing 
& 1.3 & 1.3 & 10.0 & 6.3 & 3.4 & 5.5 & 2.7 \\
\textbf{Type-4: Increasing} 
& 52.1 & 17.7 & 22.0 & 49.5 & 55.4 & 76.1 & 34.6 \\
Type-5: Non-monotonic 
& 14.2 & 16.6 & 2.0 & 23.3 & 16.6 & 12.1 & 25.7 \\
\bottomrule
\end{tabular}%
}}
\label{tab:type-nstar-qwen3}
\vspace{-0.in}
\end{table}

\begin{table}[H]
\vspace{-0in}
\caption{Sample proportion $p_i$ and expected total accuracy gain $\bar{\Delta}_i$ of each sample type on \texttt{MATH500} under majority voting. ``--'' indicates the absence of such a sample type.}
\vspace{0.05in}
\centering
\footnotesize
\renewcommand\arraystretch{1.}
\setlength{\tabcolsep}{1.8mm}{
\resizebox{1\columnwidth}{!}{
\begin{tabular}{lcccccccc}
\toprule
\multirow{2}{*}{\textbf{Type}}
& \multicolumn{2}{c}{\textbf{Qwen2.5-7B}}
& \multicolumn{2}{c}{\textbf{Llama3.1-8B}}
& \multicolumn{2}{c}{\textbf{R1-Distill-Qwen-7B}}
& \multicolumn{2}{c}{\textbf{Qwen3-4B}}
\\
\cmidrule(lr){2-3}
\cmidrule(lr){4-5}
\cmidrule(lr){6-7}
\cmidrule(lr){8-9}
& $p_i$ & $\bar{\Delta}_i$
& $p_i$ & $\bar{\Delta}_i$
& $p_i$ & $\bar{\Delta}_i$
& $p_i$ & $\bar{\Delta}_i$ \\
\midrule
Type-1: Constant-1      & 0.39 & 0.00 & 0.05 & 0.00 & 0.56 & 0.00 & 0.63 & 0.00 \\
Type-2: Constant-0      & 0.08 & 0.00 & 0.32 & 0.00 & 0.13 & 0.00 & 0.04 & 0.00 \\
Type-3: Decreasing      & 0.11 & -0.12 & 0.12 & -0.14 & 0.03 & -0.04 & 0.03 & -0.08 \\
\textbf{Type-4: Increasing} & 0.38 & 0.21 & 0.49 & 0.32 & 0.28 & 0.38 & 0.29 & 0.22 \\
Type-5: Non-monotonic   & 0.03 & -0.01 & 0.03 & 0.05 & 0.00 & 0.10 & 0.02 & 0.06 \\
\bottomrule
\end{tabular}%
}}
\label{tab:type-gain-math500}
\vspace{-0.in}
\end{table}

\begin{table}[H]
\vspace{-0in}
\caption{Sample proportion $p_i$ and expected total accuracy gain $\bar{\Delta}_i$ of each sample type on \texttt{AMC} under majority voting. ``--'' indicates the absence of such a sample type.}
\vspace{0.05in}
\centering
\footnotesize
\renewcommand\arraystretch{1.}
\setlength{\tabcolsep}{1.8mm}{
\resizebox{1\columnwidth}{!}{
\begin{tabular}{lcccccccc}
\toprule
\multirow{2}{*}{\textbf{Type}}
& \multicolumn{2}{c}{\textbf{Qwen2.5-7B}}
& \multicolumn{2}{c}{\textbf{Llama3.1-8B}}
& \multicolumn{2}{c}{\textbf{R1-Distill-Qwen-7B}}
& \multicolumn{2}{c}{\textbf{Qwen3-4B}}
\\
\cmidrule(lr){2-3}
\cmidrule(lr){4-5}
\cmidrule(lr){6-7}
\cmidrule(lr){8-9}
& $p_i$ & $\bar{\Delta}_i$
& $p_i$ & $\bar{\Delta}_i$
& $p_i$ & $\bar{\Delta}_i$
& $p_i$ & $\bar{\Delta}_i$ \\
\midrule
Type-1: Constant-1      & 0.09 & 0.00 & -- & -- & 0.46 & 0.00 & 0.30 & 0.00 \\
Type-2: Constant-0      & 0.11 & 0.00 & 0.29 & 0.00 & 0.12 & 0.00 & 0.10 & 0.00 \\
Type-3: Decreasing      & 0.31 & -0.10 & 0.13 & -0.05 & 0.10 & -0.04 & 0.02 & -0.04 \\
\textbf{Type-4: Increasing} & 0.47 & 0.32 & 0.45 & 0.45 & 0.30 & 0.49 & 0.51 & 0.19 \\
Type-5: Non-monotonic   & 0.01 & -0.05 & 0.14 & 0.03 & 0.02 & 0.13 & 0.07 & -0.02 \\
\bottomrule
\end{tabular}%
}}
\label{tab:type-gain-amc}
\vspace{-0.in}
\end{table}

\begin{table}[H]
\vspace{-0in}
\caption{Sample proportion $p_i$ and expected total accuracy gain $\bar{\Delta}_i$ of each sample type on \texttt{AIME24} under majority voting. ``--'' indicates the absence of such a sample type.}
\vspace{0.05in}
\centering
\footnotesize
\renewcommand\arraystretch{1.}
\setlength{\tabcolsep}{1.8mm}{
\resizebox{1\columnwidth}{!}{
\begin{tabular}{lcccccccc}
\toprule
\multirow{2}{*}{\textbf{Type}}
& \multicolumn{2}{c}{\textbf{Qwen2.5-7B}}
& \multicolumn{2}{c}{\textbf{Llama3.1-8B}}
& \multicolumn{2}{c}{\textbf{R1-Distill-Qwen-7B}}
& \multicolumn{2}{c}{\textbf{Qwen3-4B}}
\\
\cmidrule(lr){2-3}
\cmidrule(lr){4-5}
\cmidrule(lr){6-7}
\cmidrule(lr){8-9}
& $p_i$ & $\bar{\Delta}_i$
& $p_i$ & $\bar{\Delta}_i$
& $p_i$ & $\bar{\Delta}_i$
& $p_i$ & $\bar{\Delta}_i$ \\
\midrule
Type-1: Constant-1      & -- & -- & -- & -- & 0.27 & 0.00 & 0.03 & 0.00 \\
Type-2: Constant-0      & 0.53 & 0.00 & 0.53 & 0.00 & 0.20 & 0.00 & 0.10 & 0.00 \\
Type-3: Decreasing      & 0.20 & -0.02 & -- & -- & 0.13 & -0.07 & 0.07 & -0.11 \\
\textbf{Type-4: Increasing} & 0.23 & 0.49 & 0.07 & 0.72 & 0.37 & 0.52 & 0.67 & 0.27 \\
Type-5: Non-monotonic   & 0.03 & 0.00 &  0.40 & 0.02 & 0.03 & 0.00 & 0.13 & 0.00 \\
\bottomrule
\end{tabular}%
}}
\label{tab:type-gain-aime24}
\vspace{-0.in}
\end{table}

\begin{table}[H]
\vspace{-0in}
\caption{Sample proportion $p_i$ and expected total accuracy gain $\bar{\Delta}_i$ of each sample type on \texttt{AIME25} under majority voting. ``--'' indicates the absence of such a sample type.}
\vspace{0.05in}
\centering
\footnotesize
\renewcommand\arraystretch{1.}
\setlength{\tabcolsep}{1.8mm}{
\resizebox{1\columnwidth}{!}{
\begin{tabular}{lcccccccc}
\toprule
\multirow{2}{*}{\textbf{Type}}
& \multicolumn{2}{c}{\textbf{Qwen2.5-7B}}
& \multicolumn{2}{c}{\textbf{Llama3.1-8B}}
& \multicolumn{2}{c}{\textbf{R1-Distill-Qwen-7B}}
& \multicolumn{2}{c}{\textbf{Qwen3-4B}}
\\
\cmidrule(lr){2-3}
\cmidrule(lr){4-5}
\cmidrule(lr){6-7}
\cmidrule(lr){8-9}
& $p_i$ & $\bar{\Delta}_i$
& $p_i$ & $\bar{\Delta}_i$
& $p_i$ & $\bar{\Delta}_i$
& $p_i$ & $\bar{\Delta}_i$ \\
\midrule
Type-1: Constant-1      & -- & -- & -- & -- & 0.20 & 0.00 & 0.07 & 0.00 \\
Type-2: Constant-0      & 0.63 & 0.00 & 0.67 & 0.00 & 0.27 & 0.00 & 0.27 & 0.00 \\
Type-3: Decreasing      & 0.13 & -0.07 & 0.03 & -0.01 & 0.03 & -0.05 & 0.1 & -0.06 \\
\textbf{Type-4: Increasing} & 0.20 & 0.68 & 0.30 & 0.02 & 0.33 & 0.57 & 0.47 & 0.33 \\
Type-5: Non-monotonic   & 0.03 & -0.14 & 0.10 & 0.00 & 0.07 & 0.05 & 0.10 & 0.02 \\
\bottomrule
\end{tabular}%
}}
\label{tab:type-gain-aime25}
\vspace{-0.in}
\end{table}

\subsection{Proof of \textcolor{deepred}{Theorem \ref{thm:type4-dominance}}}
\label{appe:proof}

\begin{proof}
Recall that $\bar{\Delta}_i$ denotes the expected total accuracy gain of $\mathcal{D}_i$ over the full budget domain, i.e.,
\begin{equation}
\bar{\Delta}_i
=
\mathbb{E}_{\bm{x}\sim P_{\mathcal{D}_i}}
\left[
\mathcal{A}_{\bm{x}}(N_{\max})-\mathcal{A}_{\bm{x}}(1)
\right].
\end{equation}
Let $B_4=\lceil \mu_4\rceil$ be the high-budget reference induced by $\mathcal{D}_4$.
For the proof, we introduce the interval-wise total gain from $N$ to $B_4$:
\begin{equation}
\bar{\Delta}_i(N,B_4)
=
\mathcal{A}_{\mathcal{D}_i}(B_4)
-
\mathcal{A}_{\mathcal{D}_i}(N),
\end{equation}
where
\begin{equation}
\mathcal{A}_{\mathcal{D}_i}(N)
=
\mathbb{E}_{\bm{x}\sim P_{\mathcal{D}_i}}
\left[
\mathcal{A}_{\bm{x}}(N)
\right].
\end{equation}

Under the monotonic setting, the full-domain dominance condition
$p_4\bar{\Delta}_4>p_3[-\bar{\Delta}_3]_+$
is assumed to transfer to the high-budget interval before $B_4$:
\begin{equation}
p_4\bar{\Delta}_4(N,B_4)
>
p_3[-\bar{\Delta}_3(N,B_4)]_+,
\qquad
\forall N<B_4.
\end{equation}
This is the interval form of the observed Type-4 dominance: $\mathcal{D}_4$ provides the dominant positive gain, while $\mathcal{D}_3$ provides only a weaker negative pull.
Since $\mathcal{D}_1$ and $\mathcal{D}_2$ have zero total gain and $\mathcal{D}_5$ is negligible in practice, the dataset-level accuracy difference between $B_4$ and any $N<B_4$ can be reduced to
\begin{equation}
\mathcal{A}_{\mathcal{D}}(B_4)
-
\mathcal{A}_{\mathcal{D}}(N)
=
p_3\bar{\Delta}_3(N,B_4)
+
p_4\bar{\Delta}_4(N,B_4).
\end{equation}
Because
\begin{equation}
p_3\bar{\Delta}_3(N,B_4)
\ge
-p_3[-\bar{\Delta}_3(N,B_4)]_+,
\end{equation}
we have
\begin{equation}
p_3\bar{\Delta}_3(N,B_4)
+
p_4\bar{\Delta}_4(N,B_4)
>
0.
\end{equation}
Therefore,
\begin{equation}
\mathcal{A}_{\mathcal{D}}(B_4)
>
\mathcal{A}_{\mathcal{D}}(N),
\qquad
\forall N<B_4.
\end{equation}
Thus, no budget smaller than $B_4$ can be system-optimal, implying
\begin{equation}
N_{\mathcal{D}}^*
\ge
B_4.
\end{equation}
Next, since $\mu_4>\mu_i$ for all $i\neq 4$ and $\sum_i p_i=1$ with $p_4<1$, the dataset-level expected sample-optimal budget satisfies
\begin{equation}
\mathbb{E}_{\bm{x}\sim P_{\mathcal{D}}}
\left[
N_{\bm{x}}^*
\right]
=
\sum_{i=1}^{5}p_i\mu_i
<
\mu_4
\le
B_4.
\end{equation}
Under the global-budget setting, each sample is assigned $N_{\mathcal{D}}^*$, so
\begin{equation}
\mathcal{U}_{\bm{x}}
=
\frac{N_{\bm{x}}^*}{N_{\mathcal{D}}^*}.
\end{equation}
Taking expectation over $\mathcal{D}$ gives
\begin{equation}
\mathbb{E}_{\bm{x}\sim P_{\mathcal{D}}}
\left[
\mathcal{U}_{\bm{x}}
\right]
=
\frac{
\mathbb{E}_{\bm{x}\sim P_{\mathcal{D}}}
\left[
N_{\bm{x}}^*
\right]
}{
N_{\mathcal{D}}^*
}
\le
\frac{
\mathbb{E}_{\bm{x}\sim P_{\mathcal{D}}}
\left[
N_{\bm{x}}^*
\right]
}{
B_4
}
<1.
\end{equation}
This proves the claim.
\end{proof}

%% file: 8D-LanBo-setting.tex
\section{Detailed Settings of \latbp{}}
\label{appe:LBA-setting}

\subsection{Training Data and Hyperparameters}
\label{appe:LBA-implementation}

\paragraph{Training/Validation Data.}

To ensure both difficulty and domain diversity in the training data, we use two sources:
\begin{enumerate}
    \item For difficulty diversity, we adopt the open-source Deepmath-103K corpus \citep{he2025deepmath}, which contains 10 difficulty levels and covers diverse mathematical subdomains. We uniformly sample 250 examples per level, yielding 2,500 examples.
    \item For domain diversity, we adopt the open-source WebInstruct-verified corpus \citep{ma2025general}, which spans nearly 10 domains, including finance, business, history, biology, etc. We uniformly sample from each domain to form another 2,500 examples.
\end{enumerate}
In total, 5,000 examples are used for training.
Similarly, the validation data follow the same selection strategy and size, but with no overlap with the training data.

\paragraph{Training Hyperparameters.}
\textcolor{deepred}{Table \ref{tab:hyper}} presents the training hyperparameters of all predictors.

\begin{table}[H]
\caption{Training hyperparameters.}
\vspace{0.05in}
\centering
\footnotesize
\renewcommand\arraystretch{1.2}
\setlength{\tabcolsep}{4mm}{
\resizebox{0.4\textwidth}{!}{
\begin{tabular}{lc}

\toprule

Batch Size & 128 \\
Epoch & 300 \\
Learning Rate & 1e-3 \\
Weight Decay & 1e-4 \\
Optimizer & AdamW \\

\bottomrule
\end{tabular}}}
\label{tab:hyper}
\end{table}

\subsection{Inference-time Predictor Weighting}
\label{appe:LBA-weighting}

\paragraph{Solution of the Optimal Weights.}

In this part, we provide a detailed derivation of the optimal weights. We first restate the problem:
During inference, given an unseen sample $\bm{x}$, let $\hat{\bm n}_{\bm{x}}^* = (\hat n_{\bm{x},1}^*, \dots, \hat n_{\bm{x},L}^*)^\top$ denote the vector of normalized predictions, and let $\bm{\omega} = (\omega_1, \dots, \omega_L)^\top$ be the weight vector satisfying $\bm 1^\top \bm{\omega} = 1$. The final budget prediction is given by:
\begin{equation}
    \hat N_{\bm{x}}^* = N_{\max} \cdot \bm{\omega}^\top \hat{\bm n}_{\bm{x}}^*.
\end{equation}
Our objective is to minimize the expected squared error of the combined prediction on unseen samples:
\begin{equation}
\min_{\bm{\omega}} \; \mathbb{E}_{\bm{x} \sim P_{\mathcal{D}}} \left[ \left( \bm{\omega}^\top \hat{\bm n}_{\bm{x}}^* - n_{\bm{x}}^* \right)^2 \right] \quad \text{s.t.} \quad \bm 1^\top \bm{\omega} = 1,
\end{equation}
where $n_{\bm{x}}^* = N_{\bm{x}}^* / N_{\max}$ is the normalized optimal budget.

\begin{solution}
Define the error vector $\bm e_{\bm{x}} = \hat{\bm n}_{\bm{x}}^* - n_{\bm{x}}^* \bm 1$. Under the constraint $\bm 1^\top \bm{\omega} = 1$, the prediction error simplifies to:
\begin{equation}
\bm{\omega}^\top \hat{\bm n}_{\bm{x}}^* - n_{\bm{x}}^* = \bm{\omega}^\top (\hat{\bm n}_{\bm{x}}^* - n_{\bm{x}}^* \bm 1) = \bm{\omega}^\top \bm e_{\bm{x}}.
\end{equation}
Thus, the objective becomes:
\begin{equation}
\mathbb{E}_{\bm{x} \sim P_{\mathcal{D}}} \left[ (\bm{\omega}^\top \bm e_{\bm{x}})^2 \right] = \bm{\omega}^\top \Sigma \bm{\omega},
\end{equation}
where $\Sigma = \mathbb{E}_{\bm{x} \sim P_{\mathcal{D}}} [\bm e_{\bm{x}} \bm e_{\bm{x}}^\top]$ is the error covariance matrix across layers. The optimization problem reduces to:
\begin{equation}
\min_{\bm{\omega}} \; \bm{\omega}^\top \Sigma \bm{\omega} \quad \text{s.t.} \quad \bm 1^\top \bm{\omega} = 1.
\end{equation}
We introduce a Lagrange multiplier $\lambda$ and define the Lagrangian:
\begin{equation}
\mathcal{L}(\bm{\omega}, \lambda) = \bm{\omega}^\top \Sigma \bm{\omega} - \lambda (\bm 1^\top \bm{\omega} - 1).
\end{equation}
Taking the derivative with respect to $\bm{\omega}$ and setting it to zero:
\begin{equation}
\frac{\partial \mathcal{L}}{\partial \bm{\omega}} = 2 \Sigma \bm{\omega} - \lambda \bm 1 = 0 \quad \Rightarrow \quad \bm{\omega} = \frac{\lambda}{2} \Sigma^{-1} \bm 1.
\end{equation}
Substituting this into the constraint $\bm 1^\top \bm{\omega} = 1$:
\begin{equation}
\bm 1^\top \left( \frac{\lambda}{2} \Sigma^{-1} \bm 1 \right) = 1 \quad \Rightarrow \quad \frac{\lambda}{2} \cdot \bm 1^\top \Sigma^{-1} \bm 1 = 1 \quad \Rightarrow \quad \lambda = \frac{2}{\bm 1^\top \Sigma^{-1} \bm 1}.
\end{equation}
Plugging $\lambda$ back into the expression for $\bm{\omega}$ yields the optimal weights:
\begin{equation}
\boxed{\bm{\omega}^* = \frac{\Sigma^{-1} \bm 1}{\bm 1^\top \Sigma^{-1} \bm 1}}.
\end{equation}

\end{solution}

\paragraph{Approximation Error Analysis.}

In this part, we analyze the approximation error induced by replacing the full error covariance matrix $\Sigma$ with its diagonal surrogate.
Let $\bm{\varepsilon}=(\varepsilon_1,\dots,\varepsilon_L)^\top$ denote the layer-wise prediction errors and $\Sigma=\mathbb{E}[\bm{\varepsilon}\bm{\varepsilon}^\top]$ their covariance matrix.
We decompose $\Sigma$ as
\begin{equation}
\Sigma = D + R,
\end{equation}
where $D=\mathrm{diag}(\Sigma)$ contains the diagonal elements and $R=\Sigma-D$ contains only the off-diagonal terms.
For any aggregation weight $\bm w$ satisfying $\bm 1^\top\bm w=1$ and $w_l\ge 0$,
the resulting mean squared error can be written as
\begin{equation}
\bm w^\top \Sigma \bm w
=
\bm w^\top D \bm w
+
\bm w^\top R \bm w.
\end{equation}
The second term quantifies the approximation error introduced by ignoring cross-layer correlations.
By Hölder's inequality, we have
\begin{equation}
|\bm w^\top R \bm w|
\le
\|R\|_\infty
:=
\max_i \sum_j |R_{ij}|,
\end{equation}
which yields the bound
\begin{equation}
\bm w^\top \Sigma \bm w
\in
\Big[
\bm w^\top D \bm w - \|R\|_\infty,\;
\bm w^\top D \bm w + \|R\|_\infty
\Big].
\end{equation}
Therefore, the worst-case deviation of the diagonal surrogate from the true aggregation error is upper bounded by $\|R\|_\infty$.
In practice, we estimate the empirical covariance matrix $\hat{\Sigma}$ on the validation set and report the relative off-diagonal energy
\begin{equation}
\frac{\|\hat{R}\|_F}{\|\hat{\Sigma}\|_F},
\qquad
\hat{R}=\hat{\Sigma}-\mathrm{diag}(\hat{\Sigma}),
\end{equation}
as well as the relative MSE deviation
\begin{equation}
\frac{|\bm w^\top \hat{R} \bm w|}{\bm w^\top \hat{D} \bm w}.
\end{equation}
Empirically, both quantities are consistently small, indicating that cross-layer error correlations contribute marginally to the aggregated MSE, and validating the diagonal approximation adopted in our method.

%% file: 8E-LanBo-results.tex
\section{Empirical Validation of Latent Budget Predictor}
\label{appe:latbp}

\begin{figure}[H]
    \centering
    \begin{minipage}[b]{0.48\columnwidth}
        \centering
        \includegraphics[width=\linewidth]{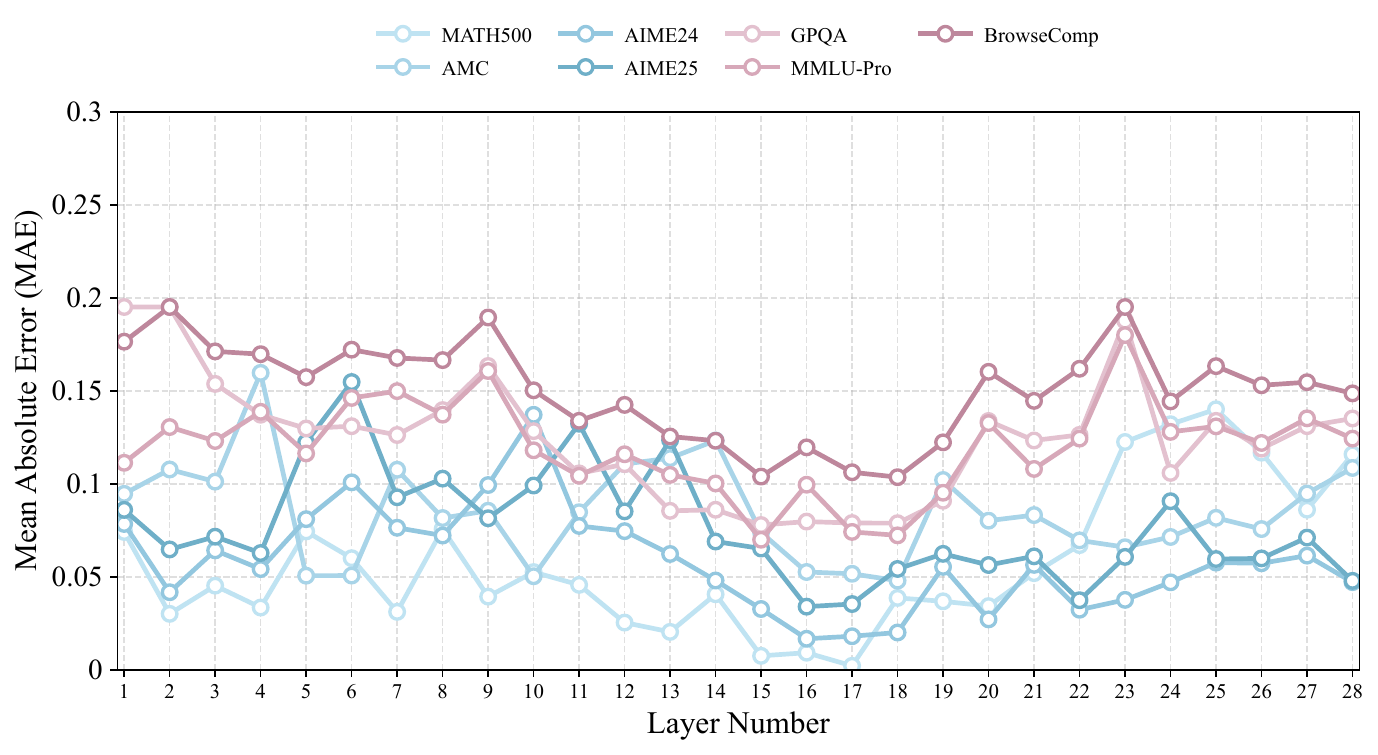}
        \subcaption{MAE results of the $L$ layer-wise predictors on the validation set, with each averaged {\it over 8 runs}.}
    \end{minipage}
    \hfill
    \begin{minipage}[b]{0.48\columnwidth}
        \centering
        \includegraphics[width=\linewidth]{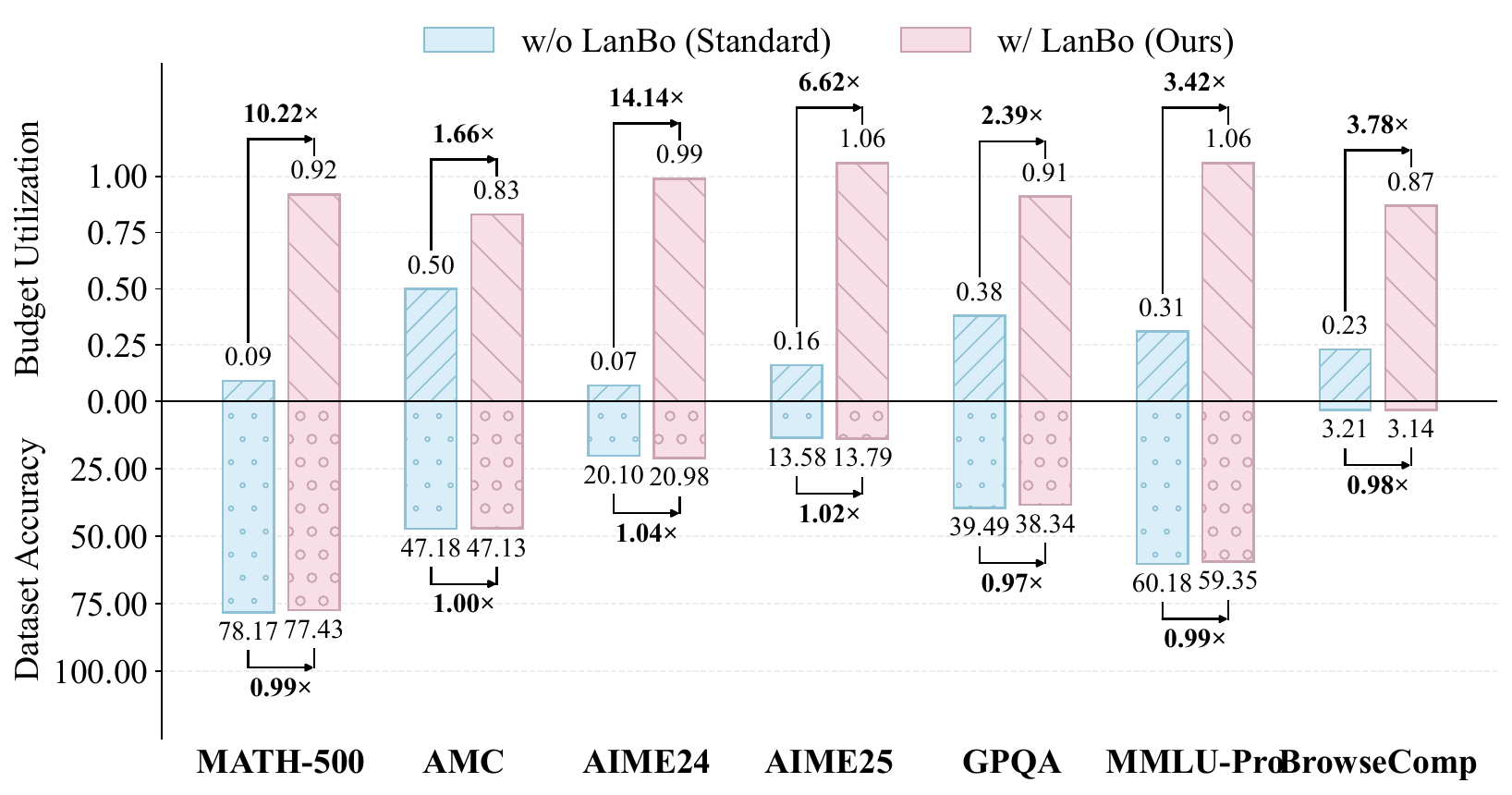}
        \subcaption{Comparison of system-level budget utilization and dataset accuracy before and after applying \latbp.}
    \end{minipage}
    \caption{{\bf Empirical Validation of \latbp~for Breaking the Overscaling Curse}. Here we use \texttt{Qwen2.5-7B} with {\it voting} aggregation.}
    \label{fig:latbp-results-qwen25}
\end{figure}

\begin{figure}[H]
    \centering
    \begin{minipage}[b]{0.48\columnwidth}
        \centering
        \includegraphics[width=\linewidth]{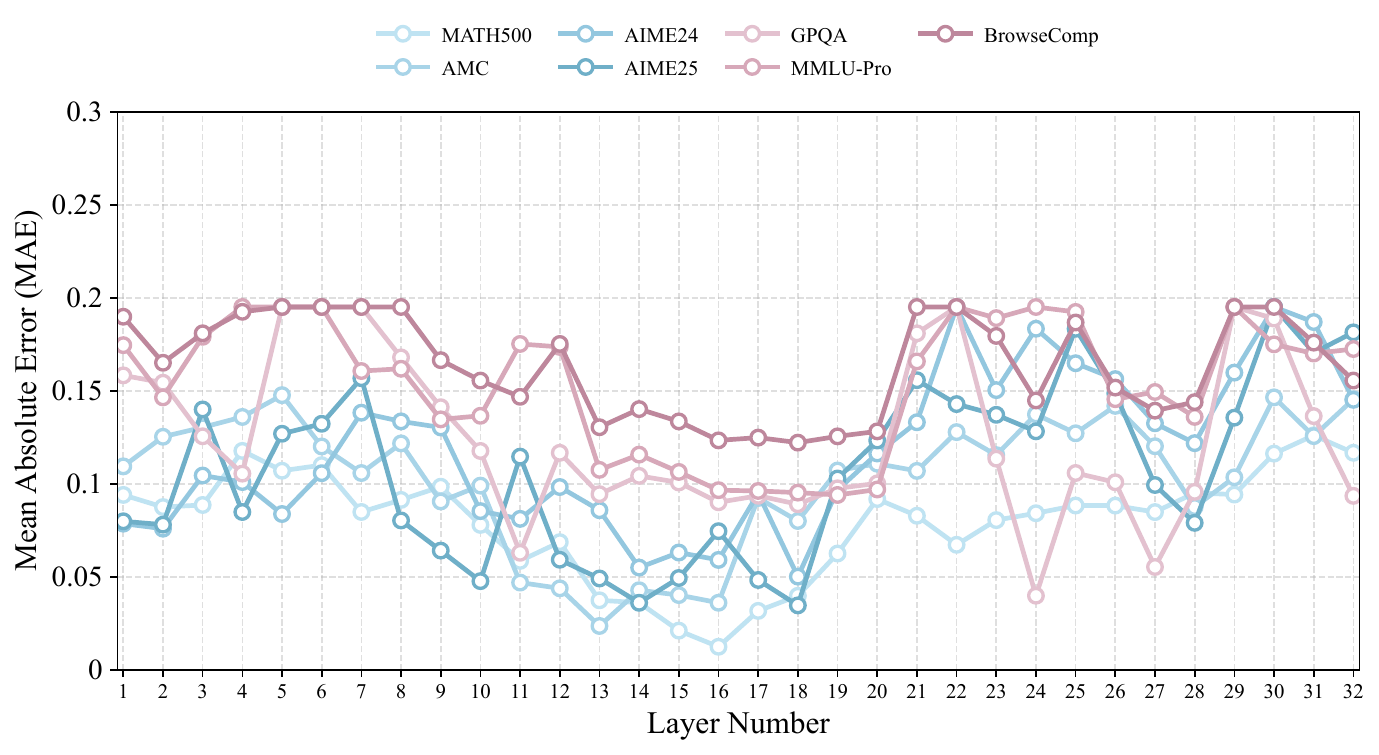}
        \subcaption{MAE results of the $L$ layer-wise predictors on the validation set, with each averaged {\it over 8 runs}.}
    \end{minipage}
    \hfill
    \begin{minipage}[b]{0.48\columnwidth}
        \centering
        \includegraphics[width=\linewidth]{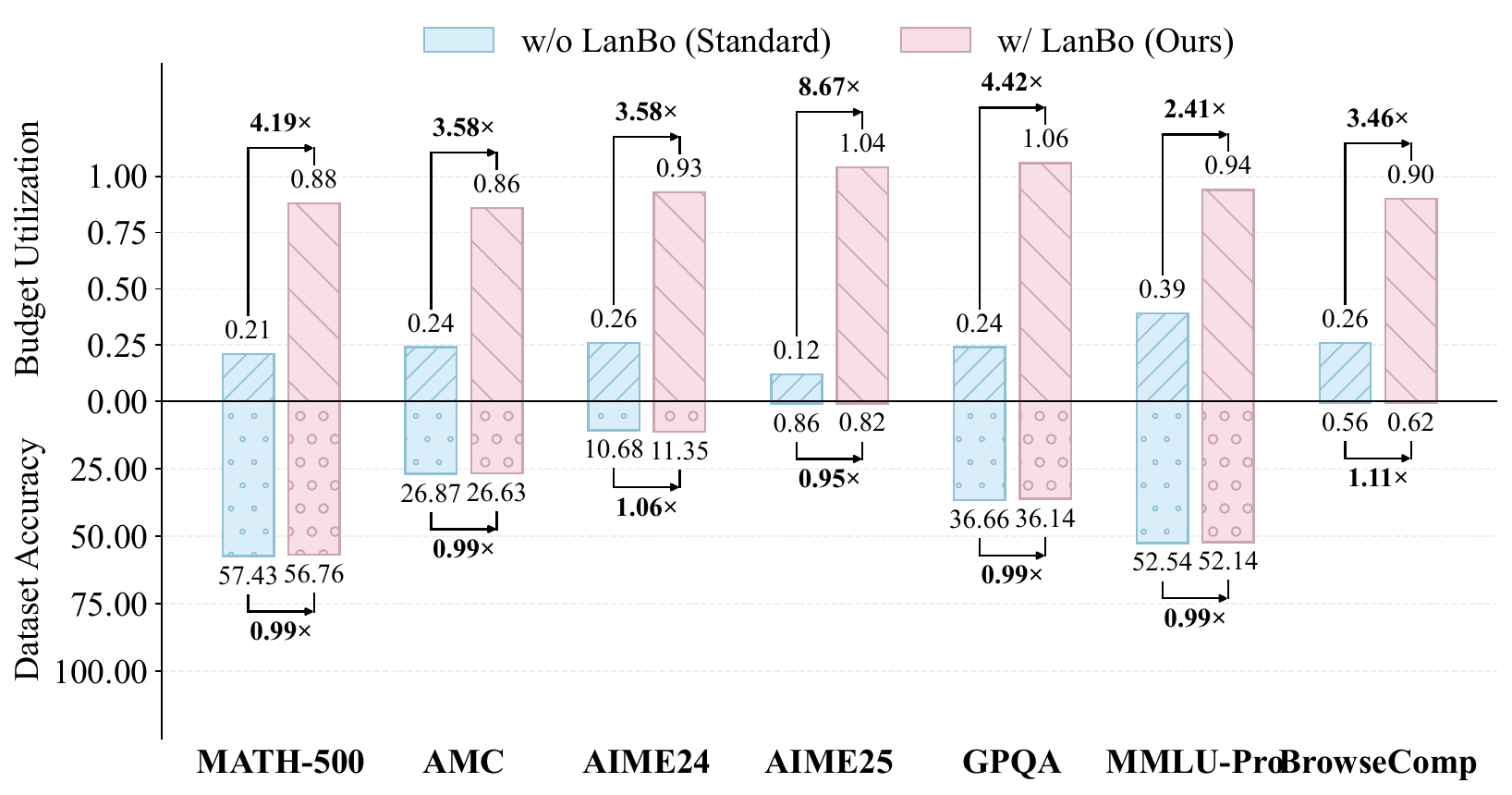}
        \subcaption{Comparison of system-level budget utilization and dataset accuracy before and after applying \latbp.}
    \end{minipage}
    \caption{{\bf Empirical Validation of \latbp~for Breaking the Overscaling Curse}. Here we use \texttt{Llama3.1-8B} with {\it voting} aggregation.}
    \label{fig:latbp-results-llama31}
\end{figure}

\begin{figure}[H]
    \centering
    \begin{minipage}[b]{0.48\columnwidth}
        \centering
        \includegraphics[width=\linewidth]{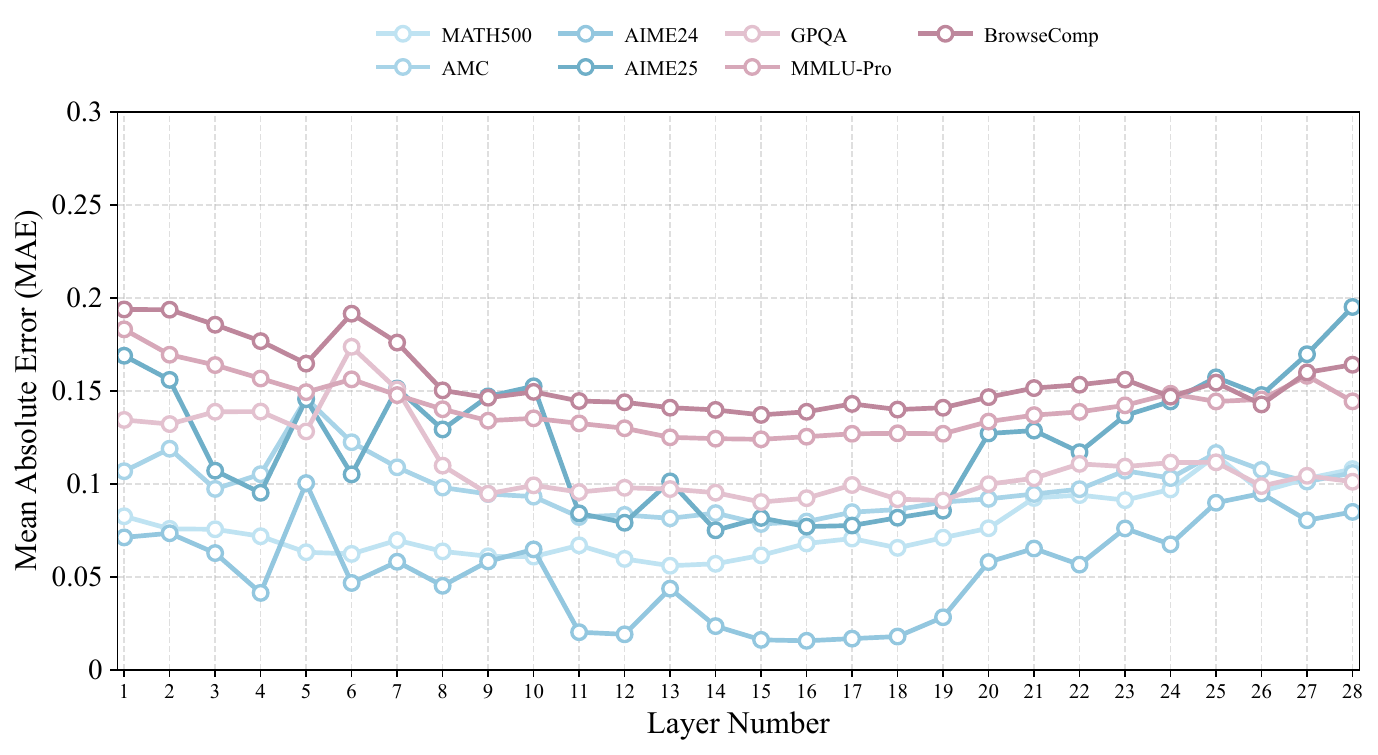}
        \subcaption{MAE results of the $L$ layer-wise predictors on the validation set, with each averaged {\it over 8 runs}.}
    \end{minipage}
    \hfill
    \begin{minipage}[b]{0.48\columnwidth}
        \centering
        \includegraphics[width=\linewidth]{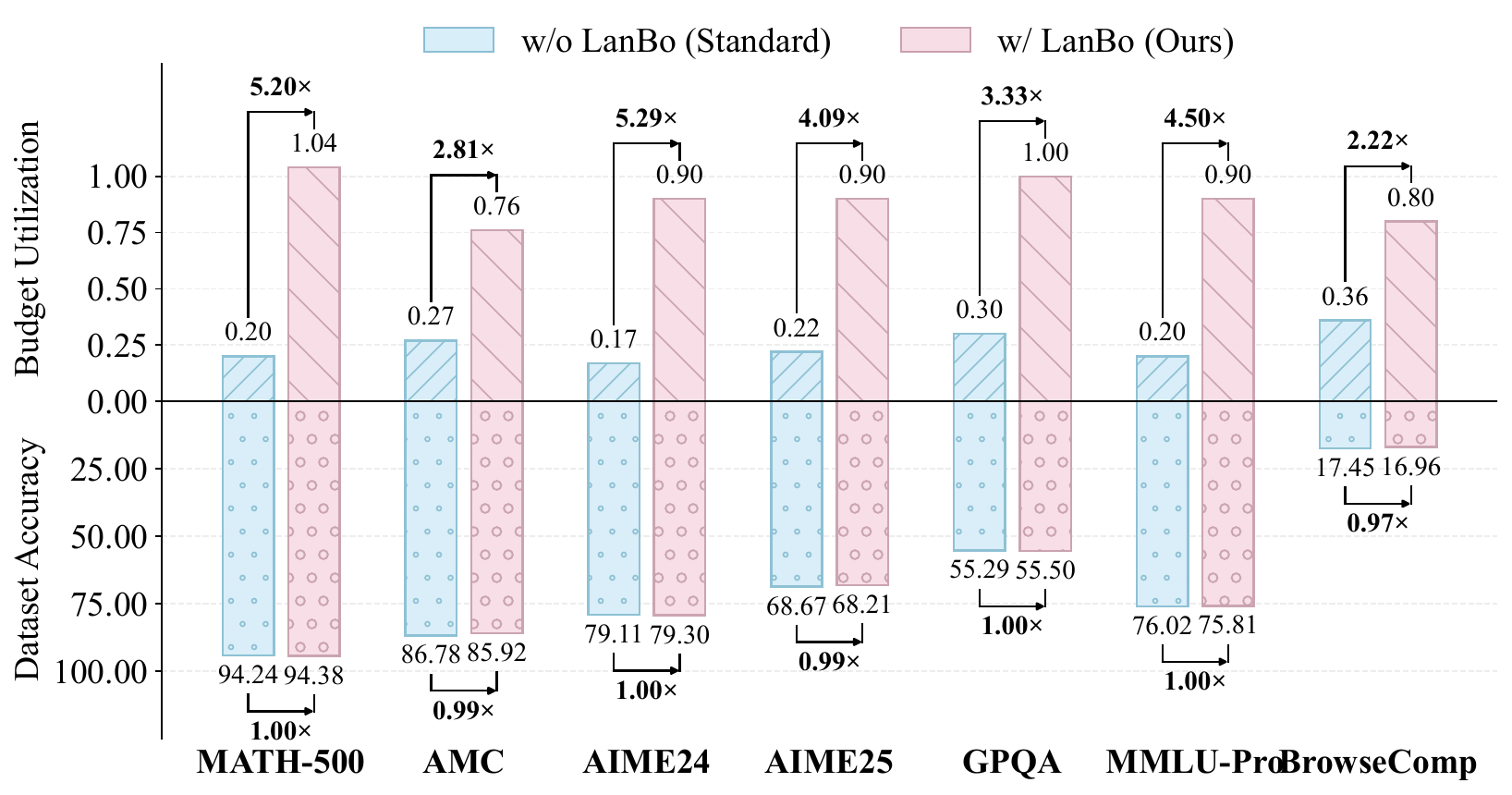}
        \subcaption{Comparison of system-level budget utilization and dataset accuracy before and after applying \latbp.}
    \end{minipage}
    \caption{{\bf Empirical Validation of \latbp~for Breaking the Overscaling Curse}. Here we use \texttt{Deepseek-R1-Distill-Qwen-7B} with {\it voting} aggregation.}
    \label{fig:latbp-results-r1}
\end{figure}

\begin{figure}[H]
    \centering
    \begin{minipage}[b]{0.48\columnwidth}
        \centering
        \includegraphics[width=\linewidth]{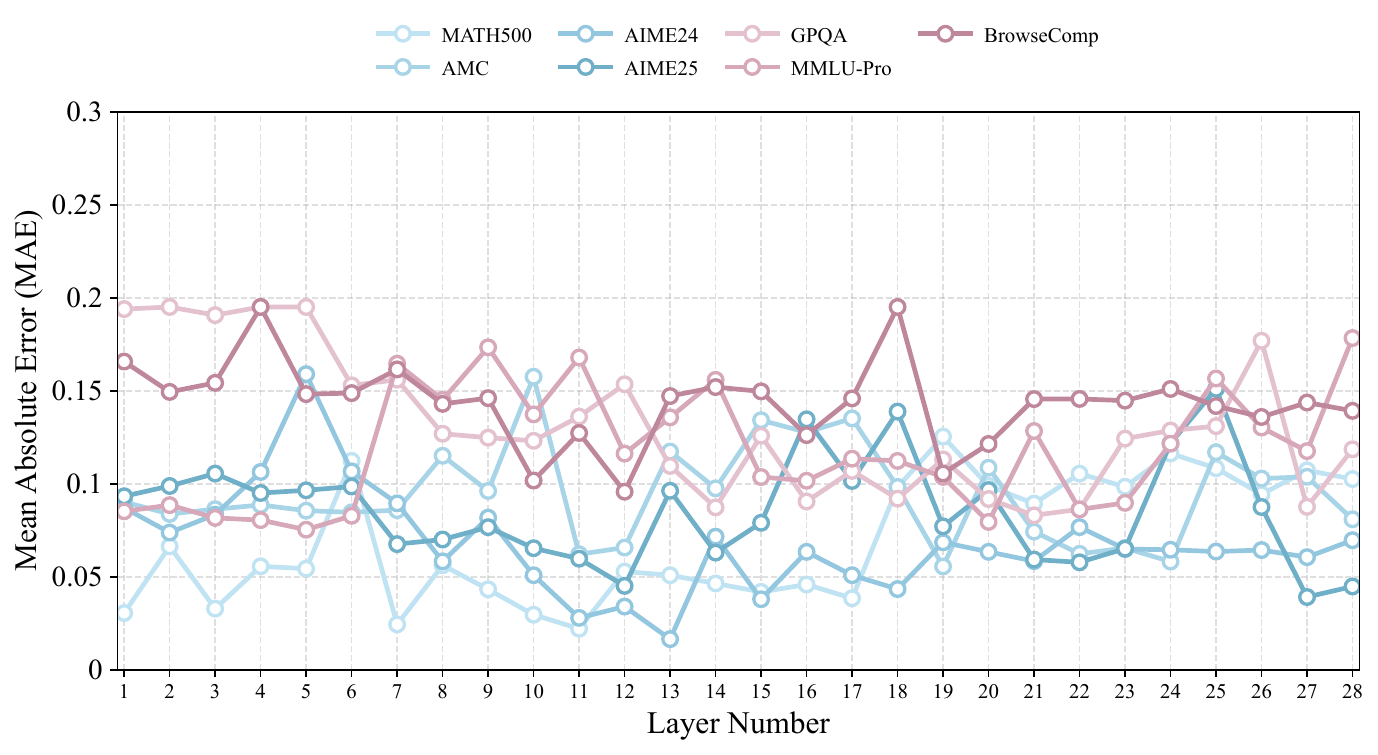}
        \subcaption{MAE results of the $L$ layer-wise predictors on the validation set, with each averaged {\it over 8 runs}.}
    \end{minipage}
    \hfill
    \begin{minipage}[b]{0.48\columnwidth}
        \centering
        \includegraphics[width=\linewidth]{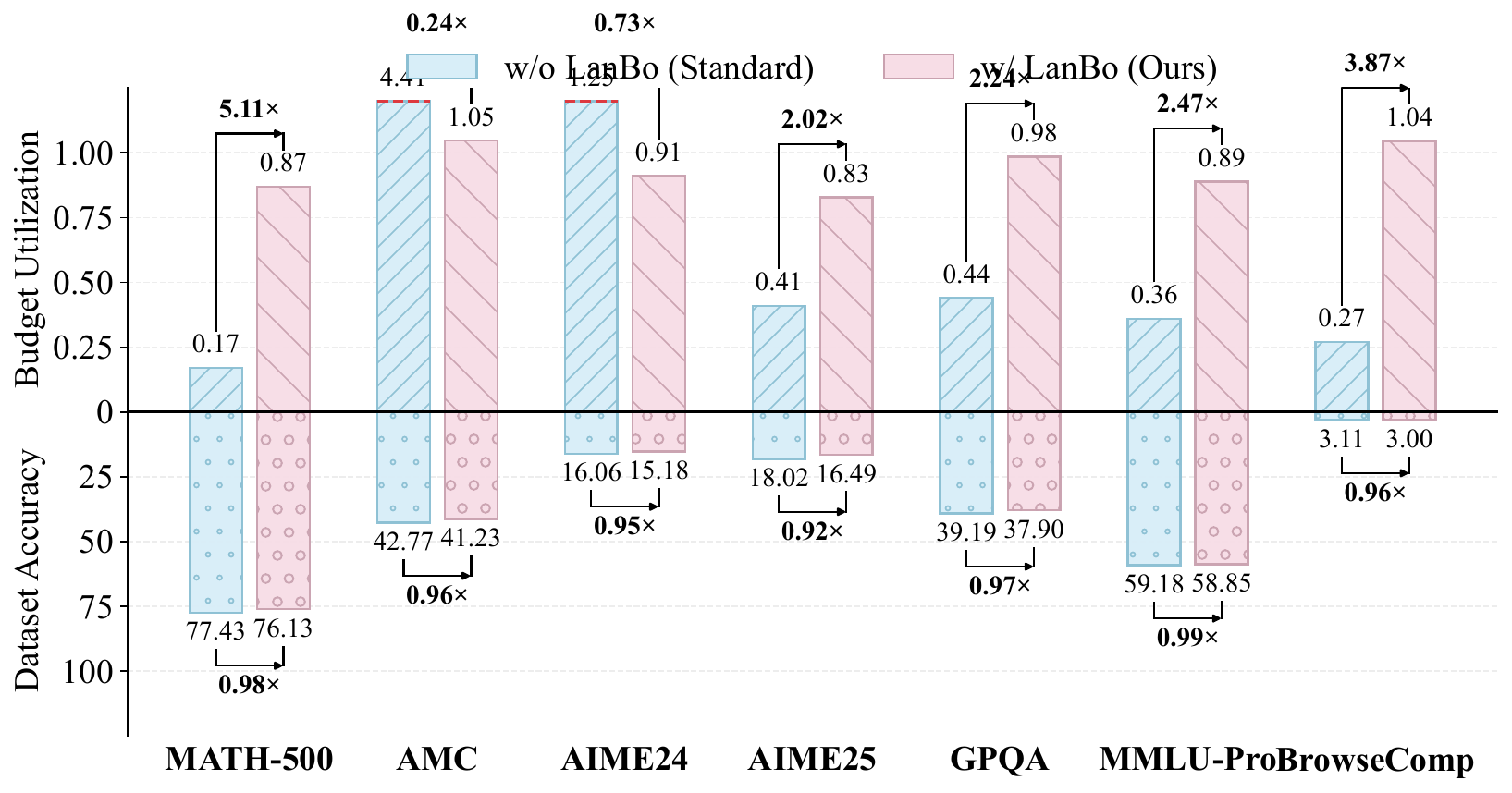}
        \subcaption{Comparison of system-level budget utilization and dataset accuracy before and after applying \latbp.}
    \end{minipage}
    \caption{{\bf Empirical Validation of \latbp~for Breaking the Overscaling Curse}. Here we use \texttt{Qwen2.5-7B} with {\it reward model} aggregation.}
    \label{fig:latbp-results-qwen25-model}
\end{figure}

\begin{figure}[H]
    \centering
    \begin{minipage}[b]{0.48\columnwidth}
        \centering
        \includegraphics[width=\linewidth]{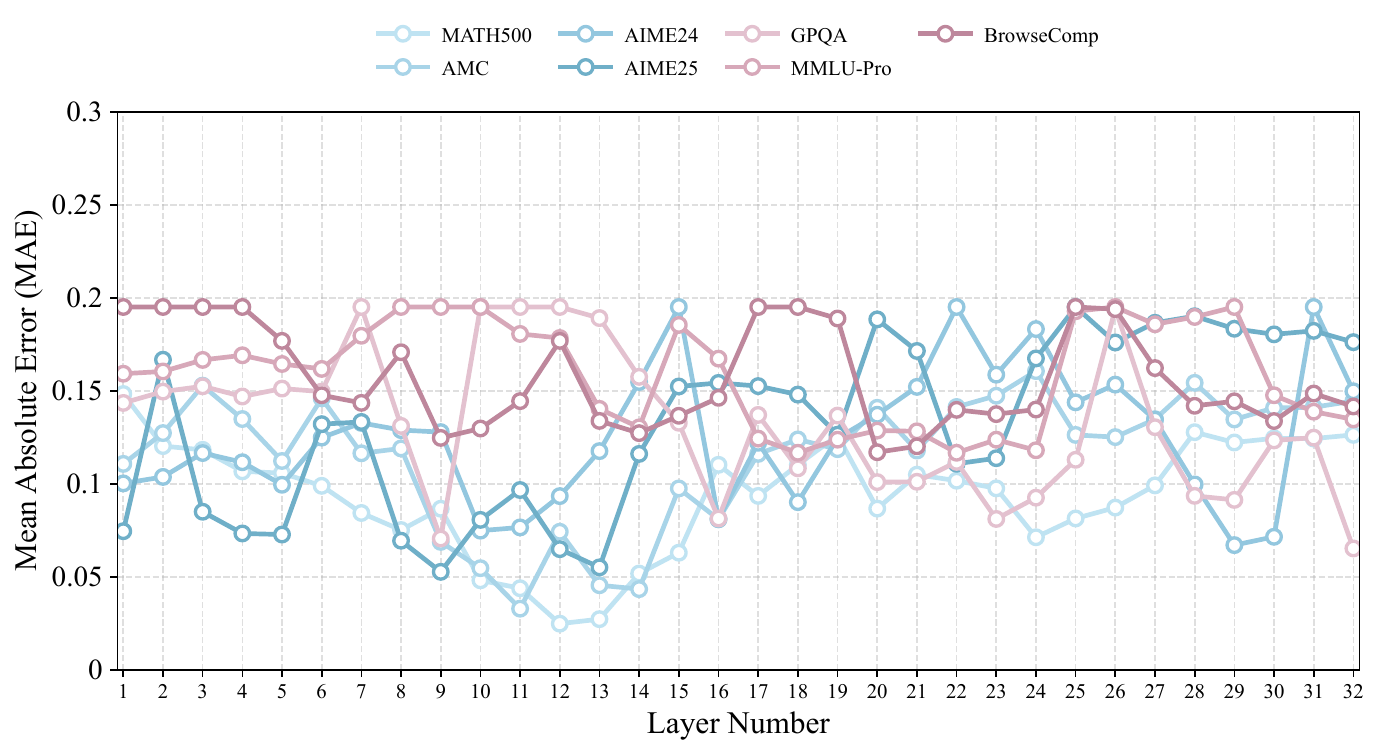}
        \subcaption{MAE results of the $L$ layer-wise predictors on the validation set, with each averaged {\it over 8 runs}.}
    \end{minipage}
    \hfill
    \begin{minipage}[b]{0.48\columnwidth}
        \centering
        \includegraphics[width=\linewidth]{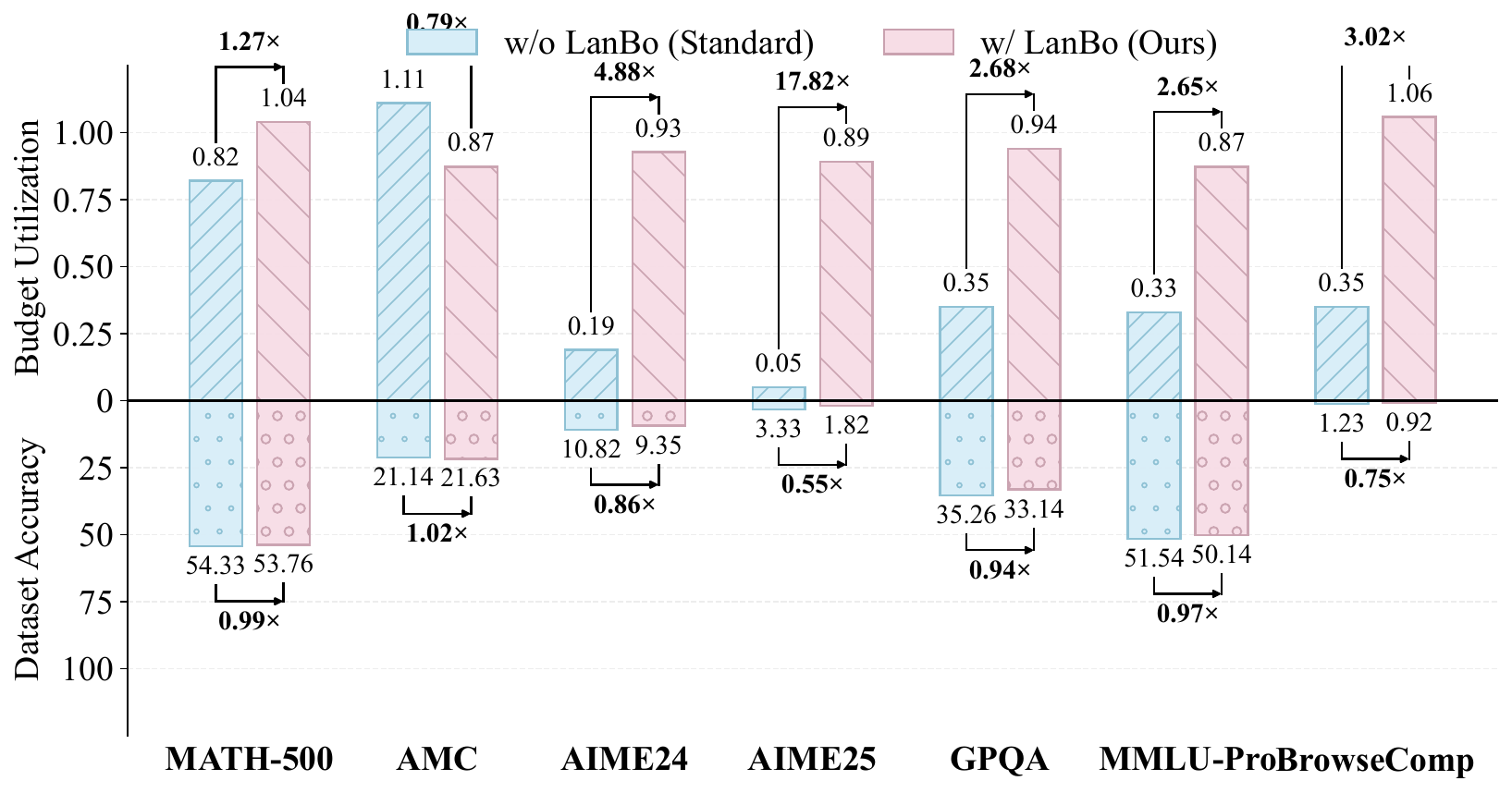}
        \subcaption{Comparison of system-level budget utilization and dataset accuracy before and after applying \latbp.}
    \end{minipage}
    \caption{{\bf Empirical Validation of \latbp~for Breaking the Overscaling Curse}. Here we use \texttt{Llama3.1-8B} with {\it reward model} aggregation.}
    \label{fig:latbp-results-llama31-model}
\end{figure}

\begin{figure}[H]
    \centering
    \begin{minipage}[b]{0.48\columnwidth}
        \centering
        \includegraphics[width=\linewidth]{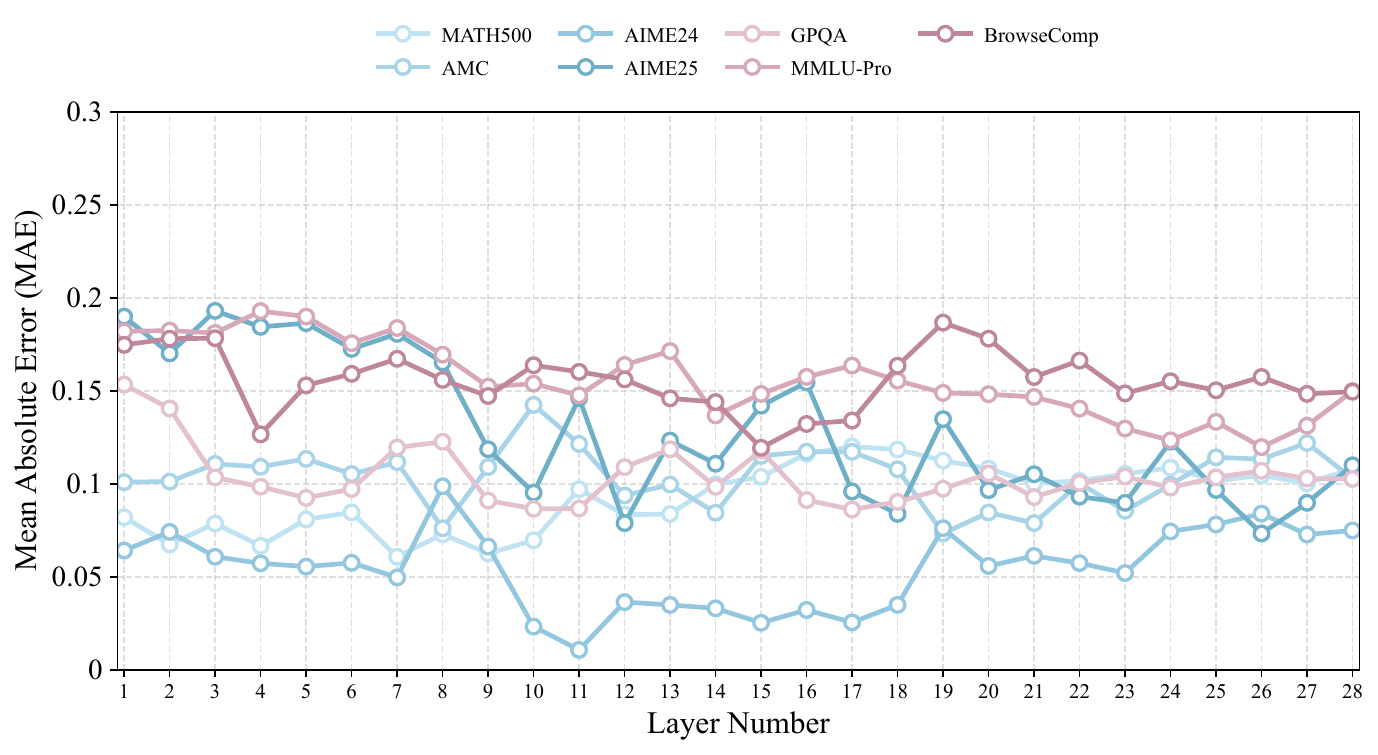}
        \subcaption{MAE results of the $L$ layer-wise predictors on the validation set, with each averaged {\it over 8 runs}.}
    \end{minipage}
    \hfill
    \begin{minipage}[b]{0.48\columnwidth}
        \centering
        \includegraphics[width=\linewidth]{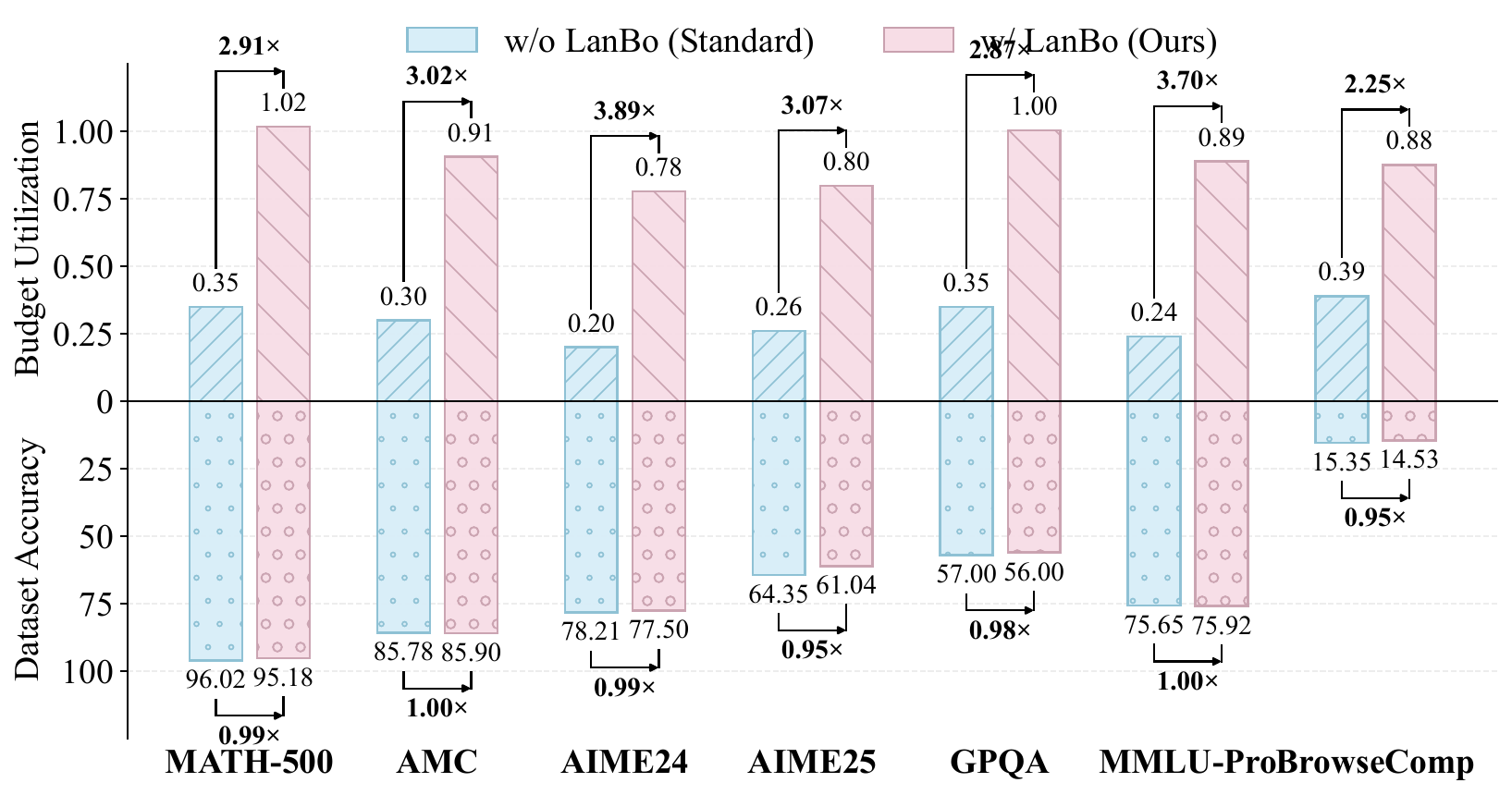}
        \subcaption{Comparison of system-level budget utilization and dataset accuracy before and after applying \latbp.}
    \end{minipage}
    \caption{{\bf Empirical Validation of \latbp~for Breaking the Overscaling Curse}. Here we use \texttt{Deepseek-R1-Distill-Qwen-7B} with {\it reward model} aggregation.}
    \label{fig:latbp-results-r1-model}
\end{figure}

\begin{figure}[H]
    \centering
    \begin{minipage}[b]{0.48\columnwidth}
        \centering
        \includegraphics[width=\linewidth]{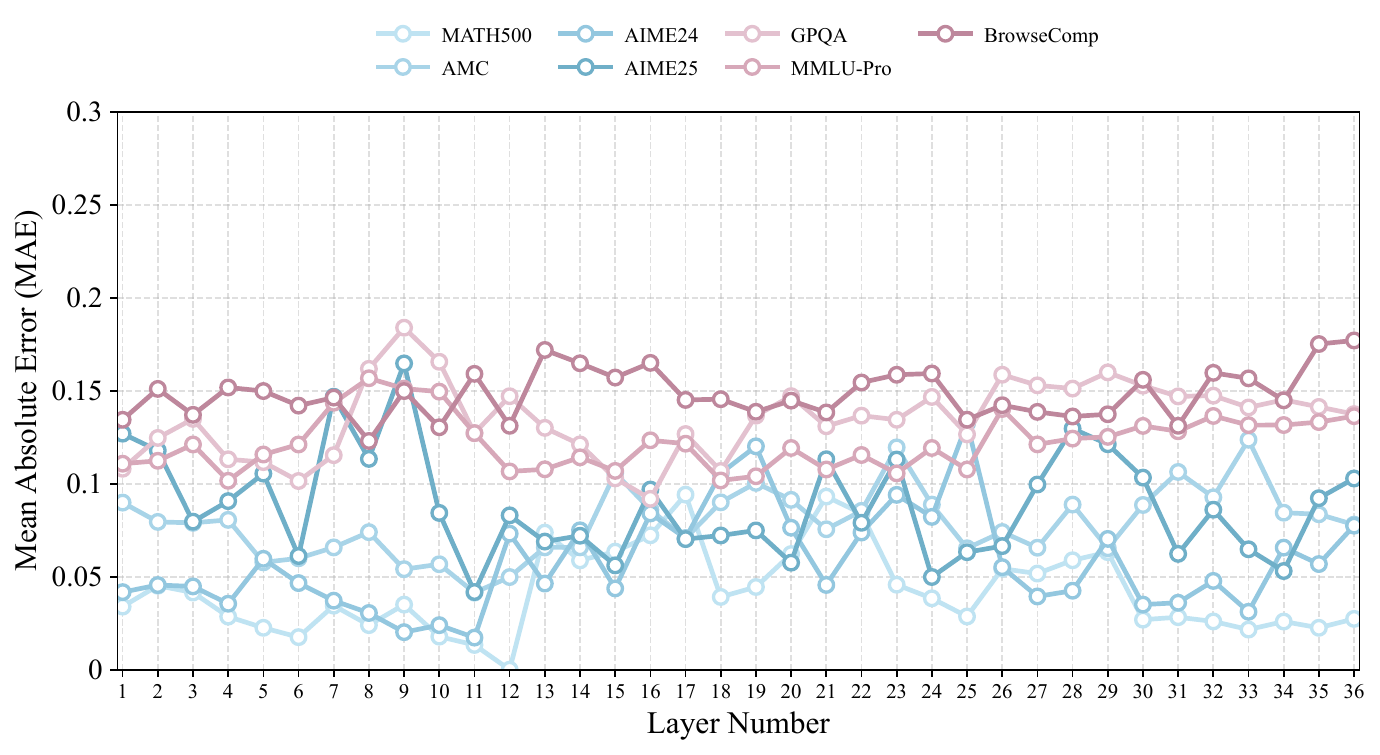}
        \subcaption{MAE results of the $L$ layer-wise predictors on the validation set, with each averaged {\it over 8 runs}.}
    \end{minipage}
    \hfill
    \begin{minipage}[b]{0.48\columnwidth}
        \centering
        \includegraphics[width=\linewidth]{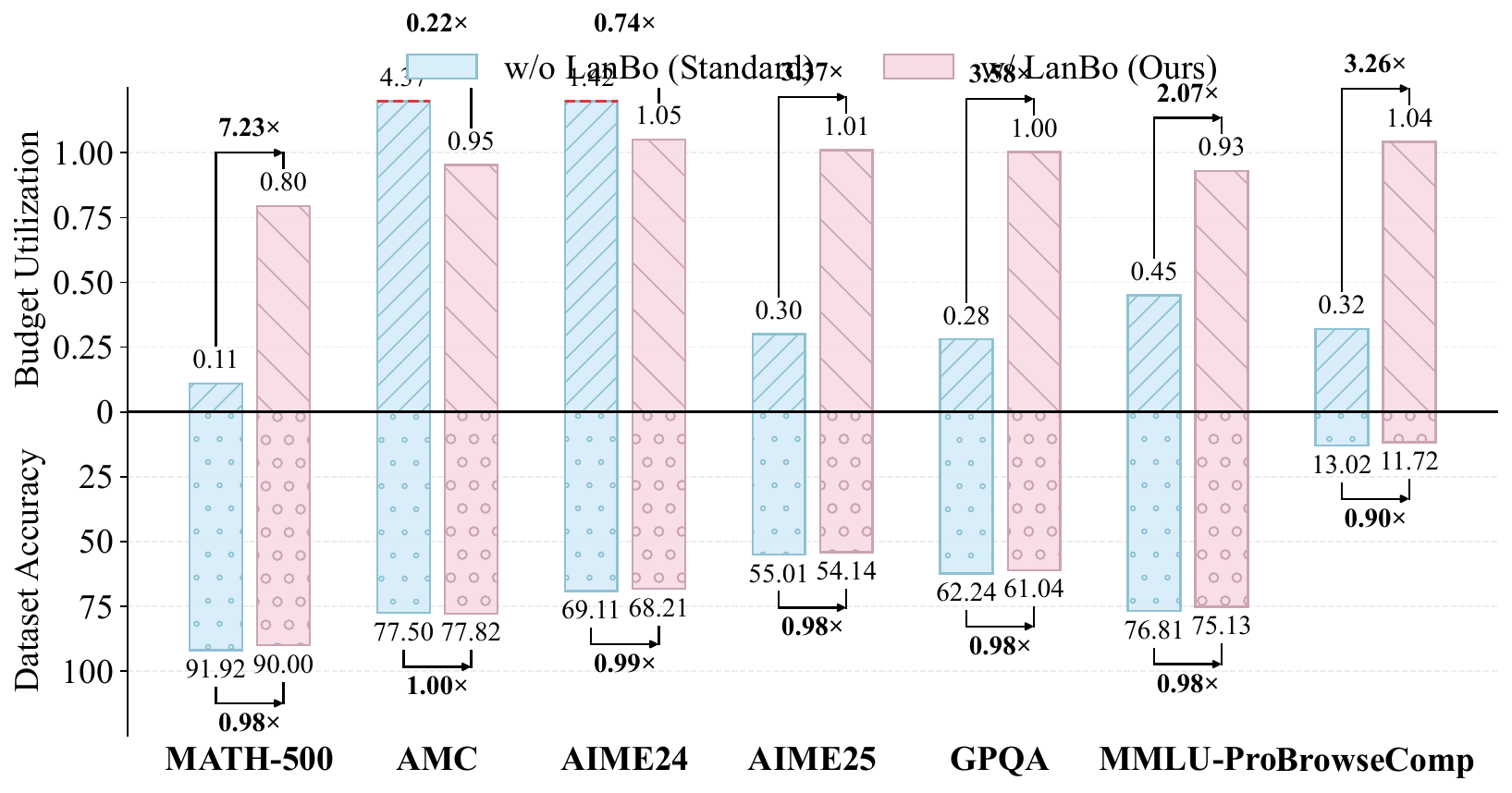}
        \subcaption{Comparison of system-level budget utilization and dataset accuracy before and after applying \latbp.}
    \end{minipage}
    \caption{{\bf Empirical Validation of \latbp~for Breaking the Overscaling Curse}. Here we use \texttt{Qwen3-4B} with {\it reward model} aggregation.}
    \label{fig:latbp-results-qwen3-model}
\end{figure}

%% file: 8F-PreAda-setting.tex
\section{Baselines of Efficient Parallel Decoding}
\label{appe:baselines}

\paragraph{AC \citep{aggarwal2023let}.}
Adaptive-Consistency (AC) introduces a Beta Stopping Criteria.
For each input, {\bf AC samples outputs one by one} and performs an early-stopping check after each sampling. Suppose that \(n\) outputs have been sampled so far, yielding \(m\) distinct results with an distribution \((p_1, p_2, \ldots, p_m)\), where \(p_1 \ge p_2 \ge \cdots \ge p_m\). If the condition $P(p_1 > p_2) \approx \int_0^{0.5} p_2^{v_2} \cdot (1-p_2)^{v_1} ~\mathrm{d}p_2 > 0.95$ is satisfied, sampling stops and the result corresponding to \(p_1\) is returned as the final answer; otherwise, sampling continues until the maximum budget \(L\) is reached.
In the original paper, $L=40$, and we adopt the same settings in our implementation.

\paragraph{ESC \citep{li2024escape}.}
Early-Stopping Self-Consistency (ESC) adopts a sliding-window entropy mechanism for adaptive early stopping. 
Specifically, for each input, {\bf ESC samples $w$ outputs in parallel at each step, where $w$ denotes the window size}. If all outputs within the current window are identical, sampling terminates and this output is returned as the final answer. Otherwise, ESC continues to sample another $w$ outputs until reaching the maximum budget $L$, after which the most frequent output is selected as the final answer. In the original paper, $w=8$ and $L=64$ are used for MATH500, while $w=5$ and $L=40$ are used for other datasets. We adopt the same settings in our implementation.

\paragraph{DSC \citep{wang2025make}.} Difficulty-Adaptive Self-Consistency (DSC) follows a three-stage procedure.
In Stage 1, the model ranks questions by estimating difficulty.
In Stage 2, it processes questions from hardest to easiest and draws $w$ samples per question. If $k$ consecutive questions are encountered for which all $w$ samples agree, it stops this pass and sets the sampling budget of all remaining (easier) questions to $1$.
In Stage 3, using the stopping point as a threshold, DSC draws one sample for each easier question, while for harder questions it adaptively increases the budget by doubling the number of $w$-sample blocks, up to a maximum of $L$ samples.
In the original paper, $w=4$, $k=32$, and $L=40$. We adopt the same settings in our implementation.

\paragraph{DeepConf \citep{fu2025deep}.}
Deep Think with Confidence (DeepConf) includes both offline and online algorithms; we mainly compare with its \textbf{online} variant, which belongs to efficient paradigms. 
The online algorithm consists of two stages. First, an \textit{offline warm-up} stage {\bf samples $N_{\text{init}}$ outputs in parallel} to compute a trace confidence threshold $s$. 
Then, in the \textit{adaptive sampling} stage, {\bf outputs are sampled one by one}. During each sampling, a confidence check is performed every $n$ tokens: if the trace confidence within a window falls below $s$, the current trace is truncated and a new sampling begins; otherwise, sampling continues until completion. 
After completing this sampling, trace-confidence-weighted majority voting is conducted over all collected outputs so for. If the highest frequency exceeds a threshold $\tau$, sampling stops and that output is returned as the final answer; otherwise, sampling continues until reaching the maximum budget $L$.
We set $N_{\text{init}} = 16$, $\tau=0.95$, and $L = 512$ as in the original paper.

%% file: 8G-PreAda-results.tex
\section{Main Results of Efficient Parallel Decoding}

\subsection{Supplementary Main Results}
\label{appe:main-results}

\textcolor{deepred}{Tables \ref{tab:qwen25-7B-results} -- \ref{tab:r1-results}} presents empirical validation of our \latbp ~for efficient parallel decoding, supplementing the \texttt{Qwen3-4B} results shown in \textcolor{deepred}{Table \ref{tab:qwen3-4B-results}} in the main text.

Under each system, we report the real-world efficiency (memory $\delta_{\mathcal{M}}$ and latency $\delta_{\mathcal{T}}$, both the ratio relative to \textit{STD.}) and accuracy ({\it Acc.}), where \textit{STD.} denotes the standard parallel decoding. {\bf Bold} and \underline{underline} denote the best and second best, respectively. \textcolor{blue}{Blue} means higher accuracy than \textit{STD.}; \textcolor{red}{red} means lower efficiency.
Each result is averaged \textit{\textbf{over 32 runs}}.

\input{tables/qwen25-7B-results}

\input{tables/llama31-8B-results}

\input{tables/r1-results}

\subsection{Standard Deviations of Main Results}
\label{appe:std}

\textcolor{deepred}{Table \ref{tab:acc-std}} reports the standard deviations of accuracy ({\it Acc.}) of all methods under each system in \textcolor{deepred}{Table \ref{tab:qwen3-4B-results}} and \textcolor{deepred}{\ref{tab:qwen25-7B-results} -- \ref{tab:r1-results}}.

\begin{table}[H]
\caption{Standard deviation of accuracy ({\it Acc.}) corresponding to \textcolor{deepred}{Table \ref{tab:qwen3-4B-results}} and \textcolor{deepred}{\ref{tab:qwen25-7B-results} -- \ref{tab:r1-results}} across 32 runs.}
\vspace{0.05in}
\centering
\footnotesize
\renewcommand\arraystretch{1.1}
\setlength{\tabcolsep}{1.2mm}{
\resizebox{0.95\textwidth}{!}{
\begin{tabular}{l l ccccccc}
\toprule
\textbf{Paradigm}
& \textbf{Method} 
& \textbf{MATH500} 
& \textbf{AMC} 
& \textbf{AIME24} 
& \textbf{AIME25} 
& \textbf{GPQA} 
& \textbf{MMLU-Pro} 
& \textbf{BrowseComp} \\
\midrule

\multicolumn{9}{c}{\textbf{\texttt{Qwen2.5-7B}}} \\
\midrule
--- & {\it STD.}  & 0.32 & 0.41 & 0.28 & 0.14 & 0.55 & 0.49 & 0.08 \\
\hdashline
\multirow{4}{*}{\tdt}
& AC          & 0.33 & 0.51 & 0.39 & 0.12 & 0.74 & 0.66 & 0.05 \\
& ESC         & 0.38 & 0.25 & 0.26 & 0.13 & 0.61 & 0.59 & 0.09 \\
& DSC         & 0.31 & 0.49 & 0.23 & 0.29 & 0.70 & 0.68 & 0.02 \\
& DeepConf    & 0.37 & 0.44 & 0.22 & 0.26 & 0.88 & 0.91 & 0.08 \\
\hdashline
\textbf{\tbt}   & \textbf{\latbp ~(Ours)}   & 0.31 & 0.35 & 0.44 & 0.11 & 0.37 & 0.33 & 0.07 \\
\midrule

\multicolumn{9}{c}{\textbf{\texttt{Llama3.1-8B}}} \\
\midrule
--- & {\it STD.}  & 0.46 & 0.33 & 0.12 & 0.19 & 0.57 & 0.52 & 0.01 \\
\hdashline
\multirow{4}{*}{\tdt}
& AC          & 0.68 & 0.46 & 0.25 & 0.11 & 0.79 & 0.64 & 0.06 \\
& ESC         & 0.61 & 0.40 & 0.14 & 0.11 & 0.66 & 0.63 & 0.03 \\
& DSC         & 0.65 & 0.33 & 0.18 & 0.15 & 0.71 & 0.69 & 0.06 \\
& DeepConf    & 0.82 & 0.51 & 0.10 & 0.03 & 0.73 & 0.77 & 0.01 \\
\hdashline
\textbf{\tbt}   & \textbf{\latbp ~(Ours)}   & 0.34 & 0.29 & 0.08 & 0.05 & 0.10 & 0.36 & 0.03 \\
\midrule

\multicolumn{9}{c}{\textbf{\texttt{Deepseek-R1-Distill-Qwen-7B}}} \\
\midrule
--- & {\it STD.}  & 0.38 & 0.44 & 0.59 & 0.56 & 0.47 & 0.45 & 0.15 \\
\hdashline
\multirow{4}{*}{\tdt}
& AC          & 0.55 & 0.63 & 0.78 & 0.74 & 0.66 & 0.61 & 0.19 \\
& ESC         & 0.49 & 0.56 & 0.71 & 0.68 & 0.58 & 0.55 & 0.14 \\
& DSC         & 0.57 & 0.65 & 0.82 & 0.79 & 0.69 & 0.66 & 0.11 \\
& DeepConf    & 0.73 & 0.81 & 0.97 & 0.93 & 0.85 & 0.88 & 0.15 \\
\hdashline
\textbf{\tbt}   & \textbf{\latbp ~(Ours)}   & 0.26 & 0.30 & 0.38 & 0.36 & 0.31 & 0.29 & 0.13 \\
\midrule

\multicolumn{9}{c}{\textbf{\texttt{Qwen3-4B}}} \\
\midrule
--- & {\it STD.}  & 0.35 & 0.41 & 0.54 & 0.52 & 0.45 & 0.43 & 0.03 \\
\hdashline
\multirow{4}{*}{\tdt}
& AC          & 0.51 & 0.60 & 0.73 & 0.71 & 0.63 & 0.59 & 0.06 \\
& ESC         & 0.47 & 0.55 & 0.68 & 0.66 & 0.57 & 0.54 & 0.12 \\
& DSC         & 0.53 & 0.61 & 0.75 & 0.72 & 0.64 & 0.60 & 0.08 \\
& DeepConf    & 0.69 & 0.78 & 0.92 & 0.89 & 0.81 & 0.84 & 0.12 \\
\hdashline
\textbf{\tbt}   & \textbf{\latbp ~(Ours)}  & 0.24 & 0.28 & 0.36 & 0.34 & 0.29 & 0.27 & 0.11 \\
\bottomrule
\end{tabular}
}}
\label{tab:acc-std}
\end{table}

\subsection{Cost Statistics}
\label{appe:latency}

\textcolor{deepred}{Table \ref{tab:LBA-training-duration}} and \ref{tab:LBA-inference-latency} report the training duration inference latency of predictors in our \latbp.
Since they depend solely on the hidden size of the applied language model, we report results on four different models.

\begin{table}[H]
\caption{Training Duration of predictors in our \latbp ({\bf the total of $L$ predictors}).}
\vspace{0.05in}
\centering
\footnotesize
\renewcommand\arraystretch{1.2}
\setlength{\tabcolsep}{1.5mm}{
\resizebox{0.7\textwidth}{!}{
\begin{tabular}{cccc}

\toprule

\texttt{Qwen2.5-7B} & \texttt{Llama3.1-8B} & \texttt{Deepseek-R1-Distill-Qwen-7B} & \texttt{Qwen3-4B} \\

\midrule

67.4s & 79.5s & 68.0s & 87.4s \\

\bottomrule
\end{tabular}}}
\label{tab:LBA-training-duration}%
\end{table}

\begin{table}[H]
\caption{Inference latency of predictors in our \latbp ({\bf the total of $L$ predictors}).}
\vspace{0.05in}
\centering
\footnotesize
\renewcommand\arraystretch{1.2}
\setlength{\tabcolsep}{1.5mm}{
\resizebox{0.7\textwidth}{!}{
\begin{tabular}{cccc}

\toprule

\texttt{Qwen2.5-7B} & \texttt{Llama3.1-8B} & \texttt{Deepseek-R1-Distill-Qwen-7B} & \texttt{Qwen3-4B} \\

\midrule

0.023ms & 0.029ms & 0.025ms & 0.031ms \\

\bottomrule
\end{tabular}}}
\label{tab:LBA-inference-latency}%
\end{table}

\textcolor{deepred}{Table \ref{tab:latency}} reports the average per-sample inference latency of all efficient decoding methods.

\begin{table}[H]
\caption{Average {\bf per-sample} inference latency (seconds) of all methods corresponding to \textcolor{deepred}{Table \ref{tab:qwen3-4B-results}} and \textcolor{deepred}{\ref{tab:qwen25-7B-results} -- \ref{tab:r1-results}}.}
\vspace{0.05in}
\centering
\footnotesize
\renewcommand\arraystretch{1.1}
\setlength{\tabcolsep}{1.2mm}{
\resizebox{0.95\textwidth}{!}{
\begin{tabular}{l l ccccccc}
\toprule
\textbf{Paradigm}
& \textbf{Method} 
& \textbf{MATH500} 
& \textbf{AMC} 
& \textbf{AIME24} 
& \textbf{AIME25} 
& \textbf{GPQA} 
& \textbf{MMLU-Pro} \\
\midrule

\multicolumn{8}{c}{\textbf{\texttt{Qwen2.5-7B}}} \\
\midrule
--- & {\it STD.}  & 3.4 & 5.2 & 31.2 & 26.3 & 9.3 & 8.5 \\
\hdashline
\multirow{4}{*}{\tdt}
& AC          & 5.1 & 12.1 & 108.0 & 74.4 & 25.1 & 17.3 \\
& ESC         & 9.4 & 9.9  & 30.0  & 38.1 & 21.1 & 15.1 \\
& DSC         & 4.9 & 11.2 & 57.1  & 25.5 & 20.3 & 31.8 \\
& DeepConf    & 11.7& 13.9 & 98.6  & 49.4 & 31.7 & 27.3 \\
\hdashline
\textbf{\tbt}   & \textbf{\latbp{} (Ours)}   & 1.6 & 2.3  & 25.6  & 17.9 & 4.7  & 3.3  \\
\midrule

\multicolumn{8}{c}{\textbf{\texttt{Llama3.1-8B}}} \\
\midrule
--- & {\it STD.}  & 3.0 & 4.6 & 28.3 & 27.3 & 7.2 & 8.9 \\
\hdashline
\multirow{4}{*}{\tdt}
& AC          & 5.1 & 10.4 & 100.7 & 95.6 & 21.5 & 28.2 \\
& ESC         & 4.2 & 9.3  & 46.4  & 66.6 & 17.1 & 19.2 \\
& DSC         & 5.2 & 10.0 & 46.1  & 26.8 & 15.0 & 25.2 \\
& DeepConf    & 7.1 & 12.0 & 113.8 & 142.5& 30.0 & 37.0 \\
\hdashline
\textbf{\tbt}   & \textbf{\latbp{} (Ours)}   & 1.2 & 1.7  & 12.2  & 14.2 & 2.7  & 3.7  \\
\midrule

\multicolumn{8}{c}{\textbf{\texttt{Deepseek-R1-Distill-Qwen-7B}}} \\
\midrule
--- & {\it STD.}  & 15.7 & 24.2 & 204.9 & 225.4 & 72.1 & 64.7 \\
\hdashline
\multirow{4}{*}{\tdt}
& AC          & 29.7 & 67.0 & 512.2  & 619.8 & 188.9 & 150.1 \\
& ESC         & 13.2 & 41.6 & 448.7  & 493.6 & 120.4 & 124.2 \\
& DSC         & 33.4 & 69.9 & 733.5  & 450.8 & 290.6 & 115.8 \\
& DeepConf    & 60.6 & 86.2 & 1020.4 & 804.7 & 274.0 & 251.0 \\
\hdashline
\textbf{\tbt}   & \textbf{\latbp{} (Ours)}   & 5.8  & 11.6 & 145.5  & 112.7 & 32.4  & 31.7  \\
\midrule

\multicolumn{8}{c}{\textbf{\texttt{Qwen3-4B}}} \\
\midrule
--- & {\it STD.}  & 6.5 & 11.1 & 65.3 & 51.4 & 33.2 & 30.7 \\
\hdashline
\multirow{4}{*}{\tdt}
& AC          & 10.1 & 29.6 & 203.7 & 123.9 & 70.1  & 72.5 \\
& ESC         & 5.7  & 20.4 & 126.0 & 115.6 & 65.4  & 68.5 \\
& DSC         & 8.8  & 19.0 & 144.3 & 142.9 & 130.1 & 91.2 \\
& DeepConf    & 18.0 & 36.3 & 302.3 & 193.3 & 106.2 & 153.2 \\
\hdashline
\textbf{\tbt}   & \textbf{\latbp{} (Ours)}   & 2.9  & 5.8  & 51.6  & 28.8  & 15.3  & 12.3  \\
\bottomrule
\end{tabular}
}}
\label{tab:latency}
\end{table}

%% file: tables/qwen25-7B-results.tex
\begin{table}[H]
\caption{{\bf Empirical Validation of Our \latbp ~for Efficient Parallel Decoding.}
This table shows \texttt{Qwen2.5-7B} results.
}
\vspace{0.05in}
\centering
\footnotesize
\renewcommand\arraystretch{1.}
\setlength{\tabcolsep}{1.3mm}{
\resizebox{1\textwidth}{!}{
\begin{tabular}{llccccccccccccccccccccc}

\toprule

\multirow{2}{*}{\bf Paradigm}
& \multirow{2}{*}{\bf Method}
& \multicolumn{3}{c}{\bf MATH500}
& \multicolumn{3}{c}{\bf AMC}
& \multicolumn{3}{c}{\bf AIME24}
& \multicolumn{3}{c}{\bf AIME25}
& \multicolumn{3}{c}{\bf GPQA}
& \multicolumn{3}{c}{\bf MMLU-Pro}
& \multicolumn{3}{c}{\bf BrowseComp}
\\
\cmidrule(lr){3-5}
\cmidrule(lr){6-8}
\cmidrule(lr){9-11}
\cmidrule(lr){12-14}
\cmidrule(lr){15-17}
\cmidrule(lr){18-20}
\cmidrule(lr){21-23}

& &
$\delta_{\mathcal{M}} \downarrow$ & $\delta_{\mathcal{T}} \downarrow$ & {\it Acc.} $\uparrow$ & 
$\delta_{\mathcal{M}} \downarrow$ & $\delta_{\mathcal{T}} \downarrow$ & {\it Acc.} $\uparrow$ & 
$\delta_{\mathcal{M}} \downarrow$ & $\delta_{\mathcal{T}} \downarrow$ & {\it Acc.} $\uparrow$ & 
$\delta_{\mathcal{M}} \downarrow$ & $\delta_{\mathcal{T}} \downarrow$ & {\it Acc.} $\uparrow$ & 
$\delta_{\mathcal{M}} \downarrow$ & $\delta_{\mathcal{T}} \downarrow$ & {\it Acc.} $\uparrow$ & 
$\delta_{\mathcal{M}} \downarrow$ & $\delta_{\mathcal{T}} \downarrow$ & {\it Acc.} $\uparrow$ &
$\delta_{\mathcal{M}} \downarrow$ & $\delta_{\mathcal{T}} \downarrow$ & {\it Acc.} $\uparrow$ \\

\midrule

--- & {\it STD.} & {\it 1.00} & {\it 1.00} & {\it 78.17} 
& {\it 1.00} & {\it 1.00} & {\it 47.18}
& {\it 1.00} & {\it 1.00} & {\it 20.10}
& {\it 1.00} & {\it 1.00} & {\it 13.58}
& {\it 1.00} & {\it 1.00} & {\it 39.49}
& {\it 1.00} & {\it 1.00} & {\it 60.18}
& {\it 1.00} & {\it 1.00} & {\it 3.21}
\\

\hdashline
\multirow{4}{*}{\tdt}
& AC 
& \underline{0.24} & \textcolor{red}{1.49} & 77.26
& 0.28 & \textcolor{red}{2.32} & 46.67
& 0.74 & \textcolor{red}{3.46} & 19.65
& 0.66 & \textcolor{red}{2.83} & 11.94
& 0.38 & \textcolor{red}{2.70} & 38.79
& 0.32 & \textcolor{red}{2.03} & 59.28
& 0.40 & \textcolor{red}{1.76} & 2.58
\\

& ESC 
& 0.52 & \textcolor{red}{2.77} & 77.58
& 0.34 & \textcolor{red}{1.90} & 46.77
& \underline{0.32} & \underline{0.96} & \underline{19.72}
& 0.33 & \textcolor{red}{1.45} & 12.02
& 0.36 & \textcolor{red}{2.27} & 39.05
& \underline{0.28} & \underline{\textcolor{red}{1.78}} & 59.38
& \underline{0.35} & \underline{\textcolor{red}{1.65}} & 2.71
\\

& DSC 
& 0.44 & \underline{\textcolor{red}{1.43}} & 77.33
& \underline{0.31} & \underline{\textcolor{red}{2.16}} & \underline{46.85}
& {\bf 0.27} & \textcolor{red}{1.83} & 18.52
& {\bf 0.28} & \underline{0.97} & 9.89
& \underline{0.31} & \underline{\textcolor{red}{2.18}} & 38.12
& 0.40 & \textcolor{red}{3.74} & 59.59
& 0.41 & \textcolor{red}{3.12} & 1.92
\\

& DeepConf 
& 0.45 & \textcolor{red}{3.45} & {\bf 77.82}
& 0.38 & \textcolor{red}{2.67} & 46.67
& \textcolor{red}{1.13} & \textcolor{red}{3.16} & 18.08
& 0.78 & \textcolor{red}{1.88} & \underline{12.25}
& 0.53 & \textcolor{red}{3.41} & {\bf 39.19}
& 0.61 & \textcolor{red}{3.21} & \underline{59.63}
& 0.65 & \textcolor{red}{2.18} & \underline{3.00}
\\

\hdashline
\rowcolor{gray!15}
{\bf \tbt} & {\bf \latbp ~(Ours)}
& {\bf 0.18} & {\bf 0.48} & \underline{77.49}
& {\bf 0.23} & {\bf 0.74} & {\bf 47.08}
& 0.49 & {\bf 0.52} & \textbf{\textcolor{blue}{20.91}}
& \underline{0.35} & {\bf 0.68} & {\bf \textcolor{blue}{13.84}}
& {\bf 0.27} & {\bf 0.51} & \underline{38.32}
& {\bf 0.23} & {\bf 0.39} & {\bf 59.85}
& {\bf 0.30} & {\bf 0.49} & {\bf 3.17}
\\

\bottomrule

\end{tabular}%
}}
\label{tab:qwen25-7B-results}%
\end{table}

%% file: tables/llama31-8B-results.tex
\begin{table}[H]
\caption{{\bf Empirical Validation of Our \latbp ~for Efficient Parallel Decoding.} 
This table shows \texttt{Llama3.1-8B} results.
}
\vspace{0.05in}
\centering
\footnotesize
\renewcommand\arraystretch{1.}
\setlength{\tabcolsep}{1.3mm}{
\resizebox{1\textwidth}{!}{
\begin{tabular}{llccccccccccccccccccccc}

\toprule

\multirow{2}{*}{\bf Paradigm}
& \multirow{2}{*}{\bf Method}
& \multicolumn{3}{c}{\bf MATH500}
& \multicolumn{3}{c}{\bf AMC}
& \multicolumn{3}{c}{\bf AIME24}
& \multicolumn{3}{c}{\bf AIME25}
& \multicolumn{3}{c}{\bf GPQA}
& \multicolumn{3}{c}{\bf MMLU-Pro}
& \multicolumn{3}{c}{\bf BrowseComp}
\\
\cmidrule(lr){3-5}
\cmidrule(lr){6-8}
\cmidrule(lr){9-11}
\cmidrule(lr){12-14}
\cmidrule(lr){15-17}
\cmidrule(lr){18-20}
\cmidrule(lr){21-23}

& &
$\delta_{\mathcal{M}} \downarrow$ & $\delta_{\mathcal{T}} \downarrow$ & {\it Acc.} $\uparrow$ & 
$\delta_{\mathcal{M}} \downarrow$ & $\delta_{\mathcal{T}} \downarrow$ & {\it Acc.} $\uparrow$ & 
$\delta_{\mathcal{M}} \downarrow$ & $\delta_{\mathcal{T}} \downarrow$ & {\it Acc.} $\uparrow$ & 
$\delta_{\mathcal{M}} \downarrow$ & $\delta_{\mathcal{T}} \downarrow$ & {\it Acc.} $\uparrow$ & 
$\delta_{\mathcal{M}} \downarrow$ & $\delta_{\mathcal{T}} \downarrow$ & {\it Acc.} $\uparrow$ & 
$\delta_{\mathcal{M}} \downarrow$ & $\delta_{\mathcal{T}} \downarrow$ & {\it Acc.} $\uparrow$ &
$\delta_{\mathcal{M}} \downarrow$ & $\delta_{\mathcal{T}} \downarrow$ & {\it Acc.} $\uparrow$ \\

\midrule

--- 
& {\it STD.} 
& {\it 1.00} & {\it 1.00} & {\it 57.43}
& {\it 1.00} & {\it 1.00} & {\it 26.87}
& {\it 1.00} & {\it 1.00} & {\it 10.68}
& {\it 1.00} & {\it 1.00} & {\it 0.86}
& {\it 1.00} & {\it 1.00} & {\it 36.66}
& {\it 1.00} & {\it 1.00} & {\it 52.54}
& {\it 1.00} & {\it 1.00} & {\it 0.56}
\\

\hdashline
\multirow{4}{*}{\tdt}
& AC 
& {\bf 0.18} & \textcolor{red}{1.71} & 55.90
& \underline{0.21} & \textcolor{red}{2.27} & 25.73
& 0.34 & \textcolor{red}{3.56} & 9.85
& 0.41 & \textcolor{red}{3.50} & \underline{0.78}
& 0.37 & \textcolor{red}{2.99} & 36.24
& 0.31 & \textcolor{red}{3.17} & \underline{\textcolor{blue}{52.58}}
& \underline{0.34} & \textcolor{red}{2.12} & \underline{0.51}
\\

& ESC
& 0.34 & \underline{\textcolor{red}{1.40}} & {\bf 56.65}
& 0.53 & \underline{\textcolor{red}{2.03}} & 25.56
& 0.33 & \textcolor{red}{1.64} & \underline{\textcolor{blue}{11.28}}
& 0.21 & \textcolor{red}{2.44} & 0.69
& 0.32 & \textcolor{red}{2.38} & \underline{36.45}
& 0.35 & \underline{\textcolor{red}{2.16}} & {\bf \textcolor{blue}{52.63}}
& 0.50 & \underline{\textcolor{red}{1.88}} & 0.50
\\

& DSC 
& 0.28 & \textcolor{red}{1.75} & 55.38
& 0.33 & \textcolor{red}{2.18} & 25.56
& \underline{0.27} & \underline{\textcolor{red}{1.63}} & 7.98
& {\bf 0.14} & \underline{0.98} & 0.35
& \underline{0.26} & \underline{\textcolor{red}{2.08}} & 35.92
& \underline{0.26} & \textcolor{red}{2.83} & 52.00
& 0.35 & \textcolor{red}{2.44} & 0.43
\\

& DeepConf
& 0.44 & \textcolor{red}{2.37} & 56.12
& 0.48 & \textcolor{red}{2.61} & \underline{25.82}
& 0.67 & \textcolor{red}{4.02} & 7.43
& 0.72 & \textcolor{red}{5.22} & 0.47
& 0.52 & \textcolor{red}{4.17} & 34.49
& 0.63 & \textcolor{red}{4.16} & 49.86
& 0.66 & \textcolor{red}{3.19} & 0.44
\\

\hdashline
\rowcolor{gray!15}
{\bf \tbt} & {\bf \latbp ~(Ours)}
& \underline{0.23} & {\bf 0.41} & \underline{56.62}
& {\bf 0.16} & {\bf 0.37} & {\bf 26.69}
& {\bf 0.22} & {\bf 0.43} & {\bf \textcolor{blue}{11.39}}
& \underline{0.17} & {\bf 0.52} & {\bf 0.80}
& {\bf 0.25} & {\bf 0.37} & {\bf 36.48}
& {\bf 0.22} & {\bf 0.42} & 52.23
& {\bf 0.26} & {\bf 0.45} & {\bf \textcolor{blue}{0.62}}
\\

\bottomrule

\end{tabular}%
}}
\label{tab:llama31-8B-results}%
\end{table}

%% file: tables/r1-results.tex
\begin{table}[H]
\caption{{\bf Empirical Validation of Our \latbp ~for Efficient Parallel Decoding.}
This table shows \texttt{Deepseek-R1-Distill-Qwen-7B} results.
}
\vspace{0.05in}
\centering
\footnotesize
\renewcommand\arraystretch{1.}
\setlength{\tabcolsep}{1.3mm}{
\resizebox{1\textwidth}{!}{
\begin{tabular}{llccccccccccccccccccccc}

\toprule

\multirow{2}{*}{\bf Paradigm}
& \multirow{2}{*}{\bf Method}
& \multicolumn{3}{c}{\bf MATH500}
& \multicolumn{3}{c}{\bf AMC}
& \multicolumn{3}{c}{\bf AIME24}
& \multicolumn{3}{c}{\bf AIME25}
& \multicolumn{3}{c}{\bf GPQA}
& \multicolumn{3}{c}{\bf MMLU-Pro}
& \multicolumn{3}{c}{\bf BrowseComp}
\\
\cmidrule(lr){3-5}
\cmidrule(lr){6-8}
\cmidrule(lr){9-11}
\cmidrule(lr){12-14}
\cmidrule(lr){15-17}
\cmidrule(lr){18-20}
\cmidrule(lr){21-23}

& &
$\delta_{\mathcal{M}} \downarrow$ & $\delta_{\mathcal{T}} \downarrow$ & {\it Acc.} $\uparrow$ & 
$\delta_{\mathcal{M}} \downarrow$ & $\delta_{\mathcal{T}} \downarrow$ & {\it Acc.} $\uparrow$ & 
$\delta_{\mathcal{M}} \downarrow$ & $\delta_{\mathcal{T}} \downarrow$ & {\it Acc.} $\uparrow$ & 
$\delta_{\mathcal{M}} \downarrow$ & $\delta_{\mathcal{T}} \downarrow$ & {\it Acc.} $\uparrow$ & 
$\delta_{\mathcal{M}} \downarrow$ & $\delta_{\mathcal{T}} \downarrow$ & {\it Acc.} $\uparrow$ & 
$\delta_{\mathcal{M}} \downarrow$ & $\delta_{\mathcal{T}} \downarrow$ & {\it Acc.} $\uparrow$ &
$\delta_{\mathcal{M}} \downarrow$ & $\delta_{\mathcal{T}} \downarrow$ & {\it Acc.} $\uparrow$ \\

\midrule

--- &
{\it STD.}
& {\it 1.00} & {\it 1.00} & {\it 94.24}
& {\it 1.00} & {\it 1.00} & {\it 86.78}
& {\it 1.00} & {\it 1.00} & {\it 79.11}
& {\it 1.00} & {\it 1.00} & {\it 68.67}
& {\it 1.00} & {\it 1.00} & {\it 55.29}
& {\it 1.00} & {\it 1.00} & {\it 76.02}
& {\it 1.00} & {\it 1.00} & {\it 17.45}
\\

\hdashline
\multirow{4}{*}{\tdt}
& AC
& 0.28 & \textcolor{red}{1.89} & 93.70
& 0.32 & \textcolor{red}{2.77} & 85.82
& \underline{0.44} & \textcolor{red}{2.50} & 78.64
& 0.37 & \textcolor{red}{2.75} & 67.79
& 0.54 & \textcolor{red}{2.62} & \underline{54.67}
& 0.47 & \textcolor{red}{2.32} & \underline{75.72}
& {\bf 0.28} & \underline{\textcolor{red}{1.76}} & \underline{16.83}
\\

& ESC
& {\bf 0.15} & 0.84 & \underline{94.02}
& \underline{0.27} & \underline{\textcolor{red}{1.72}} & {\bf 86.03}
& 0.53 & \underline{\textcolor{red}{2.19}} & 78.39
& 0.41 & \textcolor{red}{2.19} & 68.12
& \underline{0.43} & \underline{\textcolor{red}{1.67}} & 54.42
& 0.38 & \textcolor{red}{1.92} & 75.37
& 0.45 & \textcolor{red}{1.92} & 14.23
\\

& DSC 
& 0.45 & \textcolor{red}{2.13} & 91.04
& 0.39 & \textcolor{red}{2.89} & 81.28
& 0.56 & \textcolor{red}{3.58} & 73.75
& \underline{0.32} & \underline{\textcolor{red}{2.00}} & 65.30
& 0.60 & \textcolor{red}{4.03} & 52.99
& \underline{0.35} & \underline{\textcolor{red}{1.79}} & 71.14
& 0.46 & \textcolor{red}{2.11} & 15.23
\\

& DeepConf 
& 0.53 & \textcolor{red}{3.86} & 93.45
& 0.58 & \textcolor{red}{3.56} & 85.52
& 0.83 & \textcolor{red}{4.98} & {\bf \textcolor{blue}{79.58}}
& 0.75 & \textcolor{red}{3.57} & {\bf \textcolor{blue}{68.90}}
& 0.65 & \textcolor{red}{3.80} & 53.13
& 0.59 & \textcolor{red}{3.88} & 73.40
& 0.66 & \textcolor{red}{3.12} & 16.09
\\

\hdashline
\rowcolor{gray!15}
{\bf \tbt} & {\bf \latbp ~(Ours)}
& \underline{0.19} & {\bf 0.37} & {\bf \textcolor{blue}{94.32}}
& {\bf 0.22} & {\bf 0.48} & \underline{85.97}
& {\bf 0.34} & {\bf 0.71} & \underline{\textcolor{blue}{79.30}}
& {\bf 0.26} & {\bf 0.50} & \underline{68.28}
& {\bf 0.25} & {\bf 0.45} & {\bf \textcolor{blue}{55.53}}
& {\bf 0.23} & {\bf 0.49} & {\bf 75.84}
& \underline{0.31} & {\bf 0.52} & {\bf 16.96}
\\

\bottomrule

\end{tabular}%
}}
\label{tab:r1-results}%
\end{table}

%% file: 8H-analysis.tex
\section{Extended Analysis}

\subsection{Sensitivity to Sampling Hyperparameters}
\label{appe:sampling-generalization}

We further evaluate the sensitivity of \latbp{} to sampling hyperparameters.
We consider four configurations of top-$k$, top-$p$, and temperature $T$:
\begin{enumerate}
    \setlength{\itemsep}{0pt}
    \setlength{\parsep}{0pt}
    \setlength{\topsep}{2pt}
    \item[(a)] $(20, 0.95, 0.6)$, our main setup;
    \item[(b)] $(40, 0.95, 0.6)$;
    \item[(c)] $(20, 0.8, 0.6)$;
    \item[(d)] $(20, 0.95, 1.0)$.
\end{enumerate}

We evaluate two settings.
The first is \emph{In-Domain (ID) Adaptability}, where the predictor is trained and evaluated under the same sampling configuration.
This setting tests whether \latbp{} remains effective when the sampling hyperparameters are changed consistently across training and decoding.
The second is \emph{Out-of-Domain (OOD) Generalization}, where the predictor is trained under one configuration but evaluated under another.
This setting tests whether \latbp{} remains robust under training--decoding hyperparameter mismatch.
Results are reported in \textcolor{deepred}{Table \ref{tab:hyperparameter-ablation}}.

In the ID setting, \latbp{} shows strong adaptability to sampling-hyperparameter changes.
Across the four configurations, its accuracy gaps relative to the corresponding \textit{STD.} baselines are $+0.02$, $-0.23$, $-0.24$, and $-0.07$, respectively.
All gaps remain within a small range, indicating that \latbp{} remains stable when $k$, $p$, and $T$ are changed consistently between training and decoding.

In the OOD setting, \latbp{} faces a harder generalization problem, since the predictor is trained and evaluated under different sampling configurations.
Nevertheless, the degradation remains mild, with the worst accuracy drop being only $0.50$ points.
We also observe a rough correlation between cost mismatch and accuracy drop: when the training configuration induces a budget scale farther from the inference-time optimum, performance tends to degrade more.
This is expected because the predicted budget is tied to the sampling distribution used to construct training labels.
Overall, these results suggest that \latbp{} is not overly sensitive to sampling hyperparameters and remains robust under moderate training--decoding mismatch.

\begin{table}[H]
\vspace{-0in}
\caption{
Sensitivity of \latbp{} to sampling hyperparameters on \texttt{Qwen3-4B} with \texttt{AIME25}.
The upper panel reports \textit{STD.} accuracy under each inference configuration.
In the lower panel, {\bf Tr.} and {\bf In.} denote the sampling configurations used for predictor training and decoding, respectively.
\textcolor{gray}{Gray} cells indicate ID evaluation; other cells indicate OOD evaluation.
}
\vspace{0.05in}
\centering
\footnotesize
\renewcommand\arraystretch{1.}
\setlength{\tabcolsep}{1.5mm}{
\resizebox{1\columnwidth}{!}{
\begin{tabular}{lcccccccccccc}

\toprule
\multirow{2}{*}{} & \multicolumn{12}{c}{\it Acc. $\uparrow$}
\\

\cmidrule(lr){2-13}
{\it STD.} 
& \multicolumn{3}{c}{(a) 54.53}
& \multicolumn{3}{c}{(b) 54.18} 
& \multicolumn{3}{c}{(c) 54.35} 
& \multicolumn{3}{c}{(d) 53.97}
\\

\bottomrule
\\
\toprule

\multirow{2}{*}{\diagbox{\bf Tr.}{\bf In.}}
& \multicolumn{4}{c}{$\delta_{\mathcal{M}} \downarrow$}
& \multicolumn{4}{c}{$\delta_{\mathcal{T}} \downarrow$}
& \multicolumn{4}{c}{{\it Acc.} $\uparrow$}
\\

\cmidrule(lr){2-5}
\cmidrule(lr){6-9}
\cmidrule(lr){10-13}

& (a) & (b) & (c) & (d)
& (a) & (b) & (c) & (d)
& (a) & (b) & (c) & (d)
\\
\midrule

(a) 
& \cellcolor{gray!20} 0.17 & 0.17 & 0.19 & 0.18
& \cellcolor{gray!20} 0.56 & 0.55 & 0.59 & 0.56
& \cellcolor{gray!20} 54.55 & 53.81 & 54.02 & 53.57
\\

(b)
& 0.19 & \cellcolor{gray!20} 0.18 & 0.19 & 0.19
& 0.59 & \cellcolor{gray!20} 0.60 & 0.55 & 0.57
& 54.34 & \cellcolor{gray!20} 53.95 & 54.03 & 53.93
\\

(c)
& 0.15 & 0.15 & \cellcolor{gray!20} 0.15 & 0.16
& 0.54 & 0.53 & \cellcolor{gray!20} 0.53 & 0.57
& 54.32 & 53.72 & \cellcolor{gray!20} 54.11 & 53.61
\\

(d)
& 0.21 & 0.23 & 0.21 & \cellcolor{gray!20} 0.23
& 0.61 & 0.59 & 0.61 & \cellcolor{gray!20} 0.62
& 54.18 & 53.74 & 53.85 & \cellcolor{gray!20} 53.90
\\

\bottomrule

\end{tabular}%
}}
\label{tab:hyperparameter-ablation}%
\vspace{-0.in}
\end{table}

\subsection{Ablation Study of \latbp{}}
\label{appe:LBA-ablation}

\paragraph{Hidden Size of MLP Architecture.}
The only architectural hyperparameter of the predictor is the MLP hidden size, set as $\lfloor rd \rfloor$, where $d$ is the hidden dimension of the language model.
In our main experiments, we use $r=1/8$.
We vary $r$ to study how predictor capacity affects \latbp{} in real-world decoding.

\begin{figure}[H]
    \centering
    \includegraphics[width=1\textwidth]{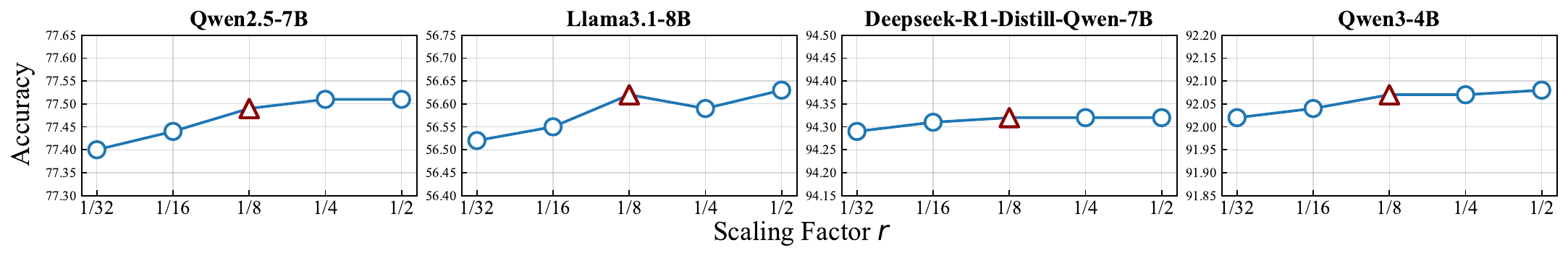}
    \vspace{-0.2in}
    \caption{Decoding accuracy under varying scaling factor $r$ of MLP hidden size $\lfloor rd \rfloor$ on the \texttt{MATH500} dataset.}
    \label{fig:ablation-hidden-size-math}
\end{figure}

\begin{figure}[H]
    \centering
    \includegraphics[width=1\textwidth]{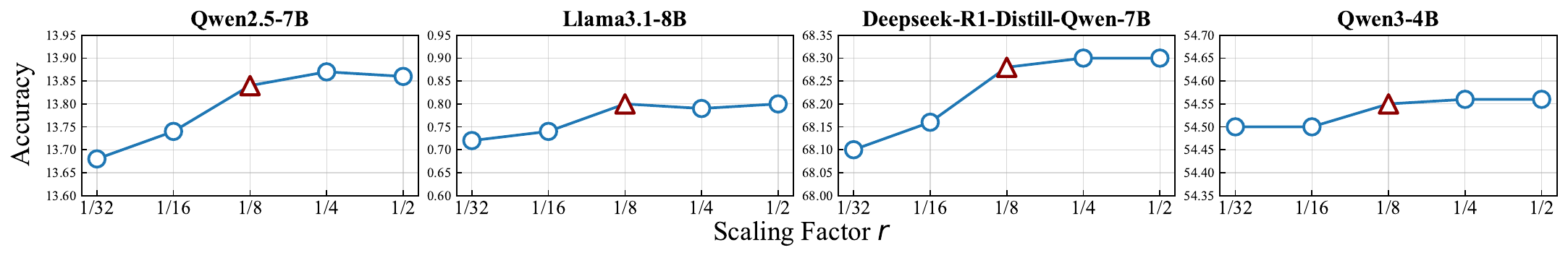}
    \vspace{-0.2in}
    \caption{Decoding accuracy under varying scaling factor $r$ of MLP hidden size $\lfloor rd \rfloor$ on the \texttt{AIME25} dataset.}
    \label{fig:ablation-hidden-size-aime25}
\end{figure}

\textcolor{deepred}{Figures \ref{fig:ablation-hidden-size-math} and \ref{fig:ablation-hidden-size-aime25}} show the results on \texttt{MATH500} and \texttt{AIME25}.
When $r<1/8$, decoding accuracy drops to different extents across models, indicating that insufficient predictor capacity harms budget prediction.
When $r>1/8$, performance largely saturates, suggesting that $r=1/8$ already provides sufficient capacity and further scaling brings limited benefit.

\paragraph{Training Data Size.}
We also study the effect of training data size for the predictor.
In our main experiments, the predictor is trained with 5k samples.
Here, we vary the number of training samples to examine how data scale affects \latbp{} in real-world decoding.

\begin{figure}[H]
    \vspace{-0.in}
    \centering
    \includegraphics[width=1\textwidth]{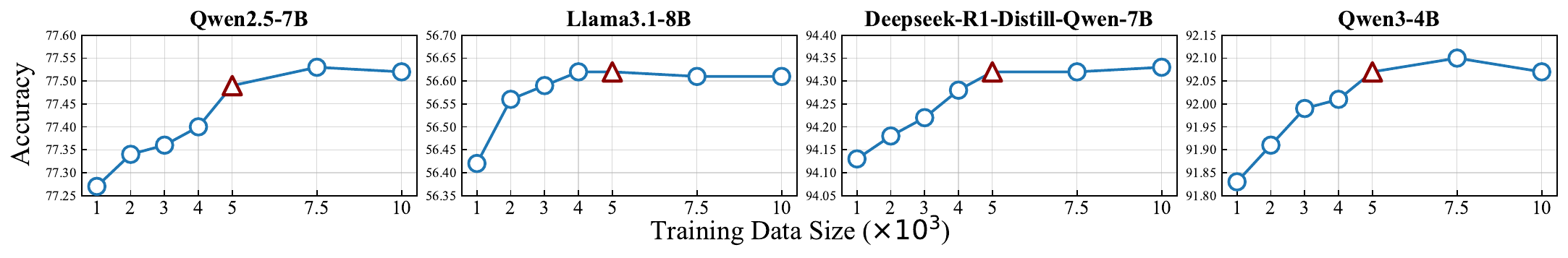}
    \vspace{-0.2in}
    \caption{Decoding accuracy under varying training data sizes on the \texttt{MATH500} dataset.}
    \vspace{-0.1in}
    \label{fig:ablation-training-size-math}
\end{figure}

\begin{figure}[H]
    \vspace{-0.in}
    \centering
    \includegraphics[width=1\textwidth]{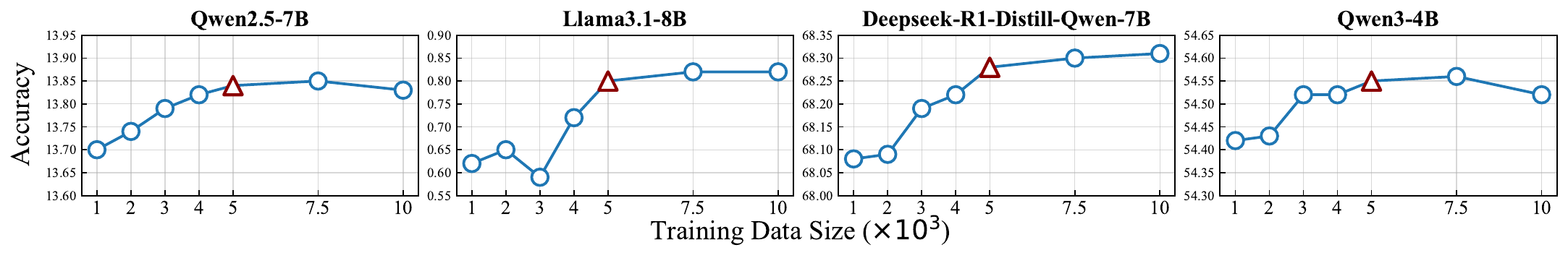}
    \vspace{-0.2in}
    \caption{Decoding accuracy under varying training data sizes on the \texttt{AIME25} dataset.}
    \vspace{-0.1in}
    \label{fig:ablation-training-size-aime25}
\end{figure}

\textcolor{deepred}{Figures \ref{fig:ablation-training-size-math} and \ref{fig:ablation-training-size-aime25}} report the results on \texttt{MATH500} and \texttt{AIME25}.
Overall, performance improves rapidly with more training data at small scales, but the gains become much slower after 5k samples.
At 10k samples, the improvement is marginal, and mild overfitting is even observed for \texttt{Qwen}.
These results suggest that 5k samples provide the best efficiency-performance trade-off, and we therefore use 5k as the default training size in our main experiments.

\paragraph{Inference Weighting Strategy.}
During inference, \latbp{} combines the outputs of the $L$ layer-wise predictors into a single budget prediction, as defined in \textcolor{deepred}{Eq.\ref{eq:aggregation-general}}.
Our theoretical derivation leads to \emph{Inverse Squared Error Weighting} (\textcolor{deepred}{Eq.\ref{eq:final-weight}}), where more reliable predictors receive larger weights.
To verify whether this choice is necessary in practice, we compare it with two simpler alternatives:
(1) \emph{Uniform Weighting}, which assigns equal weights to all layers, and
(2) \emph{One-hot Weighting}, which uses only the predictor with the lowest validation error.

\textcolor{deepred}{Table \ref{tab:weighting-strategies}} reports the results on all datasets with \texttt{Qwen3-4B}.
Inverse Squared Error Weighting consistently achieves the best performance across all datasets, confirming the practical effectiveness of our theoretical weighting rule.
Compared with One-hot Weighting, our method benefits from aggregating complementary information across multiple layers rather than relying on a single best layer.
Compared with Uniform Weighting, it avoids treating unreliable layers equally, which explains why simple averaging performs the worst.
These results suggest that both cross-layer information and reliability-aware weighting are important for stable budget prediction.

\begin{table}[H]
\vspace{-0in}
\caption{Comparison of decoding accuracy across inference weighting strategies for \latbp{} on \texttt{Qwen3-4B}.}
\vspace{0.05in}
\centering
\footnotesize
\renewcommand\arraystretch{1.}
\setlength{\tabcolsep}{1.5mm}{
\resizebox{1\columnwidth}{!}{
\begin{tabular}{lccccccc}
\toprule
& \textbf{MATH500} & \textbf{AMC} & \textbf{AIME24} & \textbf{AIME25} & \textbf{GPQA} & \textbf{MMLU-Pro} & \textbf{BrowseComp} \\
\midrule
\textbf{Inverse Squared Error Weighting (Ours)} & 92.88 & 82.18 & 59.35 & 54.53 & 65.24 & 71.87 & 11.23 \\
One-hot Weighting & 91.54 & 81.43 & 56.68 & 53.92 & 64.88 & 71.79 & 10.23 \\
Uniform Weighting & 89.49 & 80.09 & 56.63 & 53.24 & 64.34 & 70.82 & 9.93 \\
\bottomrule
\end{tabular}%
}}
\label{tab:weighting-strategies}
\vspace{-0.in}
\end{table}

\subsection{Design Principle of \latbp{}}
\label{appe:LBA-principle}

\paragraph{Layer-wise Pattern.}
We train $L$ layer-wise predictors and combine their outputs during inference.
A simpler alternative is to concatenate the hidden states from all $L$ layers and train a single predictor.
However, this concatenated-state design mixes heterogeneous representations and greatly increases the feature dimension, making the predictor more prone to sparsity and overfitting under limited training data.
In contrast, the layer-wise design preserves depth-specific signals by modeling each layer separately.

We compare these two designs to examine whether preserving layer-wise structure is important for \latbp{}.
\textcolor{deepred}{Table \ref{tab:layer-wise-pattern}} reports the results on all datasets with \texttt{Qwen3-4B}.
Our layer-wise design consistently achieves stronger decoding accuracy, supporting layer-wise prediction with inference-time combination as a more effective design for \latbp{}.

\begin{table}[H]
\vspace{-0in}
\caption{Decoding accuracy comparison of layer-wise and concatenated-state predictors for \latbp{} on \texttt{Qwen3-4B}.}
\vspace{0.05in}
\centering
\footnotesize
\renewcommand\arraystretch{1.}
\setlength{\tabcolsep}{1.5mm}{
\resizebox{1\columnwidth}{!}{
\begin{tabular}{lccccccc}
\toprule
& \textbf{MATH500} & \textbf{AMC} & \textbf{AIME24} & \textbf{AIME25} & \textbf{GPQA} & \textbf{MMLU-Pro} & \textbf{BrowseComp} \\
\midrule
\textbf{Layer-wise Predictors (ours)}
& 92.88 & 82.18 & 59.35 & 54.53 & 65.24 & 71.87 & 11.23 \\
Single Predictor w/ Concatenated States
& 90.09 & 79.80 & 59.12 & 52.90 & 64.23 & 70.16 & 10.24 \\
\bottomrule
\end{tabular}%
}}
\label{tab:layer-wise-pattern}
\vspace{-0.in}
\end{table}

\paragraph{Training Objective.}
We formulate sample-optimal budget prediction as a regression task.
A natural alternative is to treat budgets as discrete labels and train an $N_{\max}$-class classifier.
However, budget prediction is inherently distance-sensitive: a small deviation from $N_{\bm{x}}^*$ usually has limited impact, while a large deviation can substantially affect performance.
Regression naturally preserves this distance structure by penalizing larger errors more strongly.
In contrast, classification treats budgets as independent classes and does not explicitly model their distances to the target budget.

We compare these two objectives to examine whether preserving distance information is important for \latbp{}.
\textcolor{deepred}{Table \ref{tab:regression-vs-classification}} reports the results on \texttt{Qwen3-4B}.
Regression consistently outperforms classification across all datasets, confirming that distance-aware regression is a more suitable objective for predicting sample-optimal budgets.

\begin{table}[H]
\vspace{-0in}
\caption{Comparison of decoding accuracy between regression and classification objectives of predictor training for \latbp{} on \texttt{Qwen3-4B}.}
\vspace{0.05in}
\centering
\footnotesize
\renewcommand\arraystretch{1.}
\setlength{\tabcolsep}{1.5mm}{
\resizebox{1\columnwidth}{!}{
\begin{tabular}{lccccccc}
\toprule
& \textbf{MATH500} & \textbf{AMC} & \textbf{AIME24} & \textbf{AIME25} & \textbf{GPQA} & \textbf{MMLU-Pro} & \textbf{BrowseComp} \\
\midrule
\textbf{Regressor (Ours)} & 92.88 & 82.18 & 59.35 & 54.53 & 65.24 & 71.87 & 11.23 \\
$N_{\max}$-class Classifier & 88.54 & 79.64 & 55.28 & 50.13 & 62.73 & 69.22 & 8.32\\
\bottomrule
\end{tabular}%
}}
\label{tab:regression-vs-classification}
\vspace{-0.in}
\end{table}

%% file: 8I-related-work.tex
\section{Related Work}
\label{sec:related-work}

\subsection{Parallel Thinking}
\label{sec:related-parallel}

Parallel thinking is a test-time scaling paradigm that mainly consists of two stages:
\textit{\textbf{multi-path sampling}} and \textit{\textbf{answer aggregation}} \citep{li2025parallelmuse}.
In the first stage, stochastic sampling is commonly adopted, where each reasoning path is independently explored from scratch via multinomial sampling \citep{wang2022self,lightman2023let,brown2024large,chen2024more,kang2025scalable,wang2025sampling}, including variants that partially relax path independence \citep{rodionov2025hogwild,dong2025generalized}.
In the second stage, majority voting is the most widely used unsupervised aggregation method, offering strong generality, simplicity, and minimal prior assumptions \citep{wang2022self,xiong2023can,team2023gemini,liang2024encouraging}.
Beyond majority voting, some methods incorporate additional priors into aggregation, such as confidence or hidden-state signals \citep{kang2025scalable,fu2025deep,wang2024embedding,wang2024latent}, while others introduce external priors by scoring candidates with reward models \citep{aggarwal2023let,wang2024interpretable} or LLMs \citep{jiang2023llm,xiong2023can}.

Beyond this standard sampling-and-aggregation setting, some studies explore heuristic methods for multi-path exploration, providing finer-grained control during decoding.
For example, Beam Search \citep{wiseman2016sequence} improves lookahead by maintaining multiple hypotheses at each decoding step.
Other strategies, such as Tree-of-Thought (ToT) \citep{yao2023tree}, Skeleton-of-Thought (SoT) \citep{ning2023skeleton}, and Monte Carlo Tree Search (MCTS) \citep{zhang2024accessing,guan2025rstar,li2025treepo,ding2025dynamic}, build on stochastic rollouts or structured expansions, introducing additional hyperparameters to control branching, merging, truncation, and back-propagation of sampling paths.

Another line of work investigates training-based methods for further unlocking parallel thinking capabilities in LLMs, including SFT \citep{wen2025parathinker,yang2025multiverse}, RL \citep{zheng2025parallel,wu2025native}, and architectural parallelism \citep{chen2025parallel}.
Since our focus is on training-free methods, these studies are orthogonal to our work.
Overall, parallel thinking provides a promising avenue for unlocking the latent reasoning capabilities of LLMs.

\subsection{Efficient Parallel Decoding}

Efficient parallel decoding mainly follows two research directions \citep{wang2025sampling}: 
\textit{\textbf{breadth pruning}} and \textit{\textbf{depth pruning}}.
Breadth pruning adaptively reduces the number of parallel samples for each input, and our \latbp{} belongs to this line.
Prior methods, including AC \citep{aggarwal2023let}, ESC \citep{li2024escape}, and DSC \citep{wang2025make}, rely on answer frequency or problem difficulty for dynamic early stopping.
However, our experiments show that such decoding-time adaptation can incur substantial latency due to blocked hardware parallelization.

Depth pruning, in contrast, keeps the overall sampling budget fixed while shortening individual reasoning paths.
For example, SBoN \citep{sun2024fast} uses a reward model to score early partial sequences and discard low-scoring paths, ST-BoN \citep{wang2025sampling} truncates unpromising samples based on internal consistency estimates, and Slim-SC \citep{hong2025slim} prunes highly similar paths on the fly during decoding.
These two lines are {\bf orthogonal}: breadth pruning improves efficiency by reducing the number of paths, while depth pruning improves efficiency by reducing the length or cost of each path.
{\bf Our \latbp{} belongs to breadth pruning}, as its efficiency gains come from predicting and allocating sample-specific parallelism levels before decoding.

Although orthogonal, the two directions are not mutually exclusive and can be combined.
For example, DeepConf \citep{fu2025deep} performs path-level truncation within an adaptive early-stopping framework.
Similarly, our \tbt{} paradigm can be seamlessly integrated with depth-pruning methods: once \latbp{} predicts the optimal parallelism level before decoding, path-level pruning can still be applied during decoding without interfering with the pre-decoding budget allocation.

\subsection{Test-time Scaling}

Test-time scaling \citep{snell2024scaling} improves reasoning performance by increasing computation at inference time.
This paradigm mainly follows two directions \citep{muennighoff2025s1}: \textit{\textbf{sequential scaling}} and \textit{\textbf{parallel scaling}}.
Sequential scaling extends the length of a single chain-of-thought, encouraging slower and more deliberate reasoning.
This can elicit cognitive behaviors such as reflection and backtracking \citep{guo2025deepseek,gandhi2025cognitive}, thereby increasing the chance of reaching the correct answer.
Representative approaches include RL \citep{shao2024deepseekmath,guo2025deepseek,yu2025dapo}, SFT \citep{muennighoff2025s1,ye2025limo,yang2025qwen3}, and inference-time prompt forcing \citep{muennighoff2025s1,wang2025polymath}.
Parallel scaling corresponds to parallel thinking, which is discussed in detail in \textcolor{deepred}{Appendix \ref{sec:related-parallel}}.
These two paradigms are complementary and together provide a foundation for eliciting stronger reasoning capabilities from LLMs.

\subsection{Efficiency Issue in Test-time Scaling}

Among TTS paradigms, overthinking \citep{chen2025not} is a well-known efficiency issue in sequential scaling. Although the overscaling curse also ultimately results in an efficiency issue, it should not be viewed as a simple parallel counterpart of overthinking \citep{chen2025not}, even though both belong to the broader TTS paradigm. They are fundamentally different. Overthinking is a \textbf{\emph{within-sample}} issue, where excessive computation on a sample creates a trade-off between its own efficacy and efficiency. By contrast, overscaling arises from a bidirectional coupling between the system and samples: individual sample behaviors collectively determine system efficacy, which is then imposed back on the efficiency of all samples. This \textbf{\emph{sample--system--sample}} constraint creates cross-sample interactions, making overscaling far more complex.

\section{Limitations}
\label{sec:limitations}

The current work has two limitations.
First, we do not consider open-ended tasks, since the lack of reliable ground truth makes it difficult to define the sample-optimal budget.
Indeed, evaluating open-ended outputs remains an open challenge.
Once suitable evaluation protocols become available, \latbp{} can be naturally extended to such settings.
Second, \latbp{} requires access to internal model states, and therefore currently applies mainly to open-source models.
However, with the rapid growth of the open-source LLM ecosystem \citep{yang2025qwen3,guo2025deepseek}, we believe that this limitation does not diminish our contribution.

\section{Broader Impacts}
\label{sec:broader-impacts}

This work aims to improve the efficiency of parallel thinking by reducing redundant budget.
Its positive impact is that more efficient budget allocation can reduce inference latency, memory usage, and computational cost, making advanced reasoning systems more accessible under limited hardware resources.

We do not identify negative societal impacts specific to this work.
Our method does not introduce new model capabilities; it only reallocates inference budgets for existing parallel decoding pipelines.
Potential risks mainly come from the underlying LLMs and their downstream applications, rather than from the budget prediction method itself.
Therefore, responsible deployment should follow the safety and evaluation practices required for the base models and target applications.